\documentclass[10pt,twocolumn,letterpaper]{article}

\usepackage{cvpr}
\usepackage{times}
\usepackage{epsfig}
\usepackage{graphicx}
\usepackage{amsmath}
\usepackage{amssymb}
\usepackage{float}
\usepackage{authblk}

\usepackage[pagebackref=true,breaklinks=true,letterpaper=true,colorlinks,bookmarks=false]{hyperref}

\cvprfinalcopy 

\def\cvprPaperID{4806} 

\ifcvprfinal\fi
\usepackage{xcolor}
\usepackage{wrapfig}
\usepackage{amsmath}
\usepackage{amsfonts}
\usepackage{color}
\usepackage{multirow}
\usepackage{rotating}

\usepackage[]{algorithm2e}

\DeclareMathOperator{\R}{\mathbb{R}}

\begin{document}

\title{Deep Geometric Prior for Surface Reconstruction}

\author[1]{Francis Williams}
\author[1]{Teseo Schneider}
\author[1]{Claudio Silva }
\author[1]{Denis Zorin}
\author[1]{Joan Bruna}
\author[1]{Daniele Panozzo}
\affil[1]{New York University}
\affil[ ]{\tt\small francis.williams@nyu.edu, teseo.schneider@nyu.edu, csilva@nyu.edu, dzorin@cs.nyu.edu, bruna@cims.nyu.edu, panozzo@nyu.edu}

\maketitle

\begin{abstract}
The reconstruction of a discrete surface from a point cloud is a fundamental geometry processing problem that has been studied for decades, with many methods developed. 
We propose the use of a deep neural network as a \emph{geometric prior} for surface reconstruction. Specifically, we overfit a neural network representing a local chart parameterization to part of an input point cloud using the Wasserstein distance as a measure of approximation. By jointly fitting many such networks to overlapping parts of the point cloud, while enforcing a consistency condition, we compute a \emph{manifold atlas}. By sampling this atlas, we can produce a dense reconstruction of the surface approximating the input cloud. The entire procedure does not require any training data or explicit regularization, yet, we show that it is able to perform remarkably well: not introducing typical overfitting artifacts, and approximating sharp features closely at the same time. 
We experimentally show that this geometric prior produces good results for both man-made objects containing sharp features and smoother organic objects, as well as noisy inputs.
We compare our method with a number of well-known reconstruction methods on a standard surface  reconstruction benchmark.
\end{abstract}

\section{Introduction}

3D geometry is commonly acquired in the form of collections of (possibly incomplete) range images (laser scanning, structured light, etc) or measurements of more complex structure (LIDAR). 
Unordered set of points (point clouds) is a commonly used representation of combined registered results of scanning objects or scenes.  Point clouds can be obtained in other ways, e.g., by 
sampling an implicit surface using ray casting. Computing a continuous representation of a surface 
from the discrete point cloud (e.g., a polygonal mesh, an implicit surface, or a set of parameteric patches)  in way that is robust to noise, and yet retains critical surface features and approximates the sampled surface well, is a pervasive and challenging problem.

Different approaches have been proposed, mostly grouped into several categories: (1) using the points to define a volumetric scalar function whose 0 level-set corresponds to the desired surface, (2) attempt to "connect the dots" in a globally consistent way to create a mesh, (3) fit a set of primitive shapes so that the boundary of their union is close to the point cloud, and (4) fit a set of patches to the point cloud approximating the surface. 

We propose a novel method, based, on the one hand, on constructing a manifold atlas commonly used in differential geometry to define a surface, and, on the other hand, on observed remarkable properties of deep image priors \cite{ulyanov2017deep}, using an overfitted neural network for interpolation. 
We define a set of 2D parametrizations, each one mapping a square in parametric space to a region of a surface, ensuring consistency between neighbouring patches. This problem is inherently ambiguous: there are many possible valid parametrizations, and only a small subset will correspond to a faithful representation of the underlying surface. We compute each parametrization by overfiting a network to a
part of the point cloud, while enforcing consistency conditions between different patches. We observe that the result is a reconstruction which is superior  quantitatively and qualitatively to commonly used surface reconstruction methods.

We use the Wasserstein distance as a training loss, which is naturally robust to outliers, and has the added advantage of providing us explicit correspondences between the parametric domain coordinates and the fitted points, allowing us to explicitly measure, and thus minimize, the disagreement between neighbouring patches covering a surface. 

We use a standard shape reconstruction benchmark to compare our method with 12 competing methods, showing that, despite the conceptual simplicity of our algorithm, our reconstructions are superior in terms of quantitative and visual quality. 
\section{Related work}
\label{sec:related}

\paragraph{Geometric Deep Learning}
A variety of architectures were proposed for geometric applications. A few work with point clouds as input; in most cases however, these methods are designed for classification or segmentation tasks.
One of the first examples are PointNet \cite{Qi:2017} and PointNet++ \cite{Qi:2017a} originally designed for classification and segmentation, using a set-based neural architecture \cite{vinyals2015order, sukhbaatar2016learning}. PCPNet \cite{pcpnet} is version of PointNet architecture, for estimation of local shape properties. A number of learning architectures for 3D geometry work with voxel data  converting input point clouds to this form (e.g.,  \cite{varley2017shape}). The closest problems these types of networks solve is shape completion and point cloud upsampling. 

Shape completion is considered, e.g.,  in \cite{dai2017shape}, where volumetric CNN is used to predict a very course shape completion, which is then refined using data-driven local shape synthesis on small volumetric patches.  \cite{han2017high}, follows a somewhat similar approach, combining multiview and volumetric low-resolution global data at a first stage, and using a volumetric network to synthesize local patches to obtain higher resolution. Neither of these methods aims to achieve high-quality surface reconstruction. 

PU-Net, described in \cite{yu2018pu}, is to the best of our knowledge, the only learning-based work addressing point cloud upsampling directly. The method proceeds by splitting input shapes into patches and learning hierarchical features using PointNet++ type architecture. Then feature aggregation and expansion is used to perform point set expansion in feature space, followed by the (upsampled) point set reconstruction. 

In contrast to other methods, the untrained networks in our method take parametric coordinates in square parametric domains as inputs and produce surface points as output.  An important exception is the recent work \cite{groueix2018atlasnet} defining an architecture, AtlasNet, in which the decoder part is similar to ours, but with some important distinctions discussed in Section~\ref{sec:method}. 
Finally, \cite{basri2016efficient} studied the ability of neural networks to approximate low-dimensional manifolds, showing that even two-layer ReLU networks have remarkable ability to encode smooth structures with near-optimal number of parameters. In our setting, we rely on overparametrisation and leverage the implicit optimization bias of gradient descent. 

\paragraph{Surface Reconstruction}
is an established research area dating back at least to the early 1990s (e.g, \cite{hoppe1992surface}); \cite{BergerTSAGLSS17} is a recent comprehensive survey of the area. We focus our discussion on the techniques that we use for comparison, which are a superset of those included in the surface reconstruction benchmark of Berger et al~\cite{BergerLNTS13}. Berger tested 10 different techniques in their paper; we will follow their nomenclature. They separate techniques into four main categories: indicator function, point-set surfaces, multi-level partition of unity, and scattered point meshing. 

Indicator function techniques define a scalar function in space that can be used for testing if a given point is inside or outside the surface. There are multiple ways to define such a function from which a surface is generated by isocontouring. Poisson surface reconstruction ({\bf Poisson})~\cite{kazhdan2006poisson} inverts the gradient operator by solving a Poisson equation to define the indicator function. Fourier surface reconstruction ({\bf Fourier})~\cite{kazhdan2005reconstruction} represents the indicator function in the Fourier basis, while Wavelet surface reconstruction ({\bf Wavelet})~\cite{manson2008streaming} employs a Haar or a Daubechies (4-tap) basis. Screened Poisson surface reconstruction ({\bf Screened})~\cite{KazhdanH13} is an extension of \cite{kazhdan2006poisson} that incorporates point constraints to avoid over smoothing of the earlier technique. This technique is not considered in \cite{BergerLNTS13}, but is considered by us.

Point-set surfaces \cite{alexa2001point} define a projection operator that moves points in space to a point on the surface, where the surface is defined to be the collection of stationary points of the projection operator. Providing a definition of the projection operators are beyond the scope of our paper (see \cite{BergerLNTS13}). In our experiments, we have used simple point set surfaces ({\bf SPSS})~\cite{adamson2003approximating}, implicit moving
least squares ({\bf IMLS})~\cite{kolluri2005provably}, and algebraic point set surfaces ({\bf APSS})~\cite{guennebaud2007algebraic}.

Edge-Aware Point Set Resampling ({\bf EAR})~\cite{WGCAZ13} (also not considered in \cite{BergerLNTS13}, but considered by us) works by first computing reliable normals away from potential singularities, followed by a resampling step with a novel bilateral filter towards surface singularities. Reconstruction can be done using different techniques on the resulting augmented point set with normals.

Multi-level Partition of Unity defines an implicit function by integrating weight function of a set of input points. The original approach of~\cite{ohtake2003multi} ({\bf MPU}) uses linear functions as low-order implicits, while \cite{nagai2009smoothing} ({\bf MPUSm}) defines differential operators directly
on the MPU function. The method of \cite{ohtake20043d} ({\bf RBF}) uses compactly-supported radial basis functions. 

Scattered Point Meshing~\cite{ohtake2005integrating} ({\bf Scattered}) grows weighted spheres around points in order to determine the connectivity in the output triangle mesh. 

The work in \cite{pietroni2011global} uses a manifold-based approach to a direct construction of a global parametrization from a set of range images (a point cloud, or any other surface representation, can be converted to such a set by projection to multiple planes). 
It uses range images as charts with projections as chart maps; our method computes 
chart maps by fitting. \cite{xiong2014robust} jointly optimizes for connectivity and geometry to produce a single mesh for an entire input point cloud. In contrast, our method produces a global chart map using only a local optimization procedure.

\paragraph{Deep Image Prior}  Our approach is inspired, in part, by the \emph{deep image prior}. \cite{ulyanov2017deep} demonstrated that an untrained deep network can be overfitted to 
input data producing a remarkably high-quality upsampling and even hole filling without training, with the convolutional neural network structure acting as a regularizer. Our approach to surface reconstruction is similar, in that we use untrained networks to represent individual chart embedding functions. However, an important difference is that our loss function measures geometric similarity. 

\section{Method}
\label{sec:method}

\begin{figure}[t]
    \centering\footnotesize
    \includegraphics[width=1\linewidth]{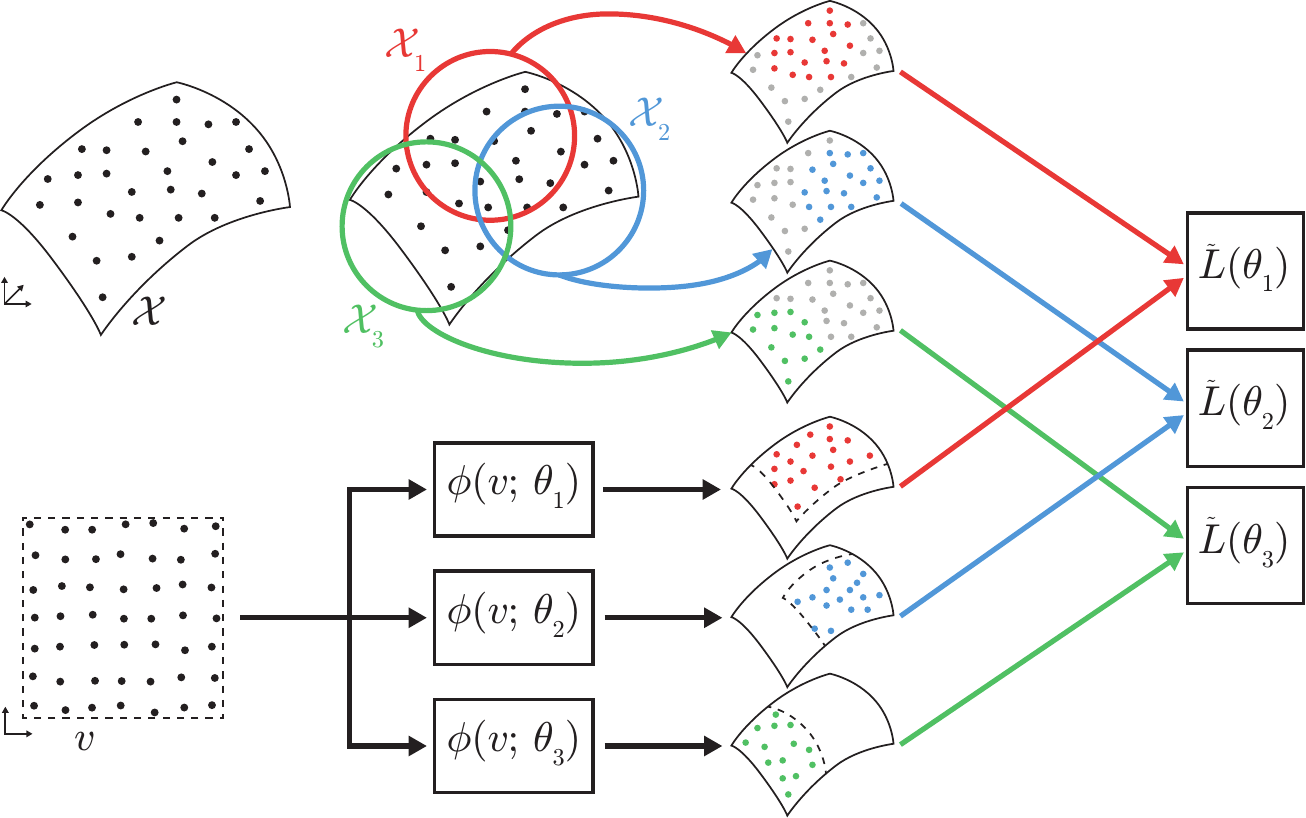}
    \caption{Overview of our proposed surface reconstruction architecture. An input point cloud $\mathcal{X}$ is split into overlapping patches $\mathcal{X}_p$, where a local chart $\varphi_p(v) = \phi(v, \theta_p)$ is obtained via the minimisation of a Wasserstein loss $\tilde{L}(\theta)$ \eqref{bibi2}. The local charts are made globally consistent thanks to the correspondences arising during the Wasserstein loss minimization (Section \ref{globalsec})}.
    \label{fig:architecture}
\end{figure}

Our method for surface reconstruction uses a set of deep ReLU networks to obtain local \emph{charts} or parametrizations (Section~\ref{localchart}). These parameterizations are then made consistent with each other on the parts where they overlap (Section~\ref{globalsec}). The networks are trained using the 2-Wasserstein distance as a loss function. The overall architecture of our technique is illustrated in Figure \ref{fig:architecture}.

In the following, we denote by $\mathcal{S}$ a smooth surface (possibly with a boundary) in  $\R^3$. The goal of surface reconstruction is to estimate $\mathcal{S}$ from a possibly noisy point cloud $\mathcal{X}= \{x_i=y_i + w_i;\,  y_i\in \mathcal{S}; i \leq N \}$, where $w_i$ models the acquisition noise. 

\subsection{Local Parametrization Model}
\label{localchart}

Let us first consider the reconstruction of a given 
neighborhood of $\mathcal{S}$ around a point $p$, denoted as
$\mathcal{U}_p$, from the corresponding point cloud $\mathcal{X}_p = \mathcal{X} \cap \mathcal{U}_p$. 
If $\mathcal{U}_p$ is sufficiently small, 
from the implicit function theorem, one can characterize $\mathcal{U}_p \cap \mathcal{S}$ as the image of the open square 
$V= (0,1)^2$ by a differentiable mapping $\varphi \colon V \to \mathcal{U}_p$. 

We propose to approximate $\varphi$ using a neural network, 
$\phi(\cdot; \theta_p) \colon V \to \R^3$, where $\theta_p$ is a vector of 
parameters, that we fit so that the image of $\phi$ approximates $\mathcal{U}_p \cap \mathcal{S}$.  For that purpose, we first consider a sample $\{v_1, \dots, v_n\}$ of $n=|\mathcal{X}_p|$ points in $V$ using a Poisson disk distribution, and the associated Earth Movers or 2-Wasserstein Distance (EMD)  between $\{ \phi(v_i; \theta_p) : i = 1 \ldots n\}$ and $\mathcal{X}_p=\{x_1\dots, x_n\}$:
\begin{equation}
\label{bibi}
  L(\theta) = \inf_{\pi \in \Pi_n} \sum_{i\leq n} \| \phi(v_i; \theta) - x_{\pi(i)} \|^2.
\end{equation}
where $\Pi_n$ is the set of permutations of $n$ points.
The computation of the EMD in \eqref{bibi} requires solving a linear assignment problem, which can be done in $O(n^3)$ 
using, for instance, the Hungarian algorithm \cite{kuhn1955hungarian}. Since this is prohibitive for typical values of $n$, we rely instead on the Sinkhorn regularized distance \cite{cuturi2013sinkhorn}: 
\begin{equation}
\label{bibi2}
    \tilde{L}(\theta) = \min_{P \in \mathcal{P}_n} \sum_{i,j \leq n} P_{i,j} \| \phi(v_j; \theta) - x_{i} \|^2 - \lambda^{-1} H(P) 
\end{equation}
where $\mathcal{P}_n$ is the set of $n \times n$ bi-stochastic matrices and $H(P)$ is the entropy, $H(P) = -\sum_{i,j} P_{ij}\log P_{ij}$. This distance provably approximates the Wasserstein metric as $\lambda \to \infty$ and can be computed in near-linear time \cite{altschuler2017near}. Figure~\ref{fig:lambda-fig} in the supplemental material shows the effect of varying the regularization parameter $\lambda$ on the results.

We choose $\phi$ to be a MLP with the half-rectified activation function: 
$$
\phi(v; \theta)= \theta_K \mathrm{ReLU}( \theta_{K-1} \mathrm{ReLU} \dots \mathrm{ReLU}(\theta_1 v)),
$$
where $\theta_i$, $i=1\ldots K$, are per-layer weight matrices. 
This choice of activation function implies that we are 
fitting a piece-wise linear approximation to $\mathcal{X}_p$.
We choose to overparametrize the network such that the total number of trainable parameters $T=\text{dim}(\theta_1) +  \dots +\text{dim}(\theta_K) $, where $\text{dim}$ refers to the total number of entries in the matrix,  satisfies $T \gg 3n$, which is the number of constraints. 

Under such overparametrized conditions, one verifies that gradient-descent converges to zero-loss in polynomial time for least-squares regression tasks \cite{du2018gradient}. 
By observing that 
$$
\min_\theta \tilde{L}(\theta) = \min_{P \in \mathcal{P}_n} \min_\theta \tilde{L}(\theta;P),\,\text{with }
$$
$$
\tilde{L}(\theta;P) = \sum_{i,j \leq n} P_{i,j} \| \phi(v_j; \theta) - x_{i} \|^2 - \lambda^{-1} H(P),
$$
and that $\tilde{L}(\theta;P)$ is convex with respect to $P$,
it follows that gradient-descent can find global minimsers of $\tilde{L}$ in polynomial time.  

As $\lambda \to \infty$, the entropic constraint disappears, 
which implies that by setting $P$ to any arbitrary permutation matrix $\Pi$, we can still obtain zero loss ($\min_\theta \tilde{L}(\theta, \Pi)=0$).
In other words, the model has enough capacity to produce any arbitrary correspondence between the points $\{v_i\}_i$ and the targets $\{x_i \}_i$ in the limit $\lambda \to \infty$. A priori, this is  an undesirable property of our model, since it would allow highly oscillatory and irregular surface reconstructions. However, our  experiments (Section \ref{sec:results}) reveal that the gradient-descent optimization of $\tilde{L}$ remarkably biases the model towards solutions with no apparent oscillation. This implicit regularisation property is reminiscent of similar phenomena enjoyed by gradient-descent in logistic regression \cite{soudry2017implicit,gunasekar2018characterizing} or matrix factorization \cite{li2017algorithmic}. In our context, gradient-descent appears to be favoring solutions with small complexity, measured for instance in terms of the Lipschitz constant of $\phi_\theta$, without the need of explicit regularization. 
We leave the theoretical analysis of such inductive bias 
for future work (Section \ref{sec:limitations}). Note that, in practice, we set $\lambda$ to a large value, which may have an extra regularizing effect. 


\subsection{Building a Global Atlas}
\label{globalsec}

Section \ref{localchart} described a procedure to obtain a local chart around a point $p \in \mathcal{S}$, with parametric domain  $V_p$ and its associated fitted parametrization $\varphi_p=\phi(\cdot;\theta_p^*): V_p \rightarrow \R^3$. In this section, we describe how to construct an atlas $\{(V_q, \varphi_q); q \in Q \}$ by appropriately selecting a set of anchor points $Q$ and by ensuring consistency between charts.

\paragraph{Consistency.}
 To define \emph{atlas consistency} more precisely, we need to separate the notions  of parametric coordinate-assignment and surface approximation, since the local chart functions $\varphi_p$  define both. 
We say that two charts $p$ and $q$ overlap, if 
$\mathcal{X}_{p,q} = \mathcal{X}_p \cap \mathcal{X}_q \neq \emptyset$.  Each discrete 
chart $\varphi_\alpha$ is equipped with a permutation $\pi_\alpha$, assigning
indices of points in $\mathcal{X}_p$ to indices of parametric positions 
in $\mathcal{V}_p$. Two overlapping charts $p$ and $q$ are \emph{consistent}
on the surface samples if 
\begin{equation*}
    \varphi_q(v_{\pi^{-1}_q(i)};\theta_q) = \varphi_p(v_{\pi^{-1}_p(i)};\theta_p),\; \forall x_i \in \mathcal{X}_p \cap \mathcal{X}_q
\end{equation*}
i.e.,  for any point in the patch overlap, the values of the two chart maps at corresponding parametric values coincide. 
If all chart maps are interpolating, then consistency is guaranteed by construction, but this is in general not the case.  We enforce consistency explicitly by minimizing a consistency loss \eqref{bibi3_phase2}. 

\paragraph{Constructing the Atlas.}
We construct the set of patch centers $Q$ using Poisson disk sampling \cite{bowers2010parallel} of $\mathcal{X}$, with a specified radius, $r$. For each 
$q \in Q$, we first extract a neighborhood $\mathcal{X}_q$ 
by intersecting a ball of radius $c r$ centered at $q$ with $\mathcal{X}$ ($\mathcal{X} \cap B(q; cr)$), where $c>1$ is another hyper-parameter. 
To reduce boundary effects, we consider a larger radius $\tilde{c}>c$ and use $\widetilde{\mathcal{X}}_q := \mathcal{X} \cap B(q; \tilde{c}r)$ to fit the local chart for $\mathcal{X}_q$. In general, the intersection of
$\mathcal{S}$ with the ball $B(q; \tilde{c}r)$ consists of multiple 
connected components, possibly of nontrivial genus. We use the heuristic described below
to filter out points we expect to be on a different sheet from the 
ball center $q$.

To ensure consistency as defined above, we fit the charts in two phases. In the first phase, we locally fit each chart to its associated points. In the second phase, we compute a joint fitting of all pairs of overlapping charts. 

Let $(\theta_p, \pi_p)$ and $(\theta_q, \pi_q)$ denote the parameters and permutations of the patches $p$ and $q$ respectively at some iteration of the optimization. We compute the first local fitting phase as
\begin{eqnarray}
\label{bibi3_phase1}
    \min_{\theta_p} \, \inf_{\pi_p} \sum_{i \leq |\mathcal{X}_{p}|} \| \phi(v_i; \theta_p) - x_{\pi_p(i)}\|^2.
\end{eqnarray}

We define the set of indices of parametric points in chart $p$  of the intersection $\mathcal{X}_{p,q}$ as
$$
\mathcal{T}_{pq}=\{i| x_{\pi_p(i)} \in \mathcal{X}_{p,q} \},
$$
where $\mathcal{T}_{qp}$ is the corresponding set in chart $q$.
The map between indices of corresponding  parametric points in two patches is given by:
$\pi_{p \to q}:= \pi_{q} \circ \pi_p^{-1}:\mathcal{T}_{pq} \to \mathcal{T}_{qp}$.

Equipped with this correspondence, we compute the second joint fitting phase between all patch pairs as:
\begin{eqnarray}
\label{bibi3_phase2}
    \min_{\theta_p, \theta_{q}} \, \inf_{\pi_p, \pi_{q}} \sum_{i \in \mathcal{T}_{p;q}} \| \phi(v_i; \theta_p) - \phi(v_{\pi_{p \to q}(i)};\theta_{q})\|^2.
\end{eqnarray}
Observe that by the triangle inequality, 
\begin{equation*}
\begin{split}
\| \phi(v_i; \theta_p) - \phi(v_{\pi_{p \to q}(i)};\theta_{q})\| \leq& \| \phi(v_i; \theta_p) - x_{\pi_p(i)} \| + \\
& \| \phi(v_{\pi_{p \to q}(i)}; \theta_{q}) - x_{\pi_{p}(i)}\|.   
\end{split}
\end{equation*}
Therefore, the joint fitting term \eqref{bibi3_phase2} is bounded by the sum of two separate fitting terms \eqref{bibi3_phase1} for each patch (note that $x_{\pi_{p}(i)} = x_{\pi_{q}(\pi_{p \to q}(i))}$). 
Consistent transitions are thus enforced by the original Wasserstein loss if the charts are \emph{interpolating}, i.e. $\phi(v_i;\theta_p) = x_{\pi_p(i)}$ for all $i,p$. 
However, in presence of noisy point clouds, the joint fitting phase enables a smooth transition between local charts without requiring an exact fit through the noisy samples.

If the Sinkhorn distance is used instead of the EMD, then we project the stochastic matrices $P_q$, $P_{q'}$ to the nearest permutation matrix by setting to one the maximum entry in each row. 

\paragraph{Filtering Sample Sets $\mathcal{X}_r$.}
In our experiments, we choose the ball radius to be sufficiently 
small to avoid most of the artifacts related to fitting patches 
to separate sheets of the surface-intersection with $B(q;cr)$.
The radius can be easily chosen adaptively, although at a significant 
computational cost, by replacing a ball by several smaller balls 
whenever the quality of the fit is bad. Instead, we use a cheaper heuristic 
to eliminate points from each set that are likely to be on a 
different sheet: We assume that the input point cloud $\mathcal{X}$ also contains normals. If normals are unavailable, they can be easily estimated using standard local fitting techniques.  We then discard all vertices in each $\mathcal{X}_p$ whose normals form an angle greater than a fixed threshold $\alpha$ with respect to the normal at the center. In all our experiments (Section \ref{sec:results}) we used $r=2.5\%$ of the bounding box diagonal enclosing the point cloud, $c=1.5$ $\tilde{c}=1.5$, $\alpha=100$ degrees.

\begin{figure}
    \centering\footnotesize
\parbox{.85\linewidth}{
    \includegraphics[width=.45\linewidth]{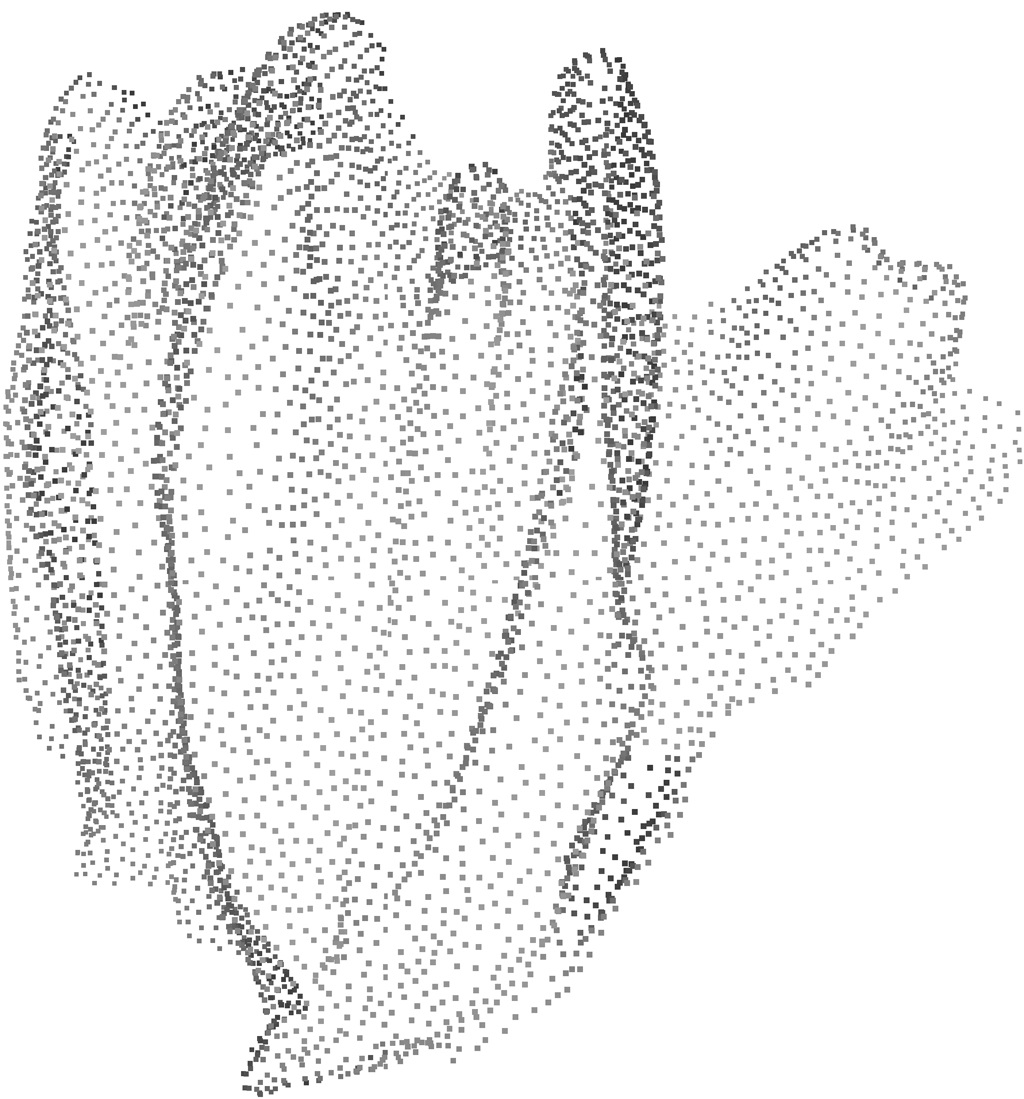}\hfill
    \includegraphics[width=.45\linewidth]{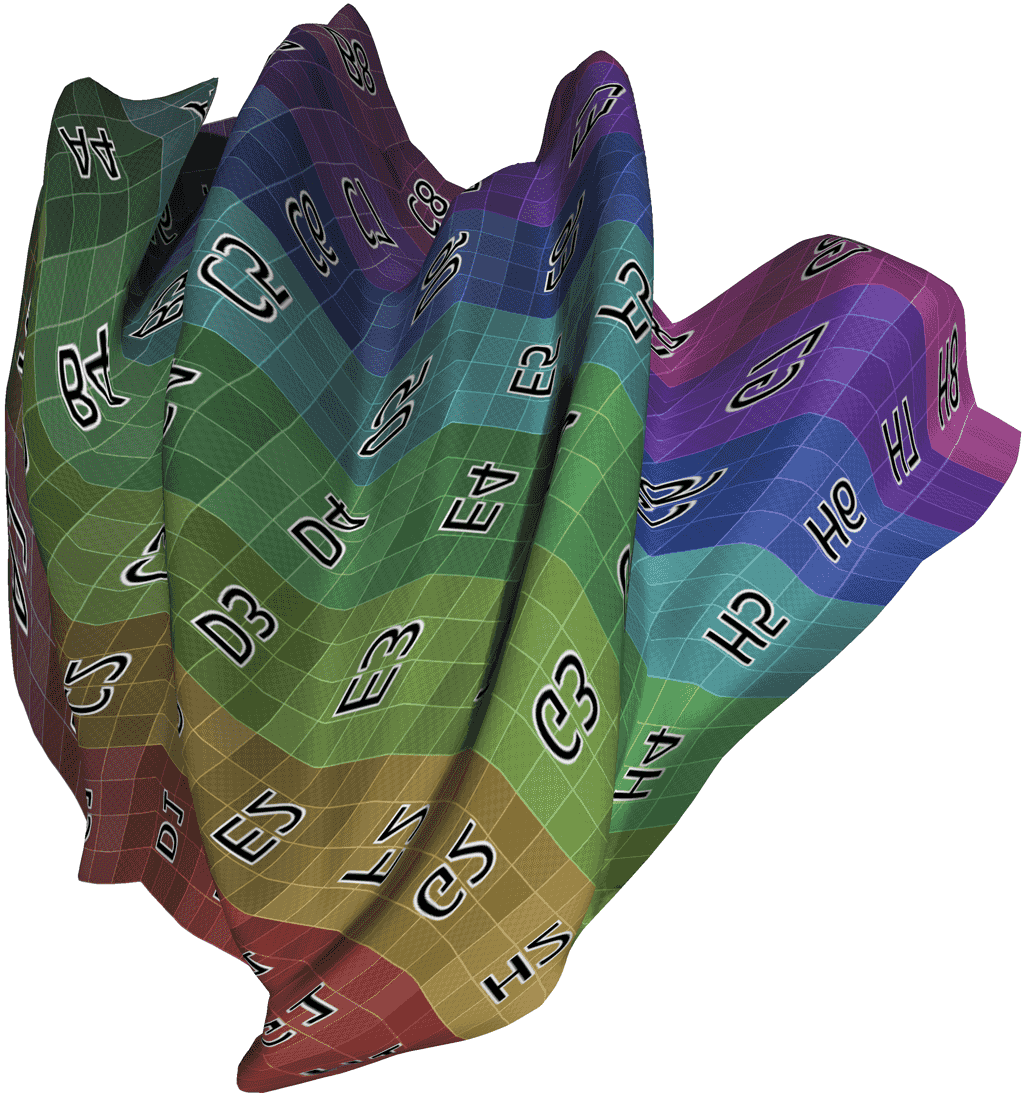}\par
    \parbox{.45\linewidth}{\centering Input}\hfill
    \parbox{.45\linewidth}{\centering Fitted surface}
}\\[2ex]
    \caption{Single patch fitting with uv-mapping illustrated with a checkerboard texture.}
    \label{fig:lion-mane}
\end{figure}

\begin{figure}
    \centering\footnotesize
    \includegraphics[width=.31\linewidth]{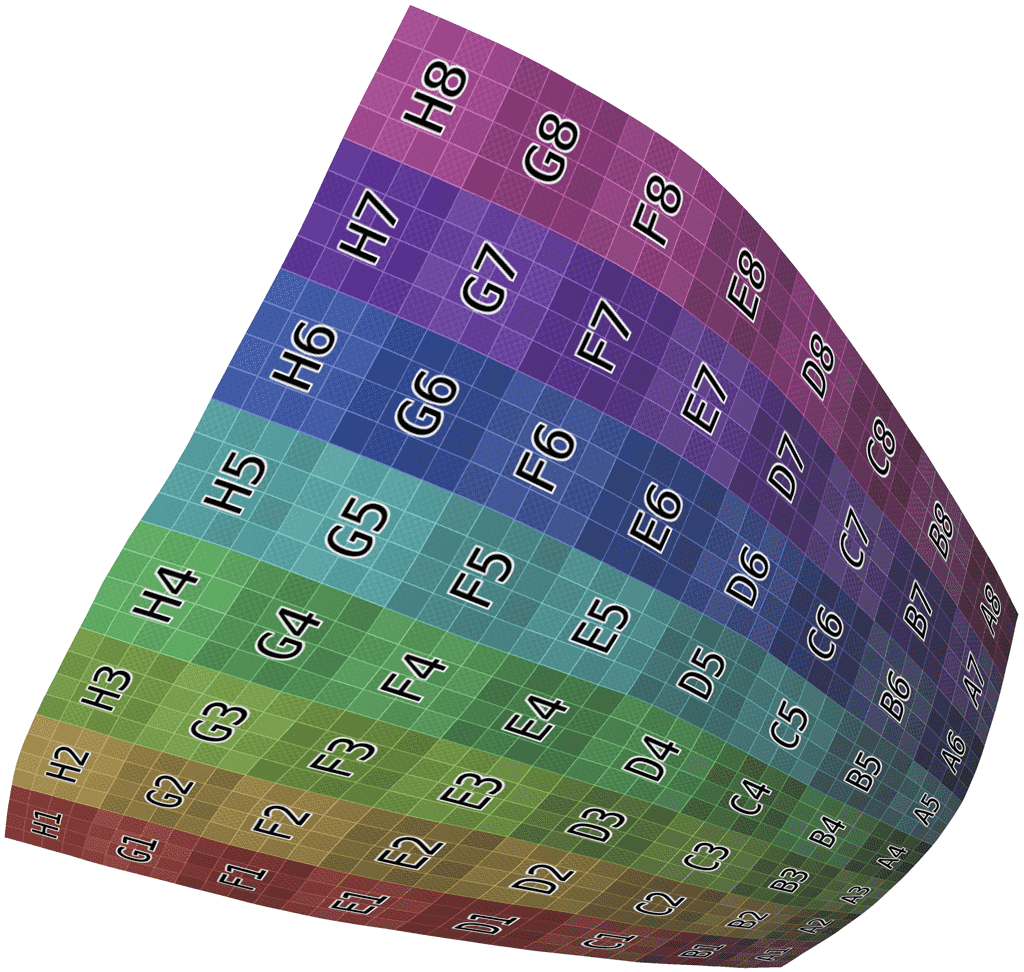}\hfill
    \includegraphics[width=.31\linewidth]{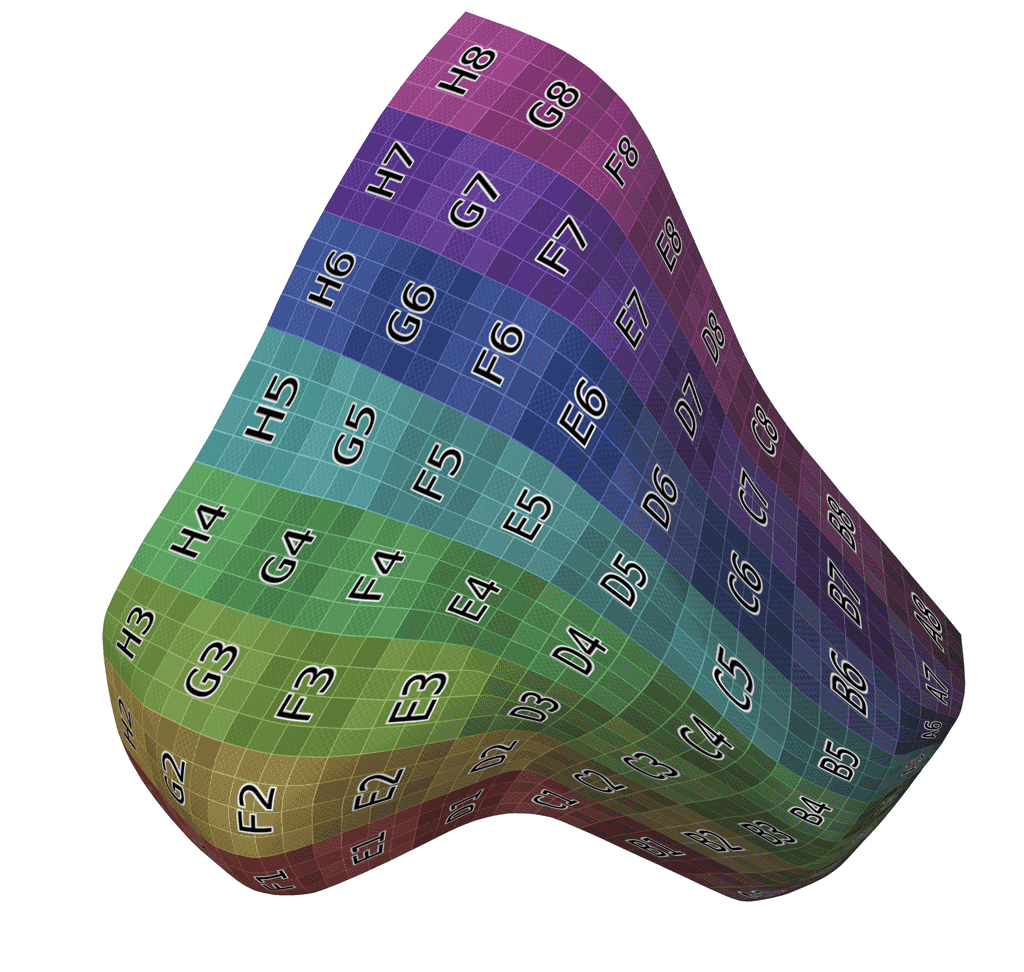}\hfill
    \includegraphics[width=.31\linewidth]{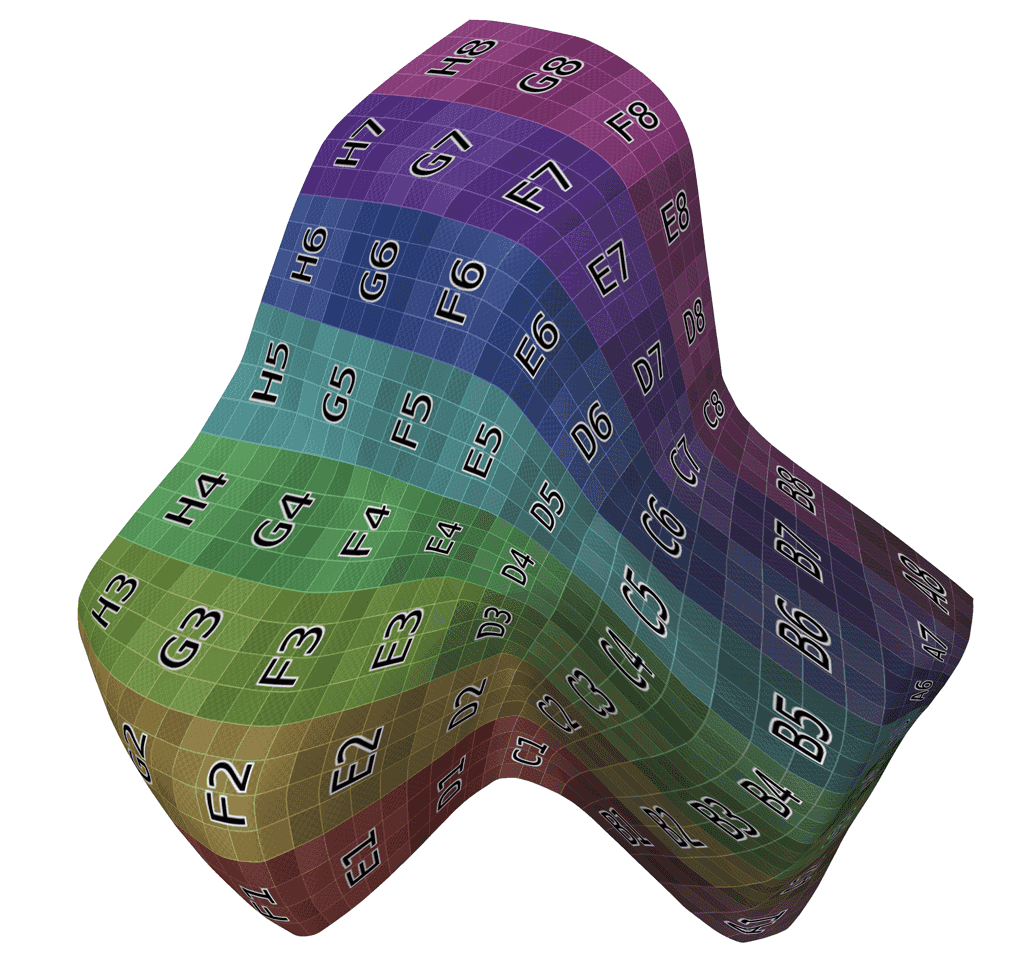}\\
    \parbox{.31\linewidth}{\centering 25}\hfill
    \parbox{.31\linewidth}{\centering 75}\hfill
    \parbox{.31\linewidth}{\centering 100}\\
    \includegraphics[width=.31\linewidth]{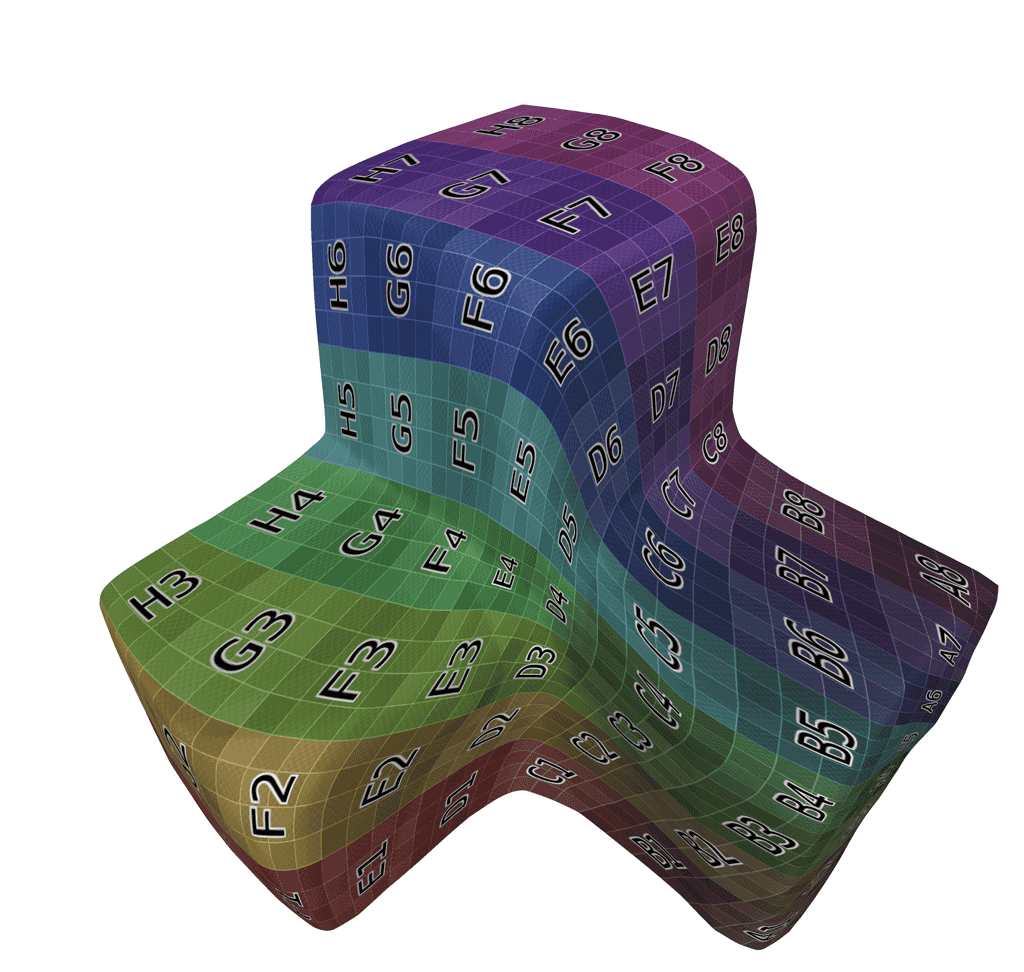}\hfill
    \includegraphics[width=.31\linewidth]{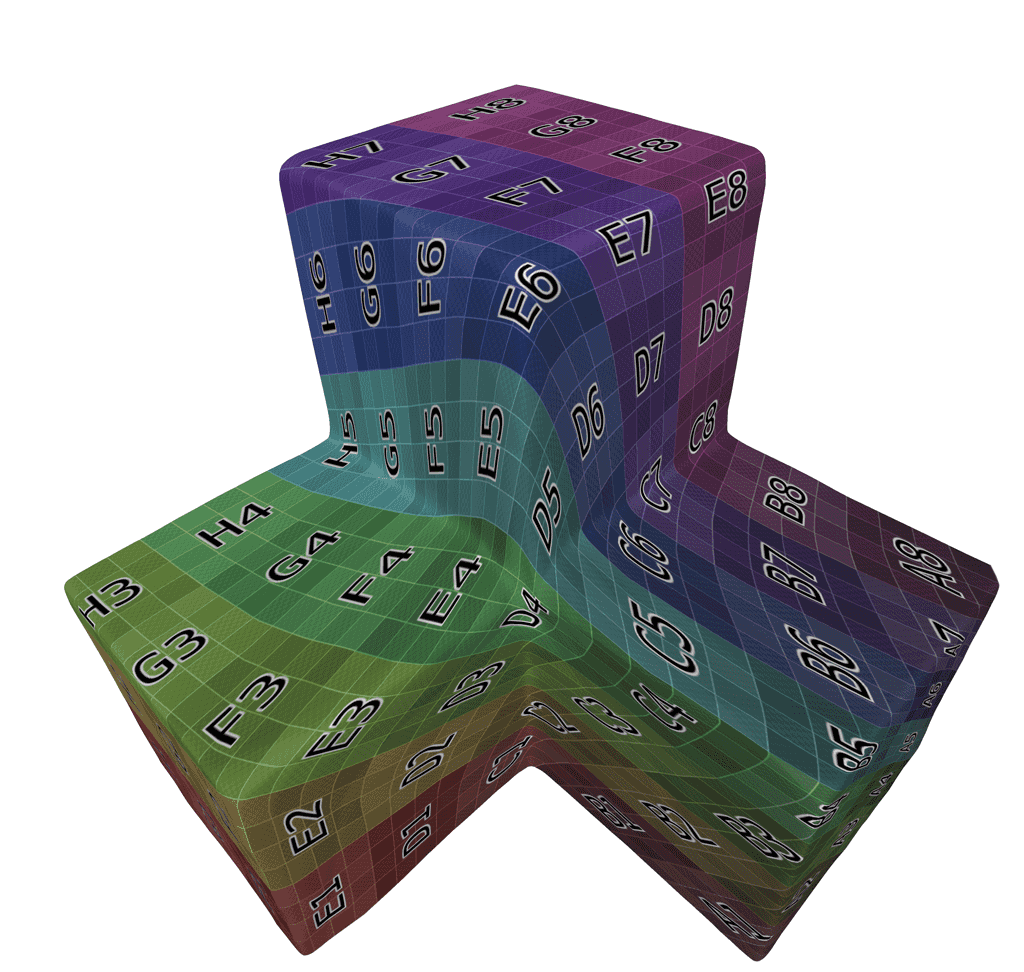}\hfill
    \includegraphics[width=.31\linewidth]{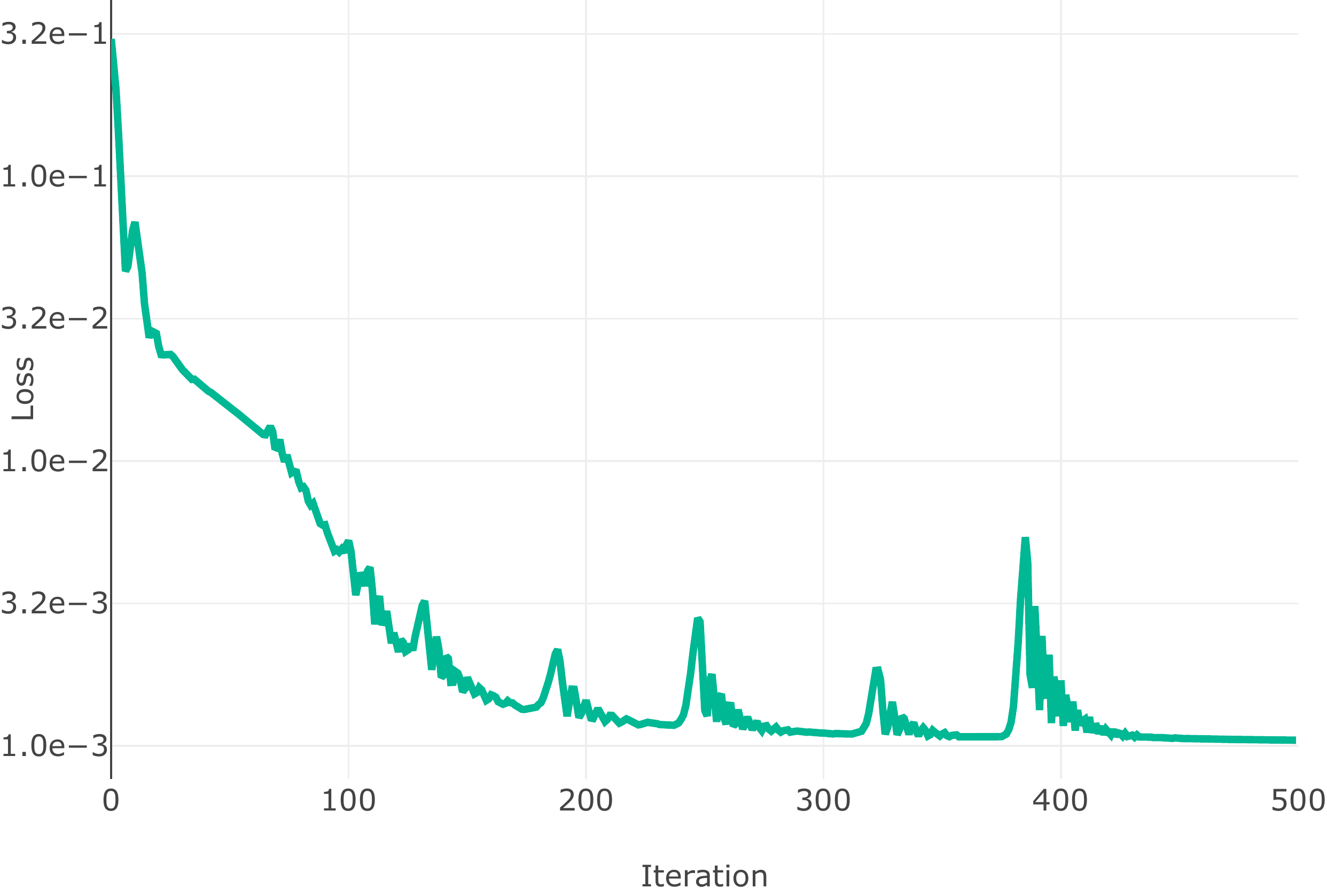}\\
    \parbox{.31\linewidth}{\centering 150}\hfill
    \parbox{.31\linewidth}{\centering 500}\hfill
    \parbox{.31\linewidth}{\centering Convergence}\\[2ex]
    \caption{Evolution of single-patch reconstruction as the loss $\tilde{L}(\theta)$ is minimised. The jumps in the loss are a side-effect of the adaptive gradient descent scheme ADAM.}
    \label{fig:lion-evolution}
\end{figure}

\section{Experiments}
\label{sec:results}

\begin{figure*}\centering\footnotesize
\includegraphics[width=.19\linewidth]{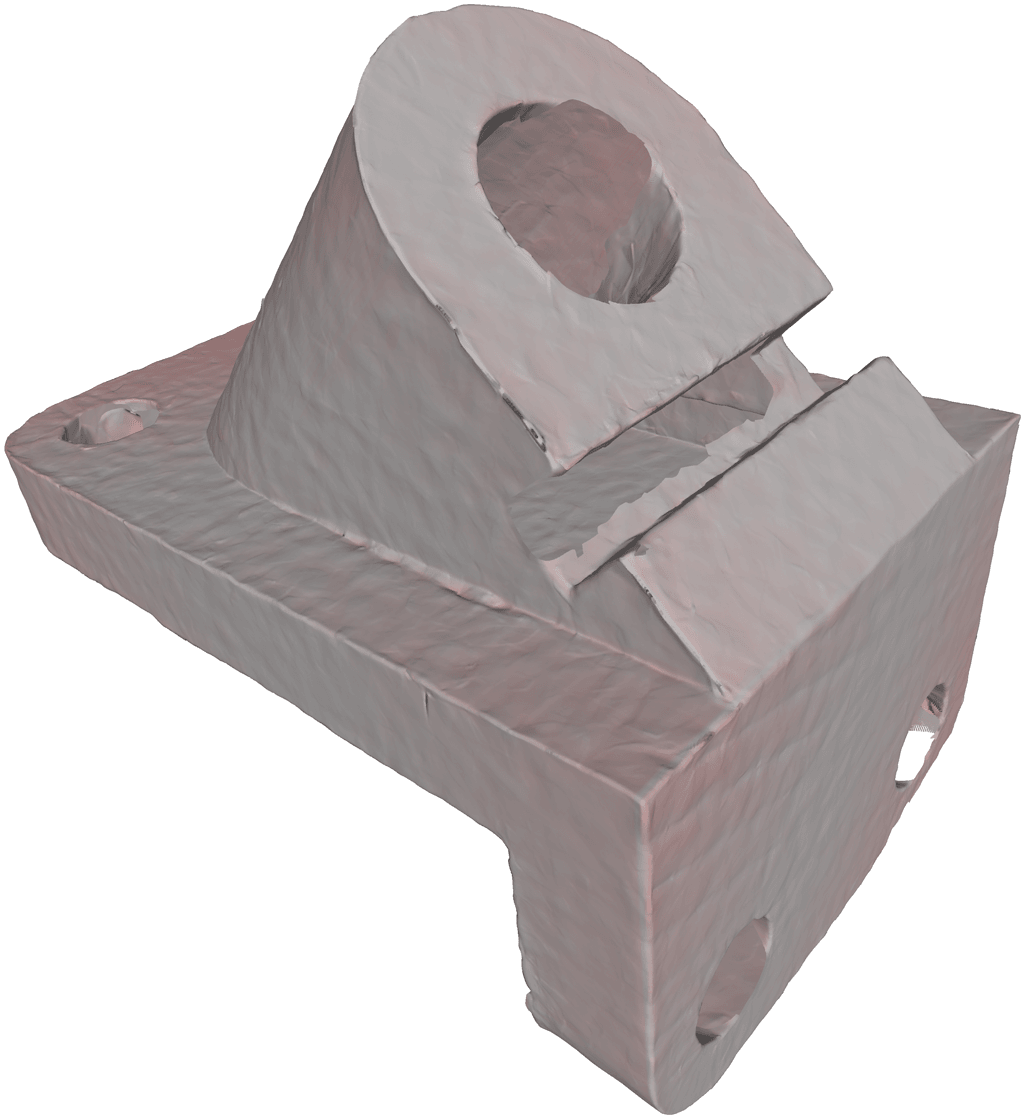}\hfill
\includegraphics[width=.22\linewidth]{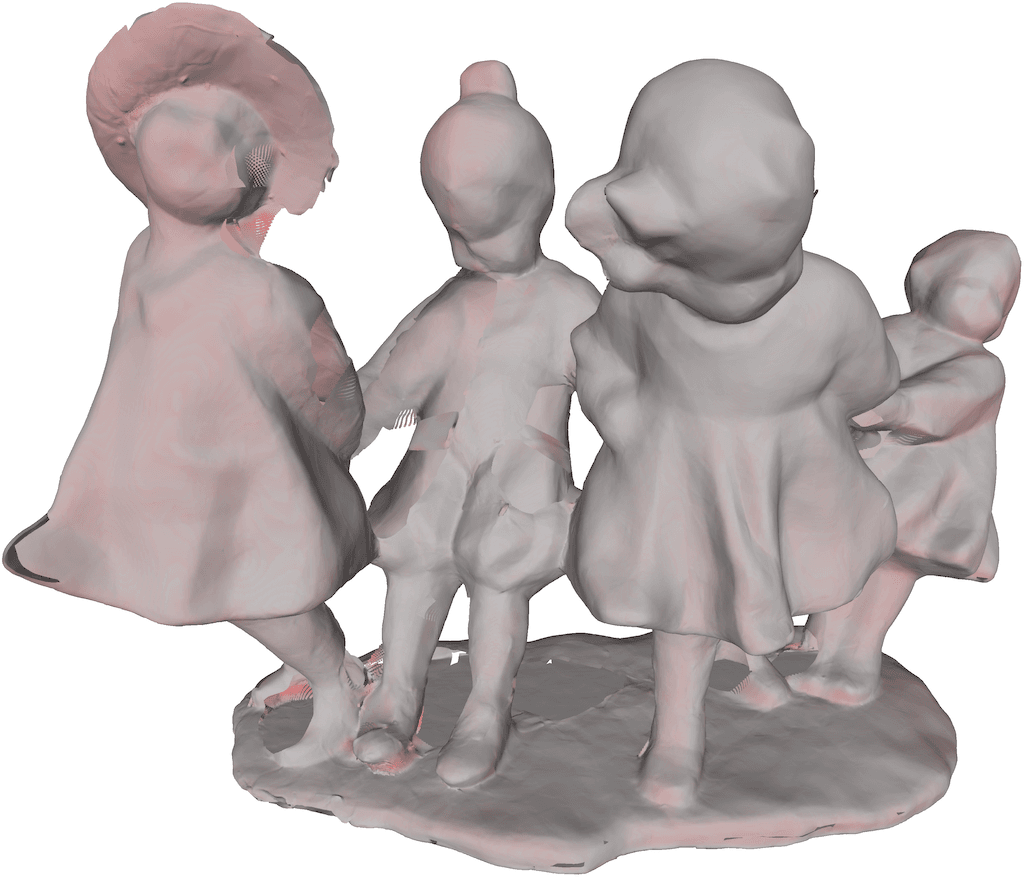}\hfill
\includegraphics[width=.21\linewidth]{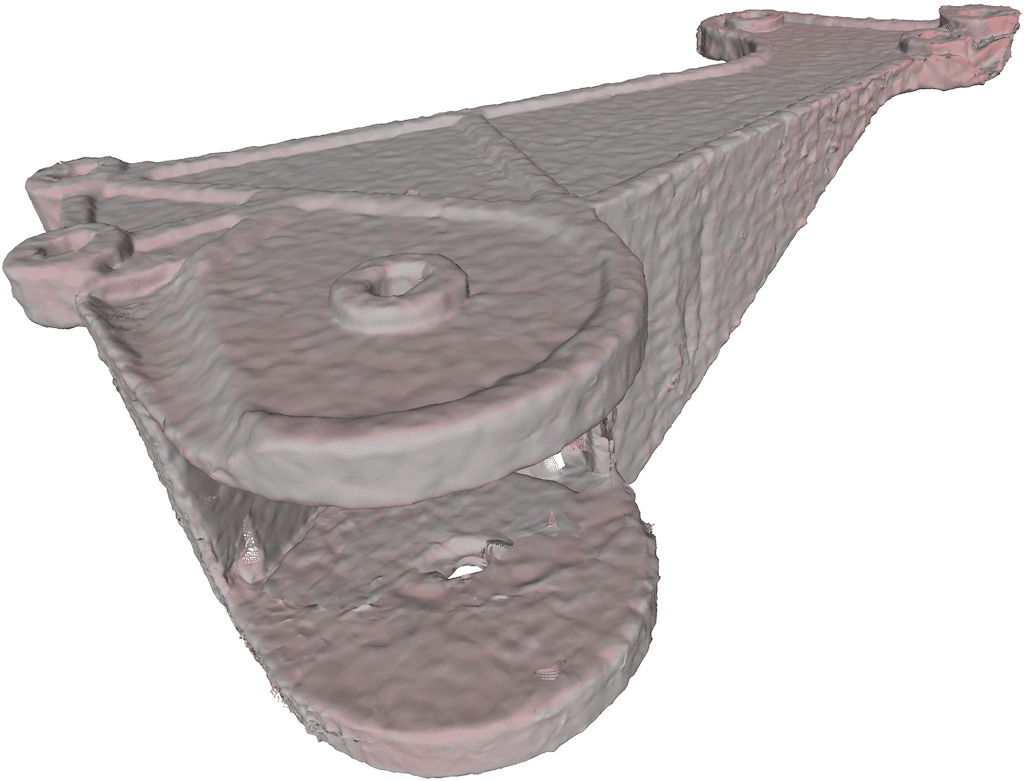}\hfill
\includegraphics[width=.17\linewidth]{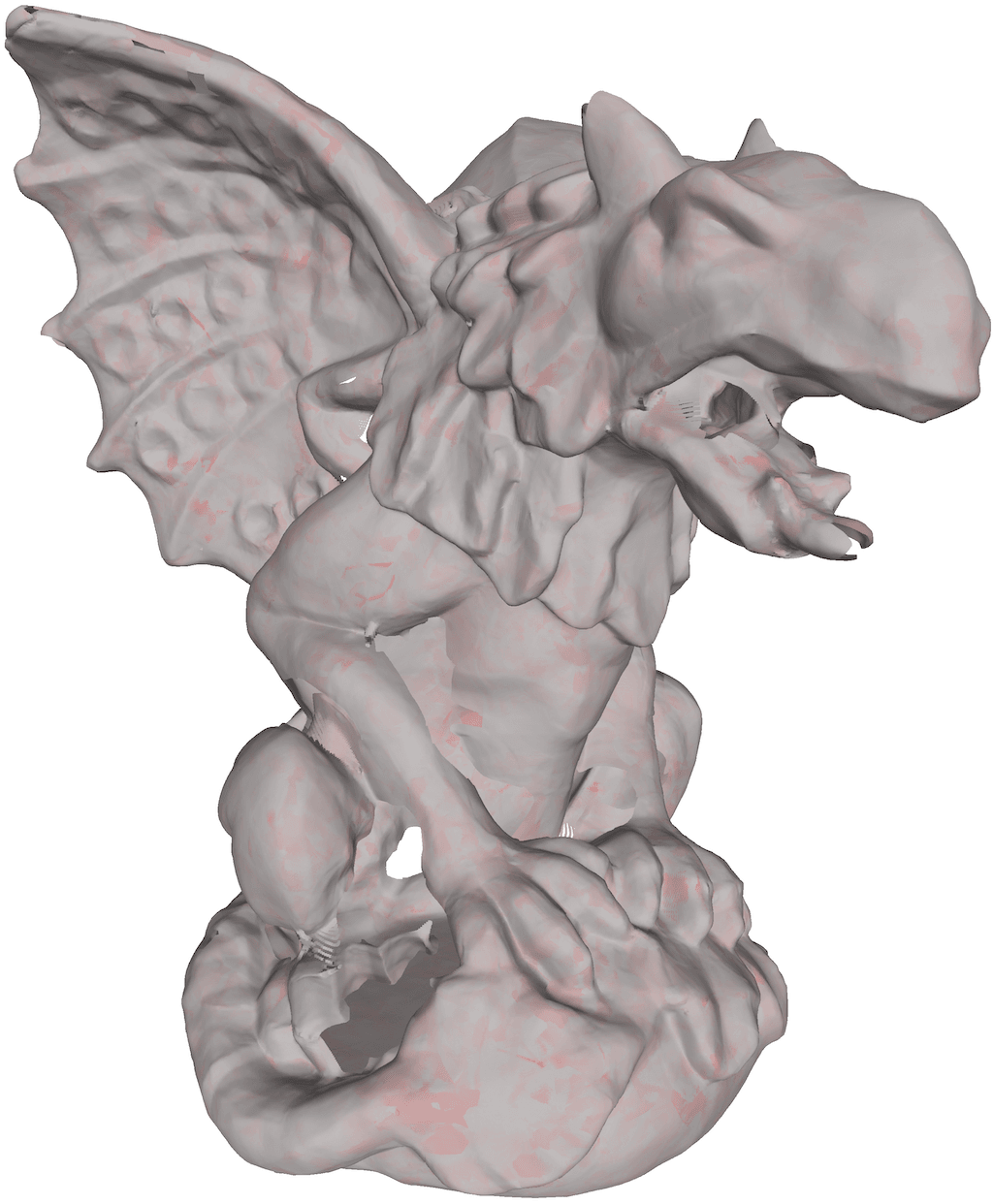}\hfill
\includegraphics[width=.12\linewidth]{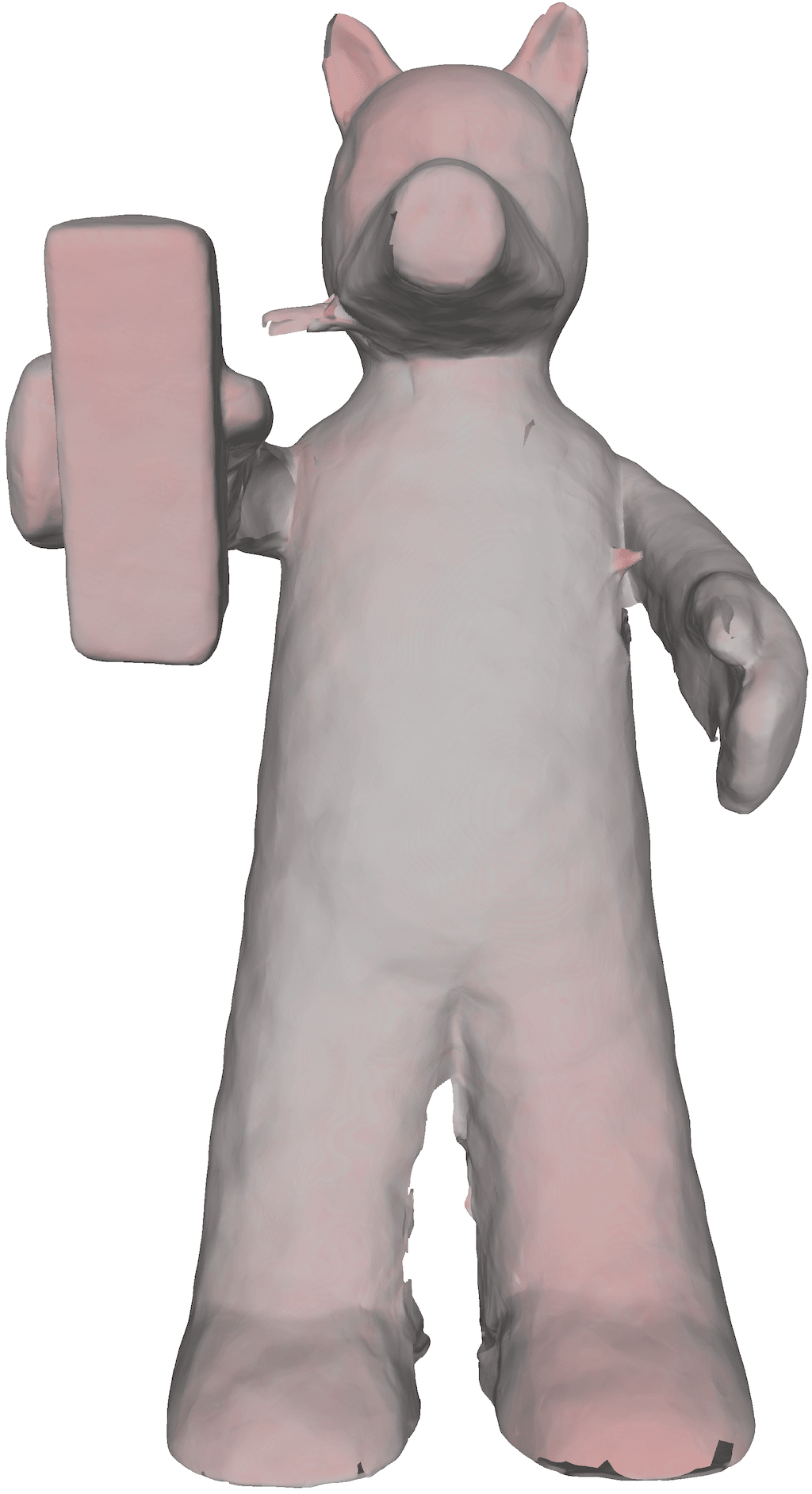}\hfill\hfill
\includegraphics[width=.07\linewidth]{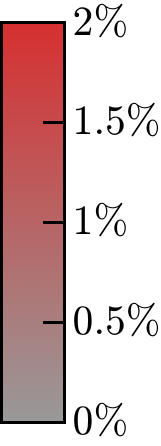}\par
\parbox{.19\linewidth}{\centering Anchor}\hfill
\parbox{.22\linewidth}{\centering Dancing Children}\hfill
\parbox{.21\linewidth}{\centering Daratech}\hfill
\parbox{.17\linewidth}{\centering Gargoyle}\hfill
\parbox{.12\linewidth}{\centering Lord Quas}\hfill\hfill
\parbox{.07\linewidth}{~}\\[2ex]
    \caption{Fitted points on the different models of the benchmark, the color illustrates the error with respect to the ground truth.}
    \label{fig:srb-modesl}
\end{figure*}

\paragraph{Experimental Setup.}
We run our experiments on a computing node with an Intel(R) Xeon(R) CPU E5-2690 v4, 256 GBgb of memory, and 4 NVIDIA Tesla P40 GPUs. The runtime of our algorithm are considerably higher than competing methods, requiring around 0.45 minutes per patch, for a total of up to 6.5 hours to optimize an the entire model. 

We optimize the losses \eqref{bibi3_phase1} and \eqref{bibi3_phase2} using the ADAM \cite{kingma2014adam} implementation in PyTorch with default parameters. Specifically for ADAM, we use a learning rate of $10^{-3}$, $\beta = (0.9, 0.999)$, $\epsilon=10^{-8}$, and no weight decay. For the Sinkhorn loss, we use a regularization parameter, $\lambda = 1000$. For the networks, $\phi$, we use an MLP with fully connected layer sizes: (2, 128, 256, 512, 512, 3) and ReLU activations. Our reference implementation is available at {\small \url{https://github.com/fwilliams/deep-geometric-prior}}. 

\paragraph{Single Patch Fitting.}
Our first experiment shows the behaviour of a single-patch network overfitted on a complex point cloud (Figure \ref{fig:lion-mane} left). Our result is a tight fit to the point cloud.  An important side effect of our construction is an explicit local surface parametrization, which can be used to compute surface curvature, normals, or to apply an image onto the surface as a texture map (Figure \ref{fig:lion-mane} right).

Figure \ref{fig:lion-evolution} shows the evolution of the fitting and of the parameterisation $\varphi_\theta$ 
as the optimization of $\tilde{L}(\theta)$ progresses. We observe that the optimization path follows a trajectory where $\phi_\theta$ does not exhibit distortions, supporting the hypothesis that gradient descent biases towards solutions with low complexity.

\begin{figure}\centering\footnotesize
\parbox{0.8\linewidth}{
    \includegraphics[width=.45\linewidth]{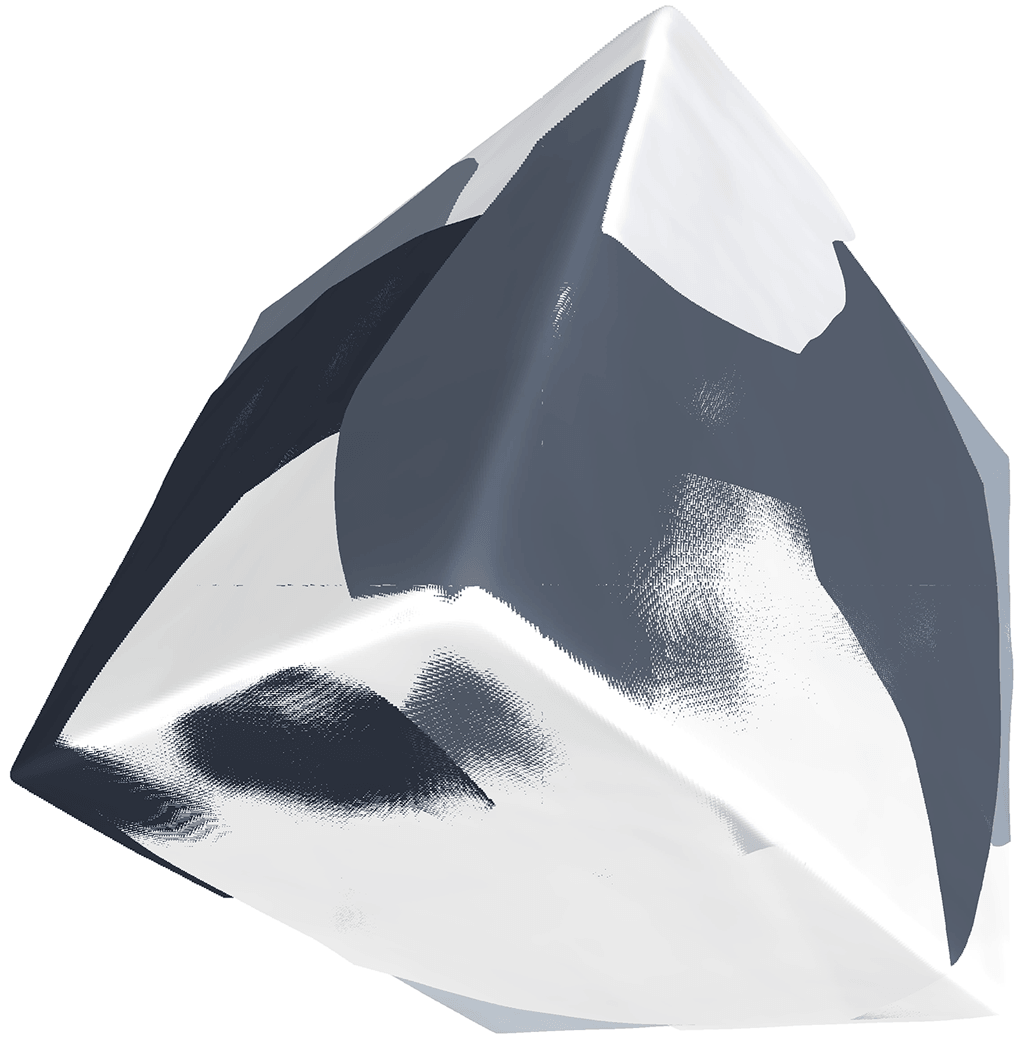}\hfill
    \includegraphics[width=.45\linewidth]{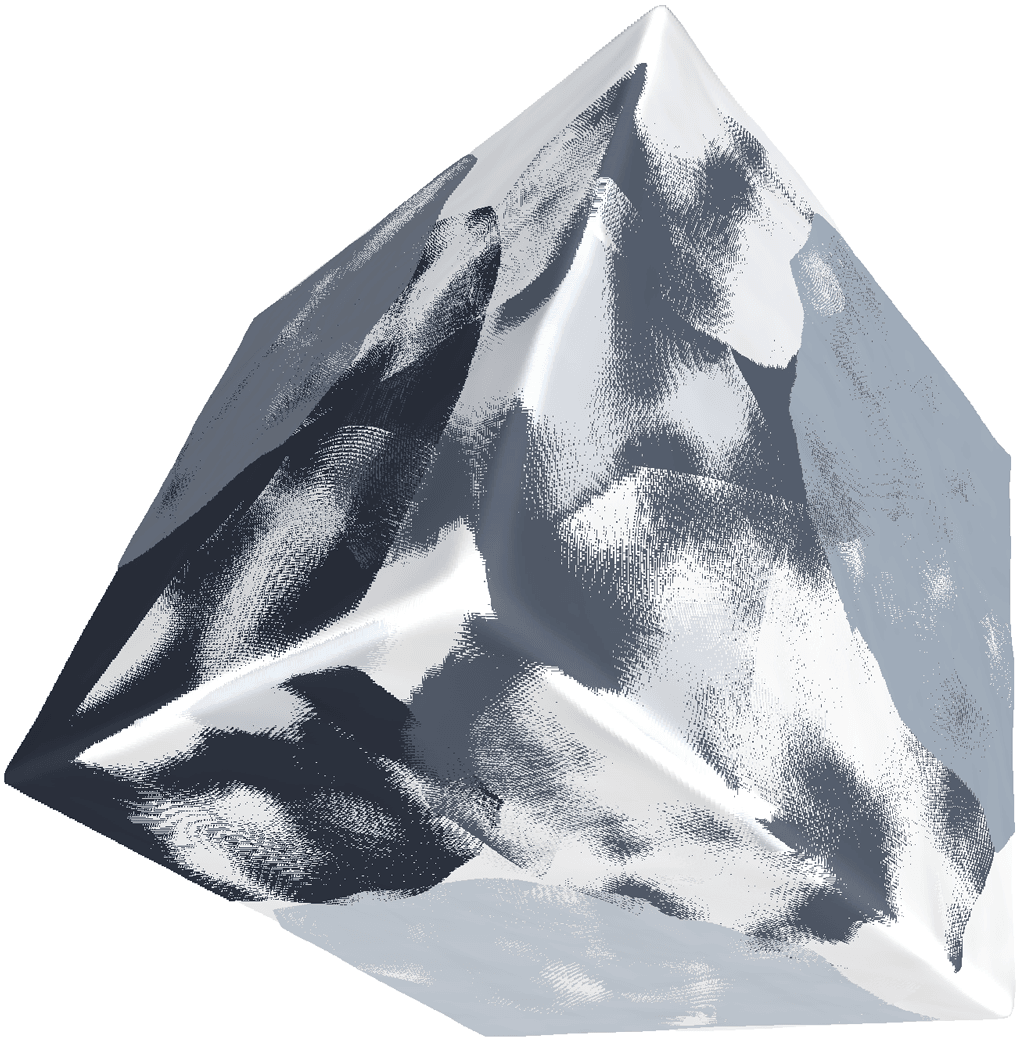}\par
    \parbox{.45\linewidth}{\centering First patches}\hfill
    \parbox{.45\linewidth}{\centering Final patches}\par
    }\\[2ex]
    \caption{Initial patch fitting (left), we clearly see the ``un-mixed'' colors due to the disagreeing patches. After optimization (right) all patches are overlapping producing a mix of color.}
    \label{fig:overlapping}
\end{figure}

\paragraph{Global Consistency.}
As described in Section \ref{globalsec}, reconstructing an entire surface from local charts requires 
consistent transitions, leading to the formulation in~\eqref{bibi3_phase1} and \eqref{bibi3_phase2} which reinforces consistency across 
overlapping patches. Figure \ref{fig:overlapping} illustrates the effect of adding the extra consistency 
terms. We verify that these terms significantly improve the consistency. 

\begin{figure}\centering\footnotesize
    \includegraphics[width=\linewidth]{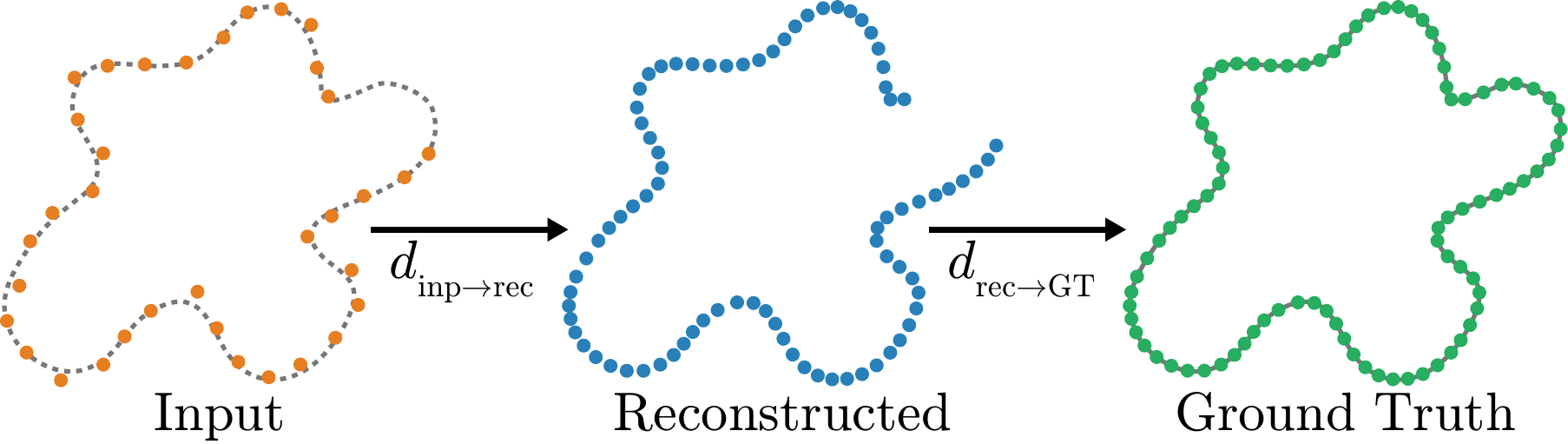}\\[2ex]
    \caption{Diagram illustrating the evaluation measures we use to compare different reconstructions.}
    \label{fig:schema_distances}
\end{figure}

\paragraph{Surface Reconstruction Benchmark.}
To evaluate quantitatively the performance of our complete reconstruction pipeline, we use the setup proposed by \cite{BergerLNTS13}, using the first set of range scans for each of the 5 benchmark models. 
Figure~\ref{fig:srb-modesl} shows the results (and percentage error) of our method on the five models of the benchmark.
We compare our results against the baseline methods described in Section \ref{sec:related},
and use the following metrics to evaluate the quality of the reconstruction: Let $\mathcal{X}$ denote the input point cloud. From the ground-truth surface $\mathcal{S}$ and the reconstruction $\widehat{\mathcal{S}}$, we obtain two dense point clouds that we denote respectively by $\mathcal{Y}$ and $\widehat{\mathcal{Y}}$. We consider 
\begin{eqnarray}
\label{metrics}
    d_{\mathrm{inp} \to \mathrm{rec}}(i) =& \min_{\widehat{y}_j \in \widehat{\mathcal{Y}}} \| x_i - \widehat{y}_j \|~,i \leq |\mathcal{X}|, \nonumber \\
    d_{\mathrm{rec} \to \mathrm{GT}}(j) =& \min_{{y}_k \in {\mathcal{Y}}} \| \widehat{y}_j - {y}_k \|~,j \leq |\widehat{\mathcal{Y}}|.
\end{eqnarray}
That is, $d_{\mathrm{rec} \to \mathrm{GT}}$ measures a notion of precision of the reconstruction, while $d_{\mathrm{inp} \to \mathrm{rec}}$ measures a notion of recall. Whereas $d_{\mathrm{rec} \to \mathrm{GT}}$ is an indication of overall quality, it does not penalize the methods for not covering undersampled regions of the input. 
Figure \ref{fig:schema_distances} illustrates these one-sided correspondence measures. 
A naive reconstruction that copies the input satisfies $d_{\mathrm{inp} \to \mathrm{rec}} \equiv 0$ but 
since in general the input point cloud consists of noisy measurements, we will have $d_{\mathrm{inp} \to \mathrm{GT}} >0$. 

\begin{figure}\centering\footnotesize
    \includegraphics[width=.49\linewidth]{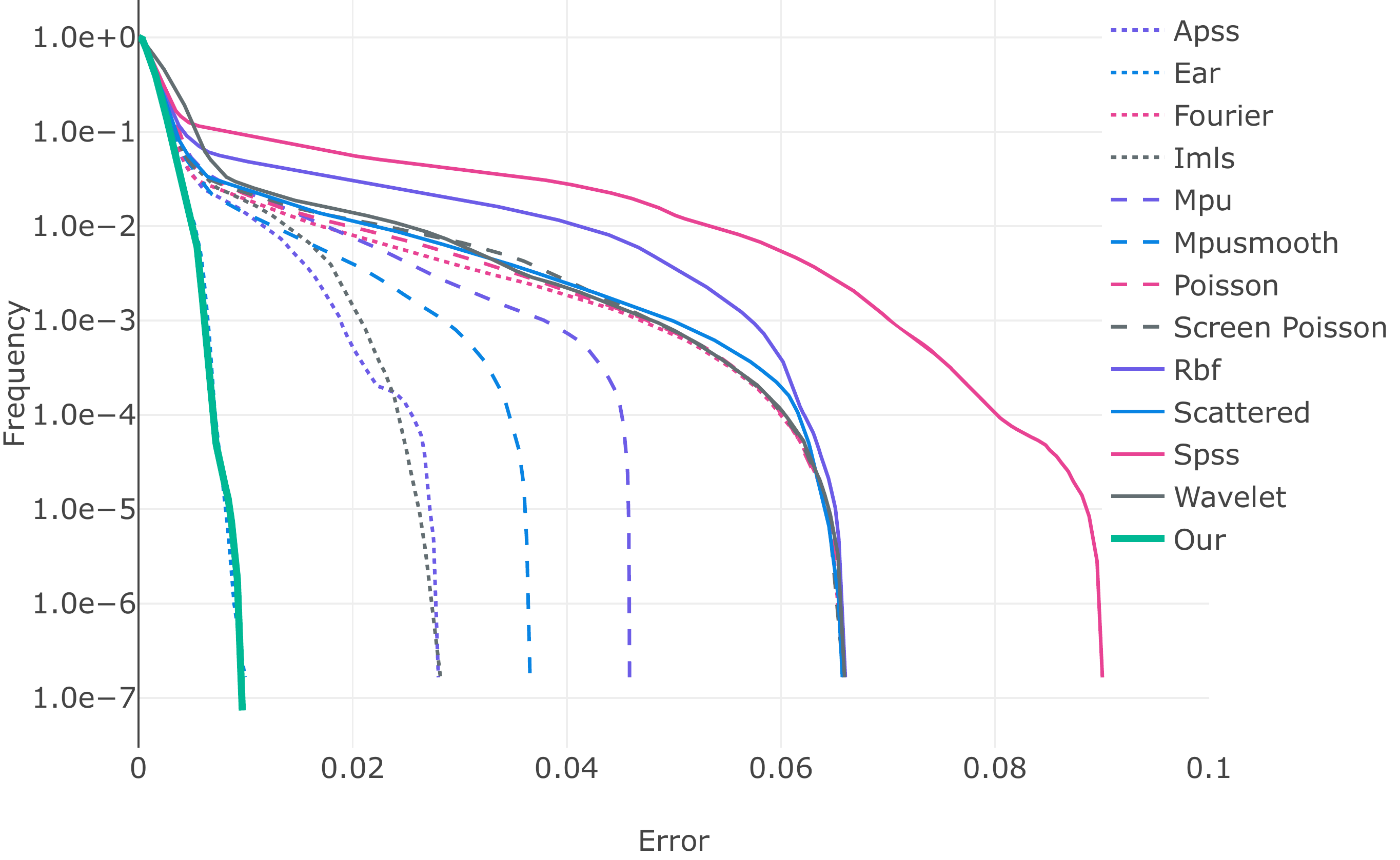}\hfill
    \includegraphics[width=.49\linewidth]{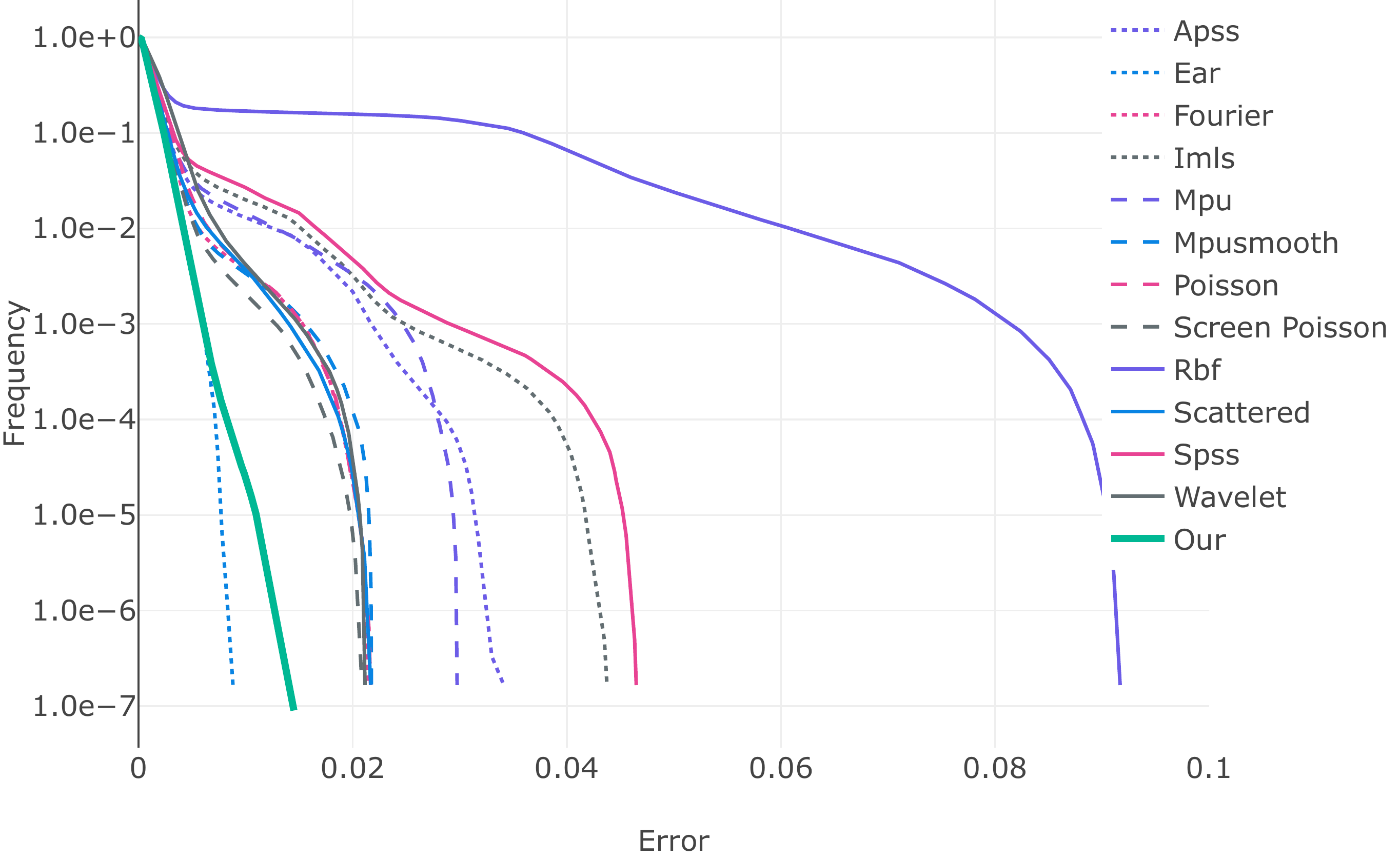}\par
    \parbox{.49\linewidth}{\centering Anchor}\hfill
    \parbox{.49\linewidth}{\centering Gargoyle}\\[2ex]
    \caption{Percentage of fitted vertices ($y$-axis, log scale) to reach a given error ($d_{\mathrm{inp} \to \mathrm{rec}}$, $x$-axis) for different methods. The errors are computed from the fitted surface to the ground truth. The plots for the remaining models of the dataset are provided in the supplementary document. }
    \label{fig:first-result}
\end{figure}

\begin{figure}\centering\footnotesize
    \includegraphics[width=.49\linewidth]{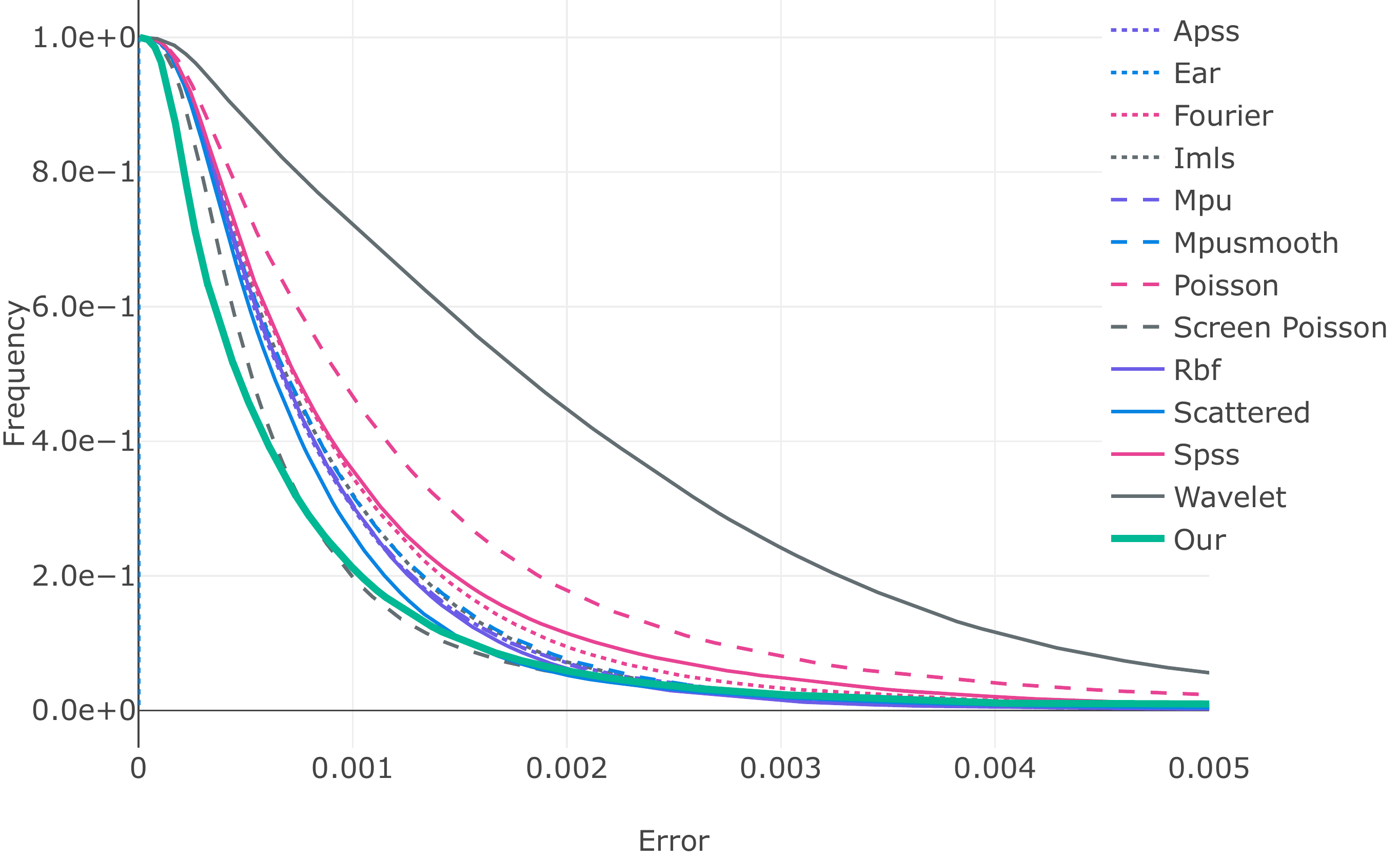}\hfill
    \includegraphics[width=.49\linewidth]{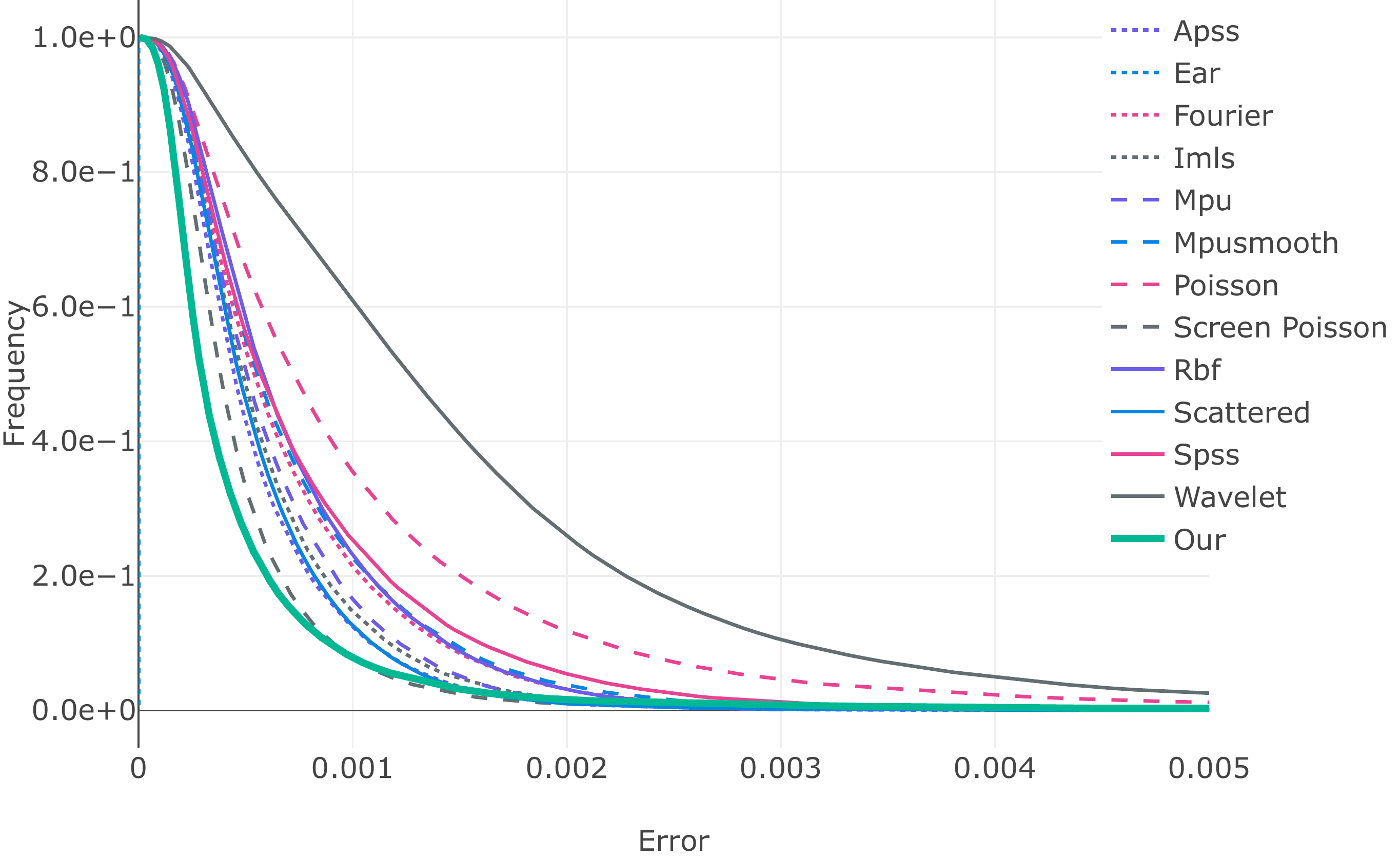}\par
    \parbox{.49\linewidth}{\centering Anchor}\hfill
    \parbox{.49\linewidth}{\centering Gargoyle}\\[2ex]
    \caption{Percentage of fitted vertices ($y$-axis) to reach a given error ($d_{\mathrm{rec} \to \mathrm{GT}}$), $x$-axis) for different methods. The errors are measured as distance from the input data to the fitted surface. The plots for the remaining models of the dataset are provided in the supplementary document.}
    \label{fig:second-result}
\end{figure}

Figures \ref{fig:first-result} and \ref{fig:second-result} show respectively the percentage of vertices of $\widehat{\mathcal{Y}}$ and $\mathcal{X}$ such that $d_{\mathrm{rec} \to \mathrm{GT}}$ and $d_{\mathrm{inp} \to \mathrm{rec}}$ is below a given error. 

\begin{figure}
    \centering\footnotesize
    \includegraphics[width=0.5\linewidth]{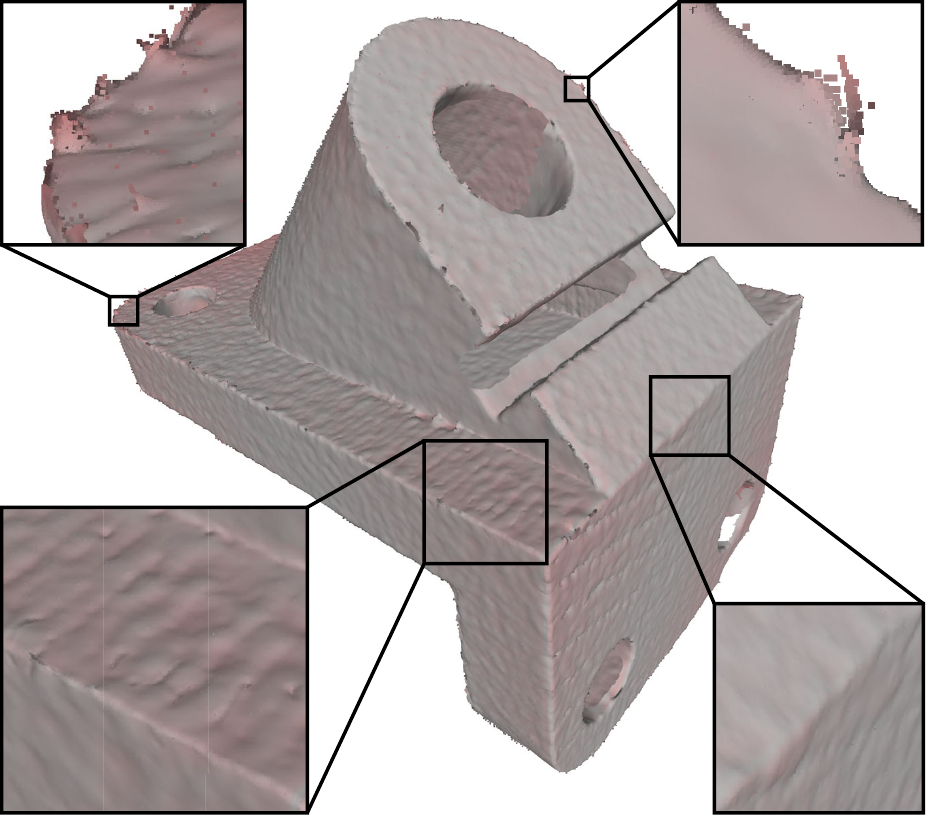}\hfill
    \includegraphics[width=0.5\linewidth]{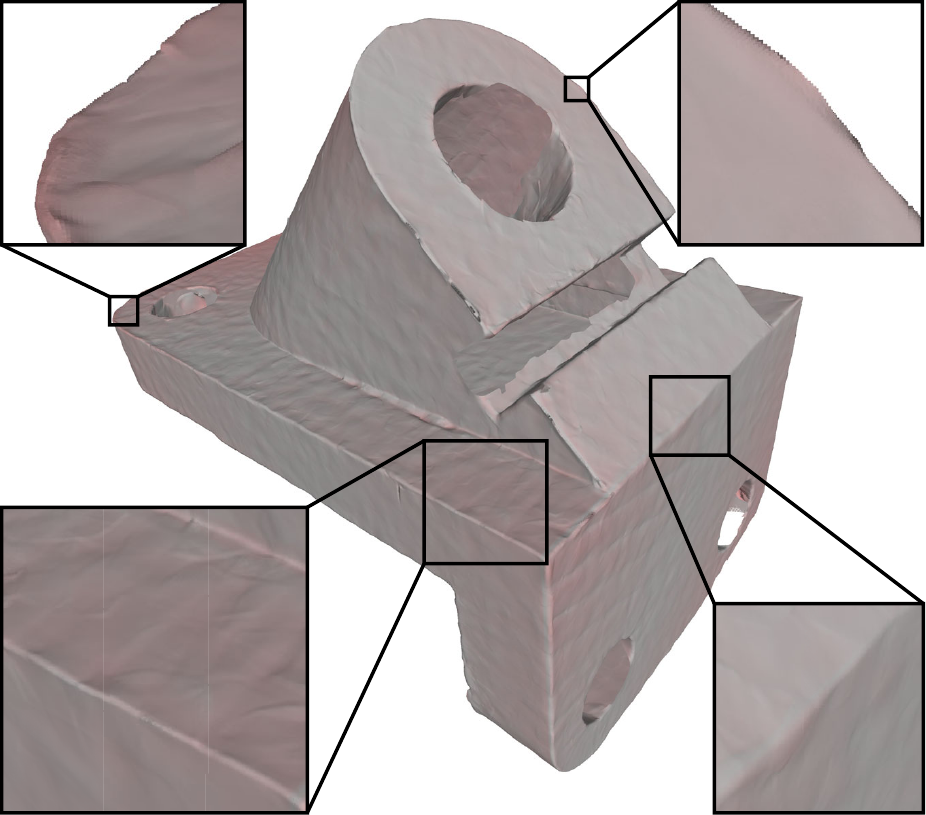}\par
    \parbox{0.5\linewidth}{\centering EAR}\hfill
    \parbox{0.5\linewidth}{\centering Ours}\\[2ex]
    \caption{EAR (left) versus Ours (right). The input point cloud contains noise which is smoothed out by our reconstruction (while preserving sharp features quite well) but interpolated by EAR. The result is that EAR produces spurious points and visible artifacts.}
    \label{fig:our-vs-ear}
\end{figure}
    
Our technique outperforms all the technique we tested, and it is on par with the state-of-the art EAR method \cite{Huang:2014}, which achieves a similar score for these 5 models. But, as we will discuss in the next paragraph, EAR is unable to cope with noisy inputs. This is a remarkable result considering that our method, differently from all the others, produces an explicit parametrization of the surface, which can be resampled at arbitrary resolutions, used for texture mapping, or to compute analytic normals and curvature values.

EAR is an interpolative method, which by construction satisfies $\mathcal{X} \subset \widehat{\mathcal{Y}}$. 
It follows that the noise in the measurements is necessarily transferred to the surface reconstruction. Figure \ref{fig:our-vs-ear} illustrates that our deep geometric prior preserves the sharp structures while successfully denoising the input point cloud. The mathematical analysis of such implicit regularisation is a fascinating direction of future work. 

\begin{figure}\centering\footnotesize
    Sharp Features\\[1.5ex]
    \includegraphics[width=.22\linewidth]{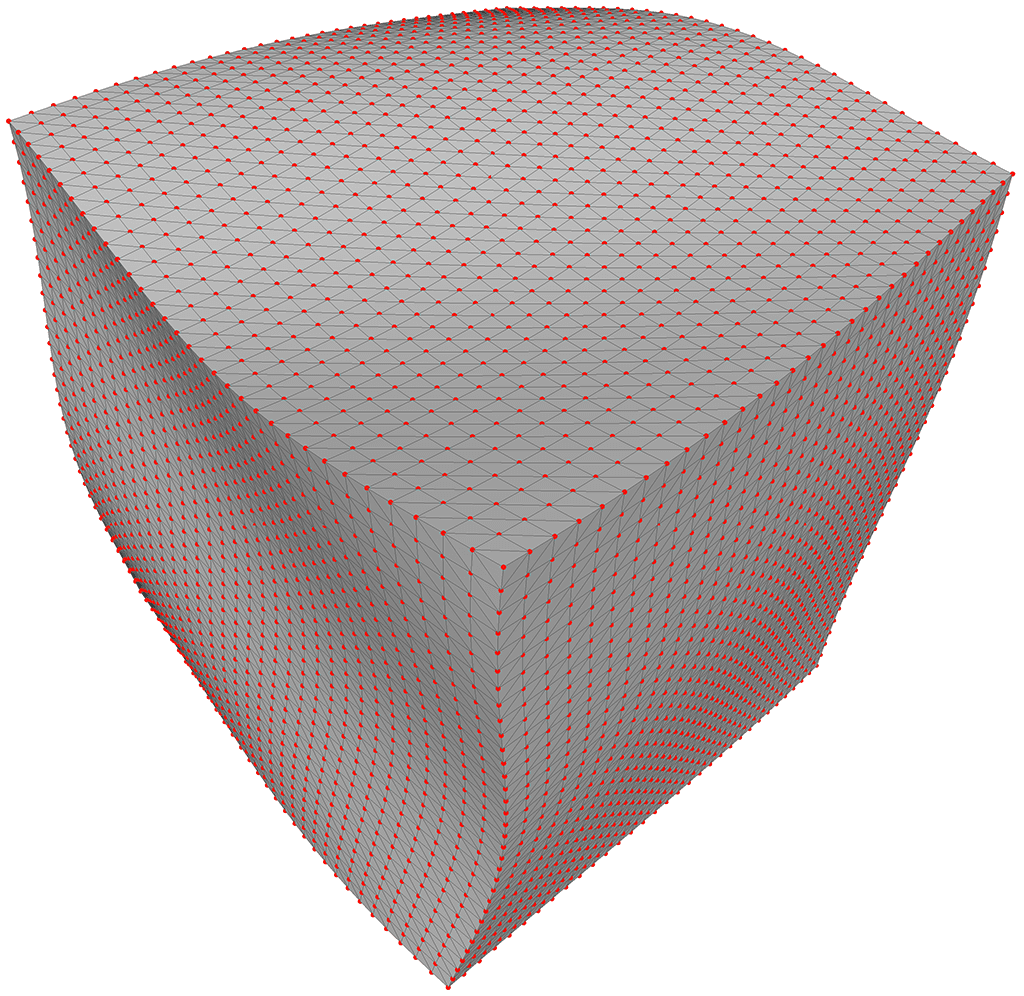}\hfill\hfill
    \includegraphics[width=.22\linewidth]{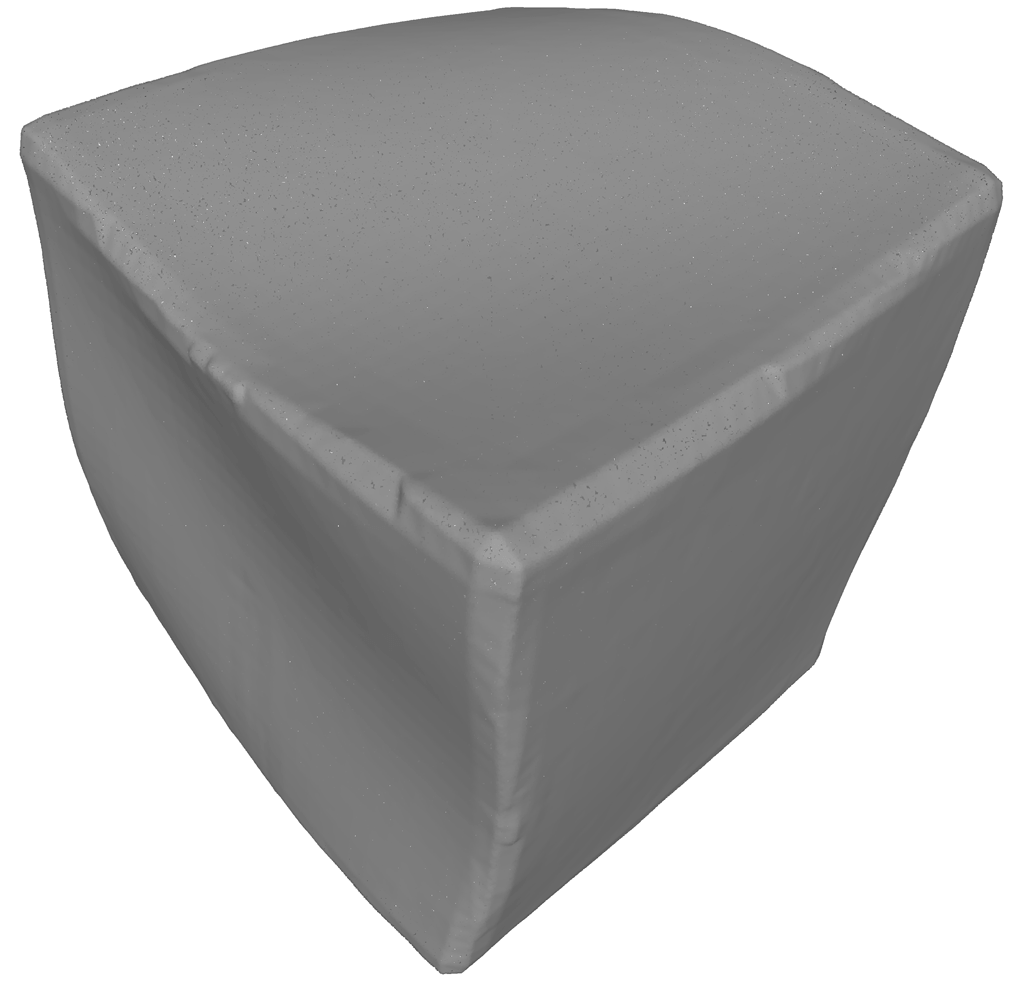}\hfill
    \includegraphics[width=.22\linewidth]{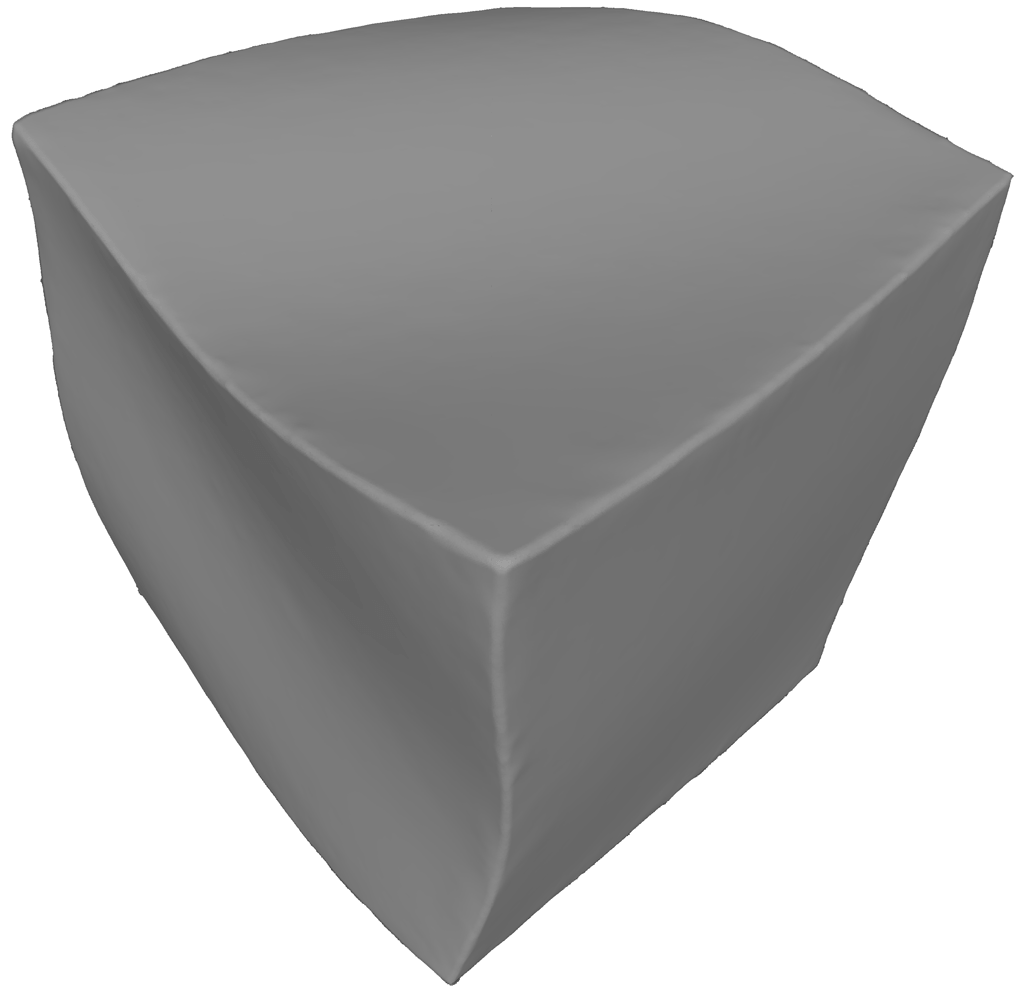}\hfill
    \includegraphics[width=.22\linewidth]{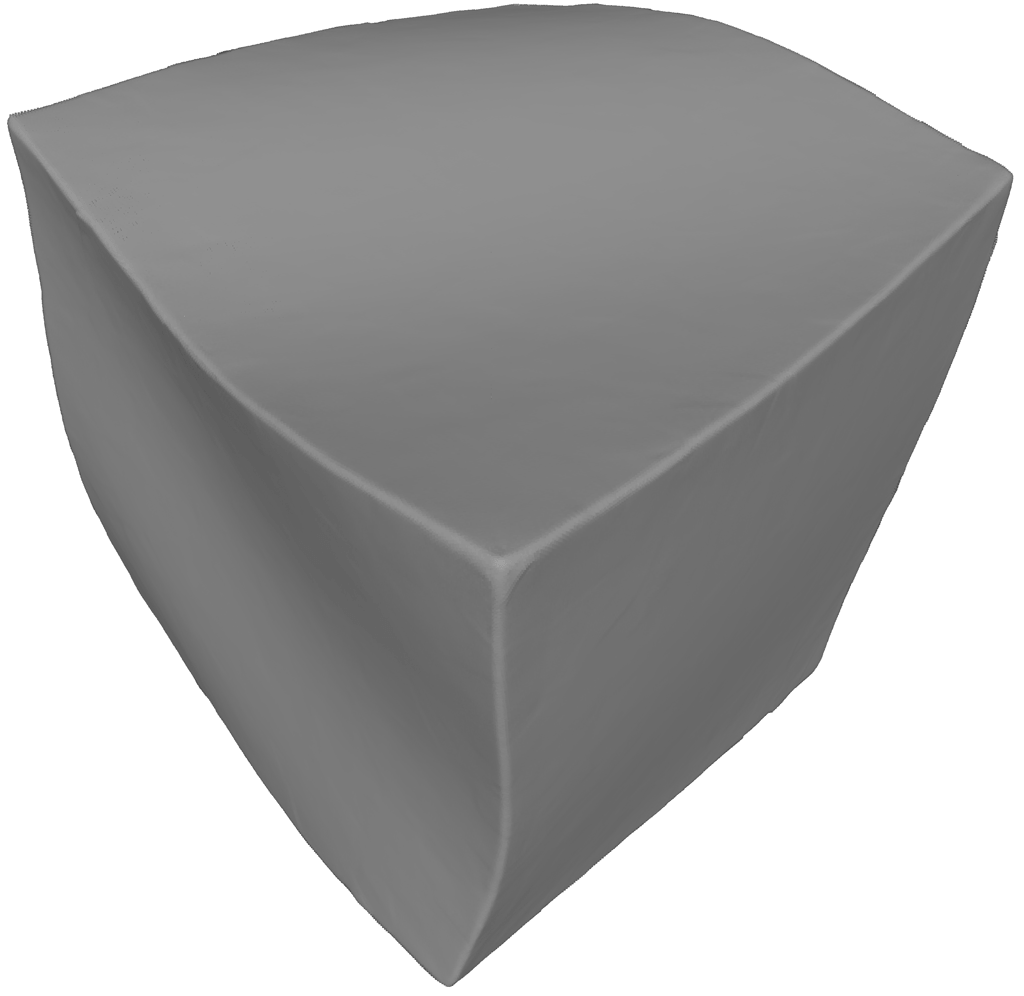}\\[2.5ex]
    Noise\\
    \includegraphics[width=.22\linewidth]{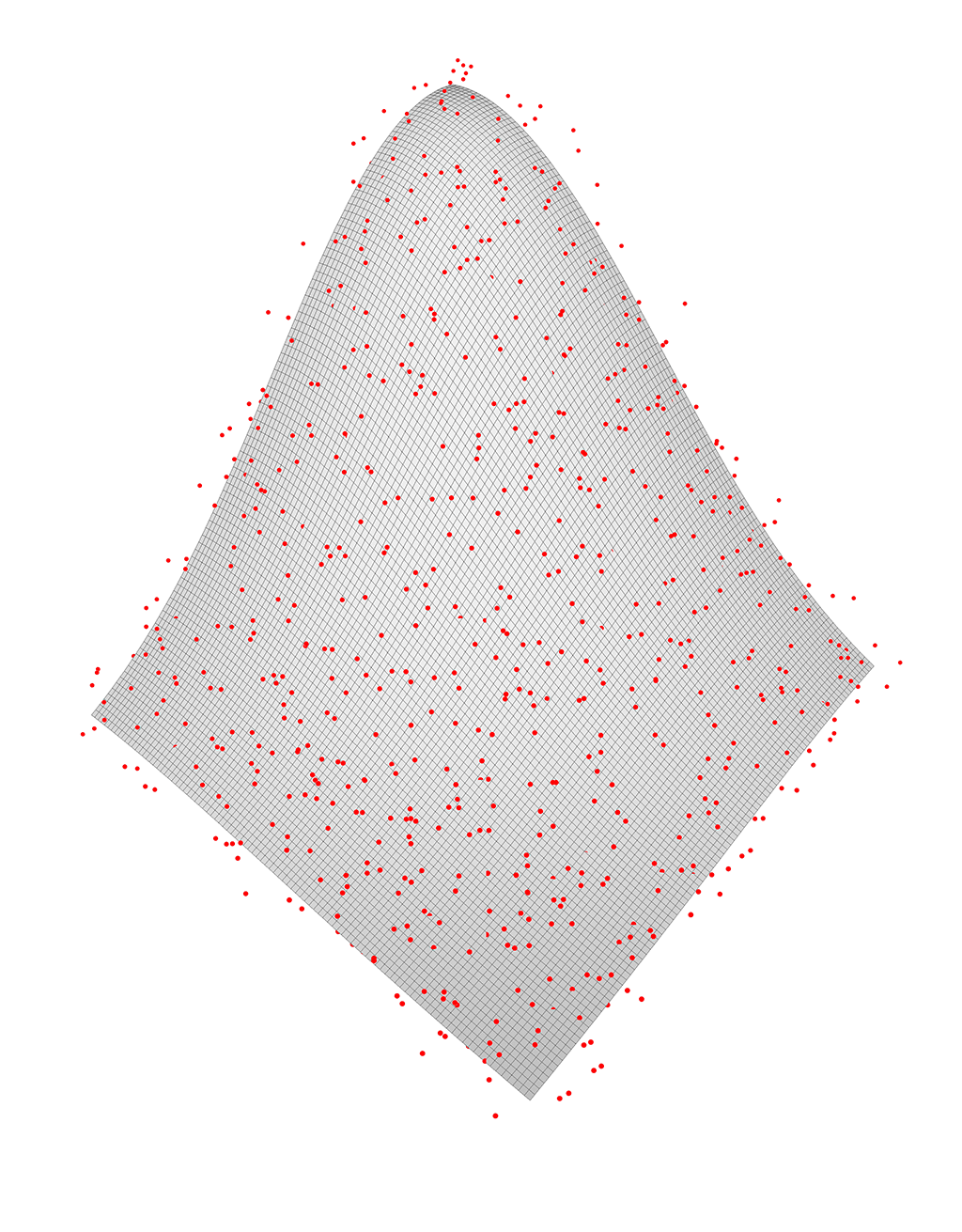}\hfill\hfill
    \includegraphics[width=.22\linewidth]{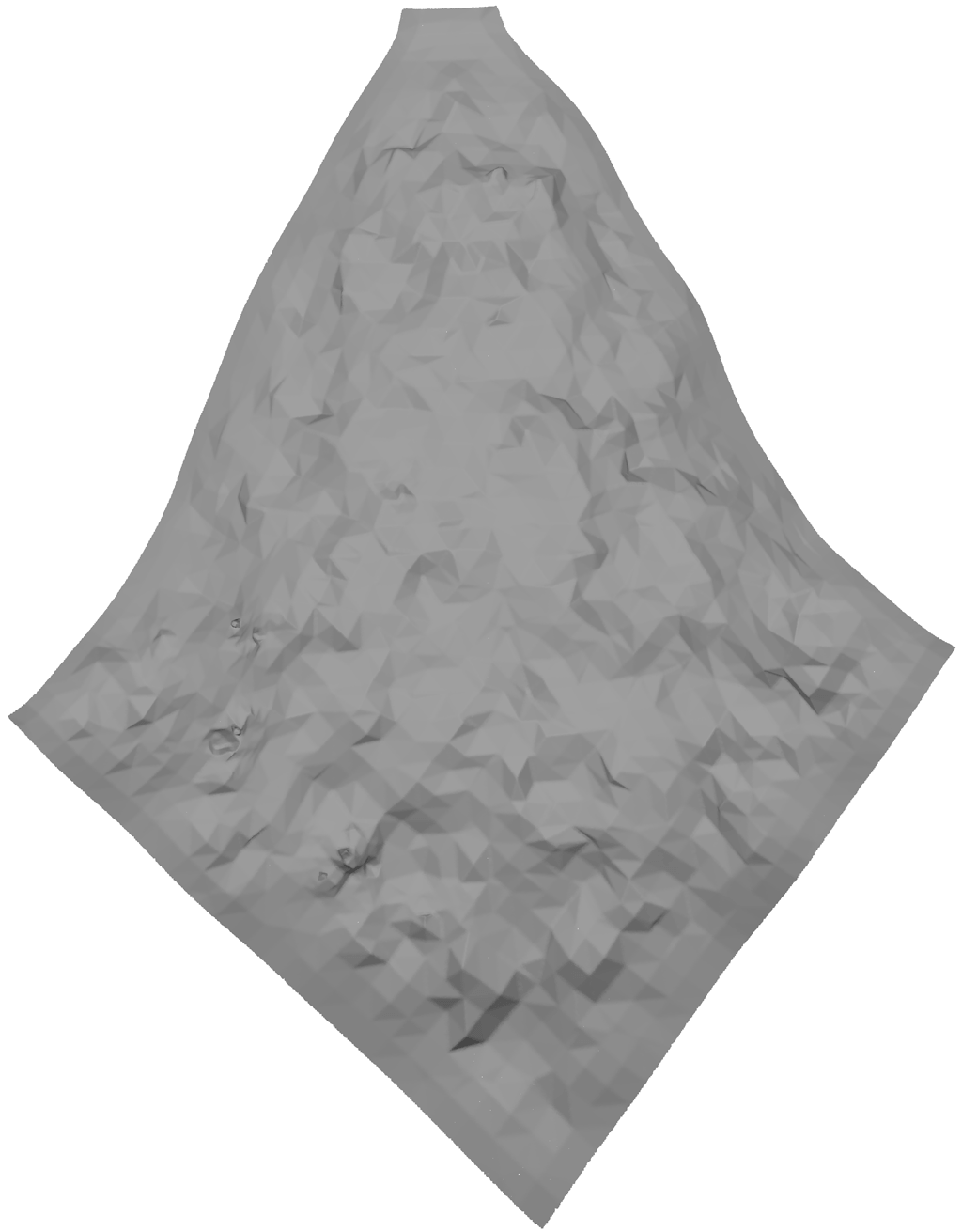}\hfill
    \includegraphics[width=.22\linewidth]{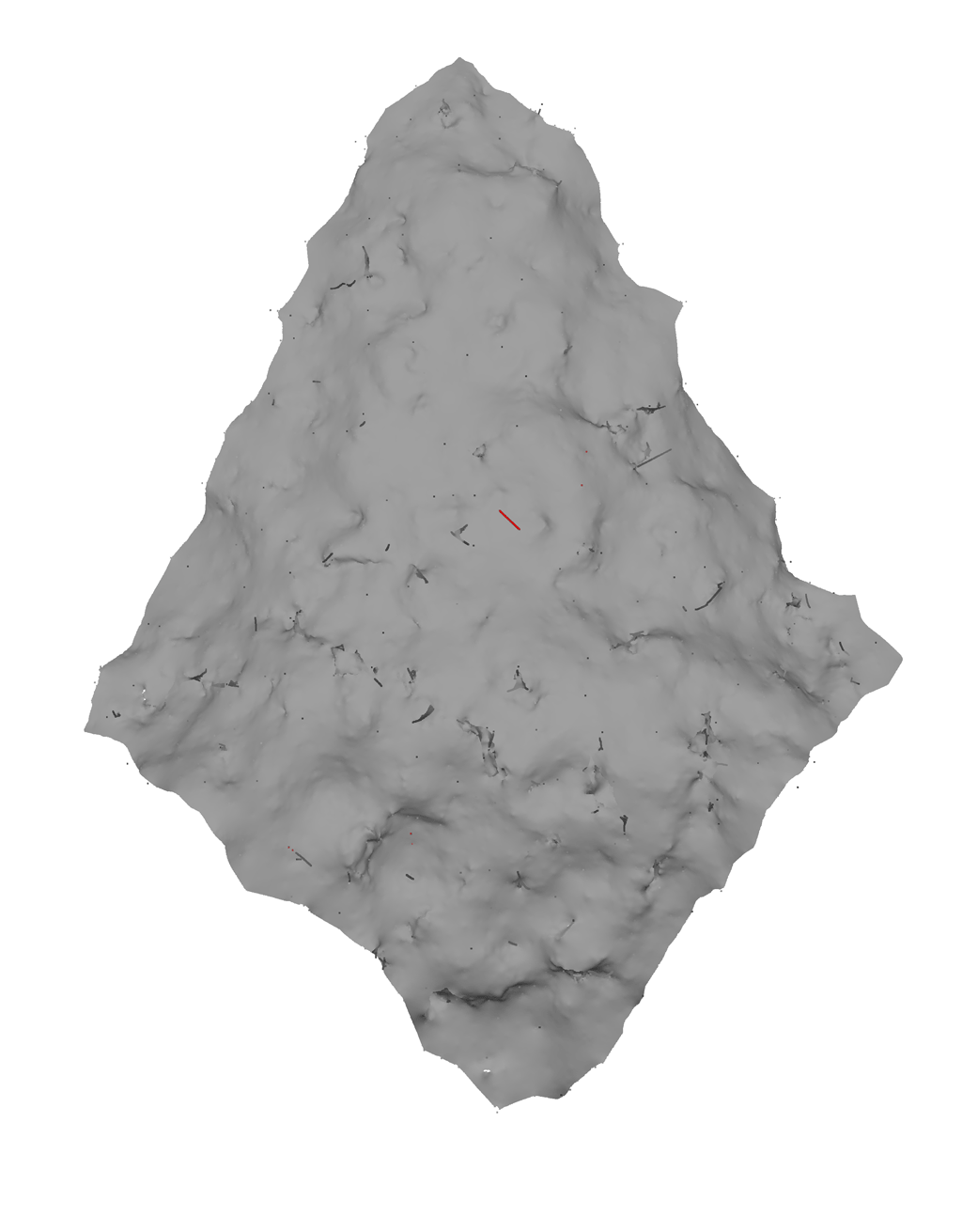}\hfill
    \includegraphics[width=.22\linewidth]{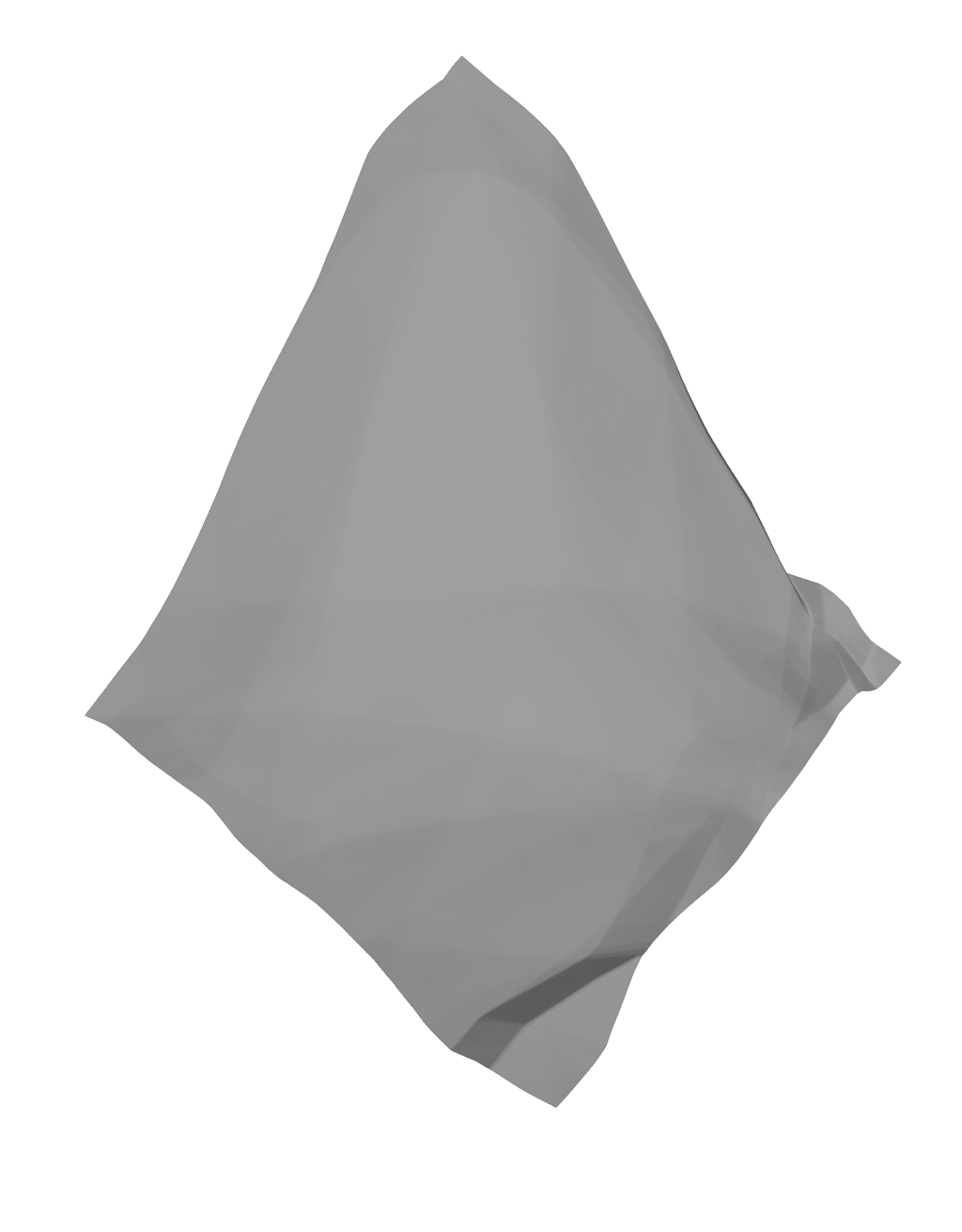}\\[2ex]
    \parbox{.22\linewidth}{\centering Reference}\hfill
    \parbox{.22\linewidth}{\centering Poisson}\hfill
    \parbox{.22\linewidth}{\centering Ear}\hfill
    \parbox{.22\linewidth}{\centering Our}\\[2ex]
    \caption{Example of reconstruction on extreme conditions for our method versus the most representative traditional methods. The red dots in the Reference are the input points for the reconstruction.\vspace{2ex}}
    \label{fig:sharp}
\end{figure}

\paragraph{Noise and Sharp Features.}
As discussed above, the behavior of surface reconstruction methods is particularly challenging in the presence of 
sharp features and noise in the input measurements. 
We performed two additional experiments to compare the behaviour of our architecture with the most representative traditional methods on both noisy point clouds and models with sharp features. Schreened Poisson Surface Reconstruction \cite{KazhdanH13} (Figure \ref{fig:sharp}, left) is very robust to noise, but fails at capturing sharp features. EAR (Figure \ref{fig:sharp}, middle) is the opposite: it captures sharp features accurately, but being an interpolatory method fails at reconstructing noisy inputs, thus introducing spurious points and regions during the reconstruction. Our method (Figure \ref{fig:sharp}, right) does not suffer from these limitations, robustly fitting noisy inputs and capturing sharp features.

\paragraph{Generating a Triangle Mesh.}
Our method generates a collection of local charts, which can be sampled at an arbitrary resolution. We can generate a triangle mesh by using off-the-shelf-techniques such as Poisson Surface Reconstruction \cite{KazhdanH13} on our dense point clouds. We provide meshes reconstructed in this way for all the benchmark models at {\small \url{https://github.com/fwilliams/deep-geometric-prior}}.

\paragraph{Comparison with AtlasNet \cite{groueix2018atlasnet}.}

\begin{figure}
    \centering\footnotesize
    \includegraphics[width=\linewidth]{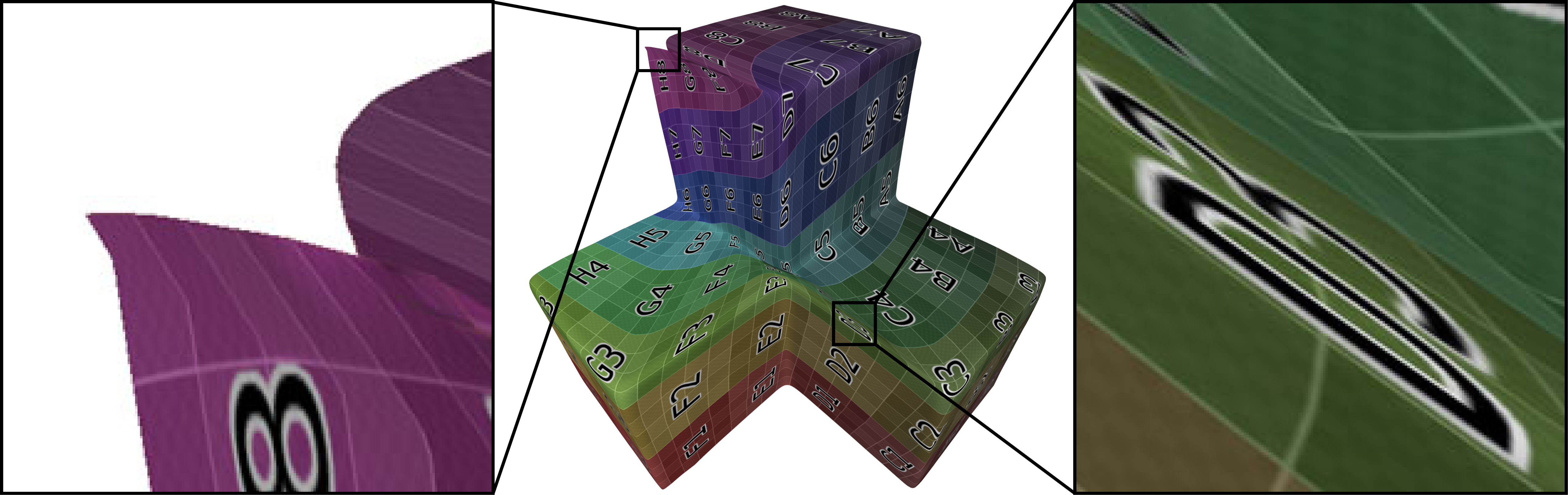}\par
    Chamfer\\[2.5ex]
    \includegraphics[width=\linewidth]{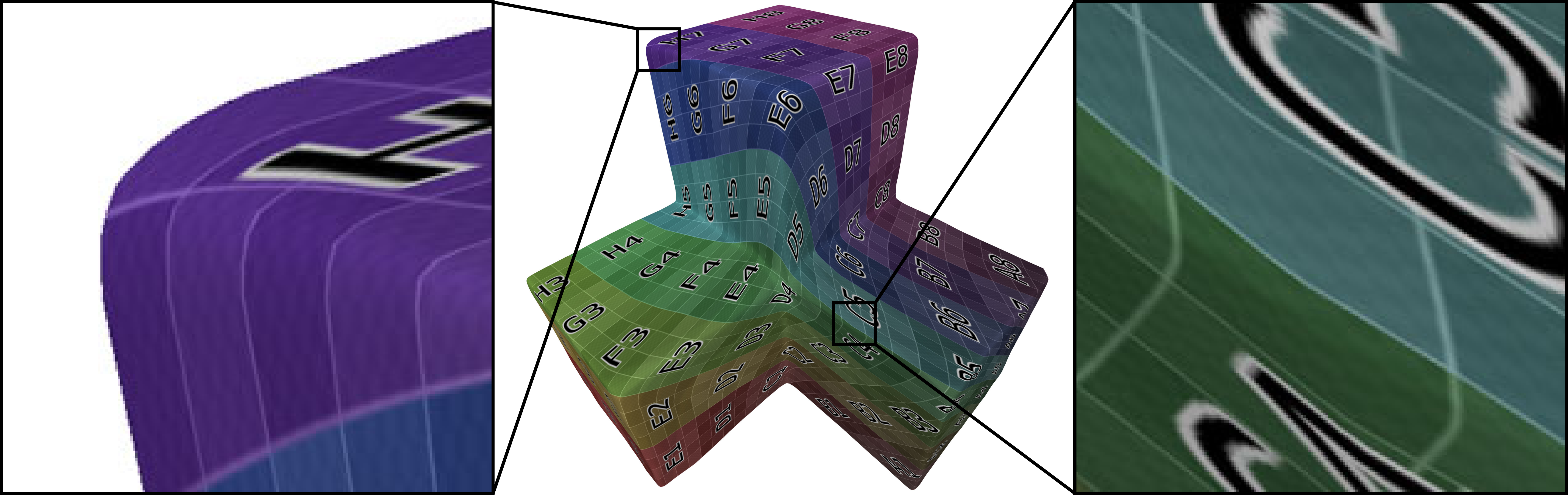}\\[1.5ex]
    Wasserstein (w/ Sinkhorn)\\[2ex]
    \caption{Surface reconstruction of stacked cubes using a single chart, with two different choices of metric. We verify that the Wasserstein metric, even with the Sinkhorn entropy-regularization, provides a more uniform parametrisation, as well as bijective correspondences between overlapping charts.\vspace{2ex}}
    \label{fig:c-vs-w}
\end{figure}

Our atlas construction is related to the AtlasNet model introduced in \cite{groueix2018atlasnet}. AtlasNet is a data-driven reconstruction method using an autoencoder architecture. While their emphasis was on leveraging semantic similarity of shapes and images on several 3D tasks, we focus on high-fidelity point cloud reconstruction in data-sparse regimes, i.e. in absence of any training data. Our main contribution is to show that in such regimes, an even simpler neural network yields state-of-the-art results on surface reconstruction. We also note the following essential differences between our method and AtlasNet.

\begin{itemize}
    \setlength\itemsep{0.4em}
    \item \emph{No Learning:} Our model does not require any training data, and, as a result, we do not need to consider an autoencoder architecture with specific encoders. 
    \item \emph{Transition Functions:}
    Since we have pointwise correspondences, we can define a transition function between overlapping patches $V_p$ and $V_q$ by consistently triangulating corresponding parametric points in $\mathcal{V}_p$ and  $\mathcal{V}_q$. In contrast, AtlasNet does not have such correspondences and thus does \emph{not} produce a true manifold atlas. 
    \item \emph{Patch Selection:} 
    We partition the input into point-sets $\mathcal{X}_p$ that we fit separately. While it is theoretically attractive to attempt to fit each patch to the whole set as it is done in AtlasNet, and let the algorithm figure out the patch partition automatically, in practice the difficulty of the optimization problem leads to unsatisfactory results. In other words, AtlasNet is approximating a \emph{global} matching whereas our model only requires \emph{local} matchings within each patch.
    \item \emph{Wasserstein vs. Chamfer Distance:} As discussed above, the EMD automatically provides transition maps across local charts. AtlasNet considers instead the Chamfer distance between point clouds, which is more efficient to compute but sacrifices the ability to construct bijections in the overlapping regions. Moreover, as illustrated in Figure \ref{fig:c-vs-w},  we observe that Chamfer distances may result in  distortion effects even within local charts.
    \item \emph{Chart Consistency:} We explicitly enforce consistency~\eqref{bibi3_phase2} which has a significant effect on quality, as illustrated in Section \ref{sec:results}, whereas AtlasNet does not produce a real manifold atlas, since it has no definition of transition maps.
\end{itemize}

\begin{figure}\centering\footnotesize
    \includegraphics[width=.49\linewidth]{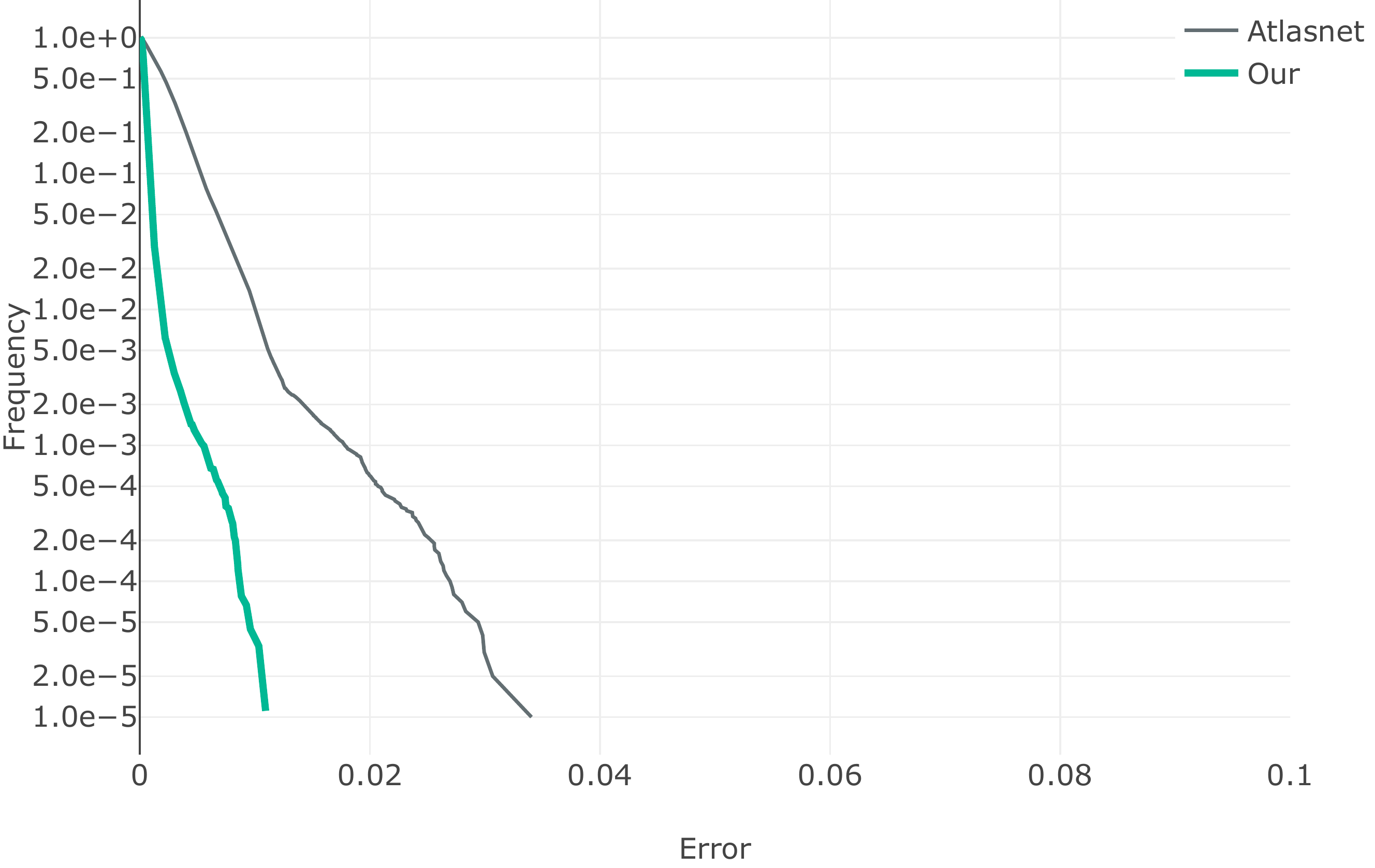}\hfill
    \includegraphics[width=.49\linewidth]{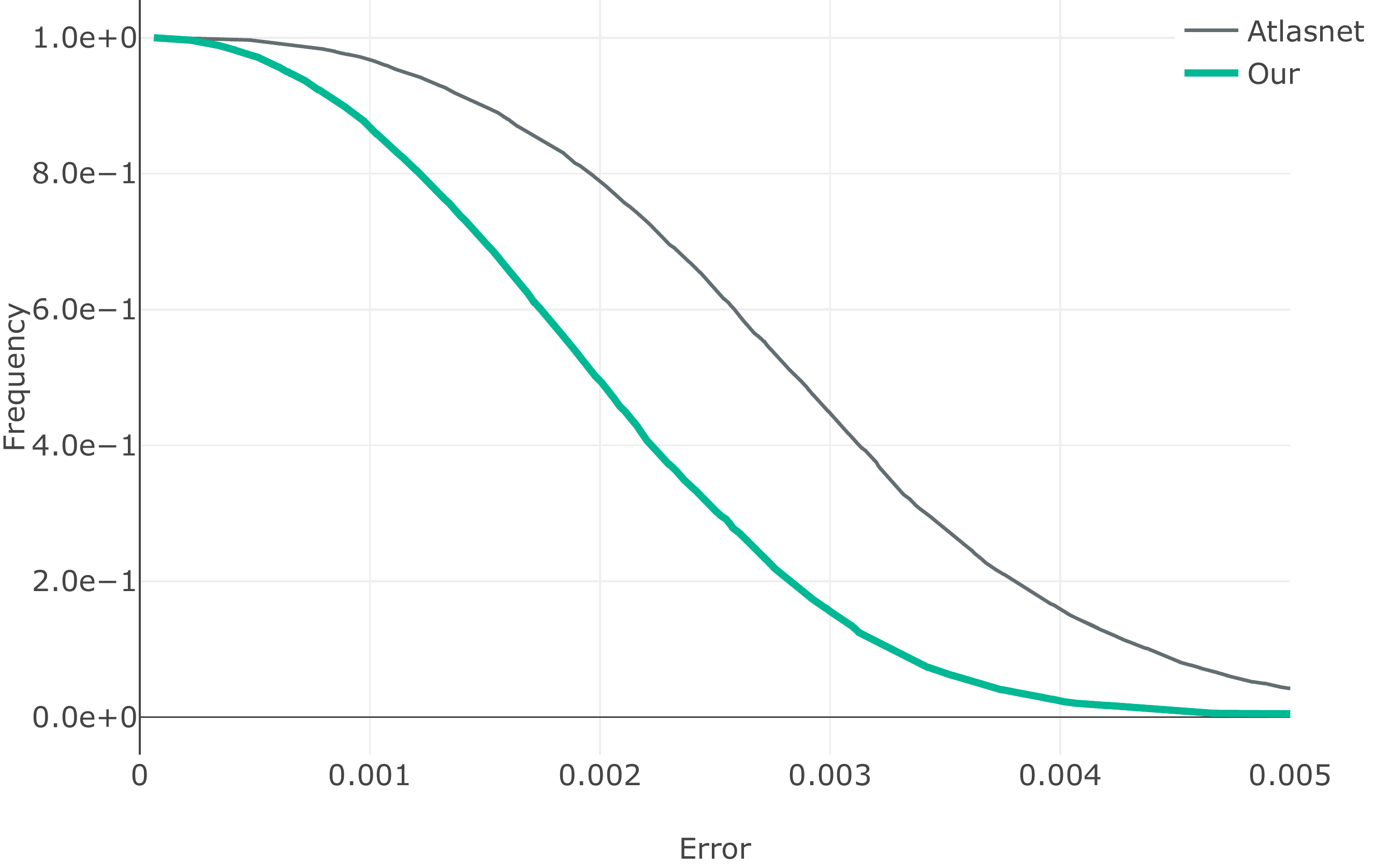}\\
    \includegraphics[width=.45\linewidth]{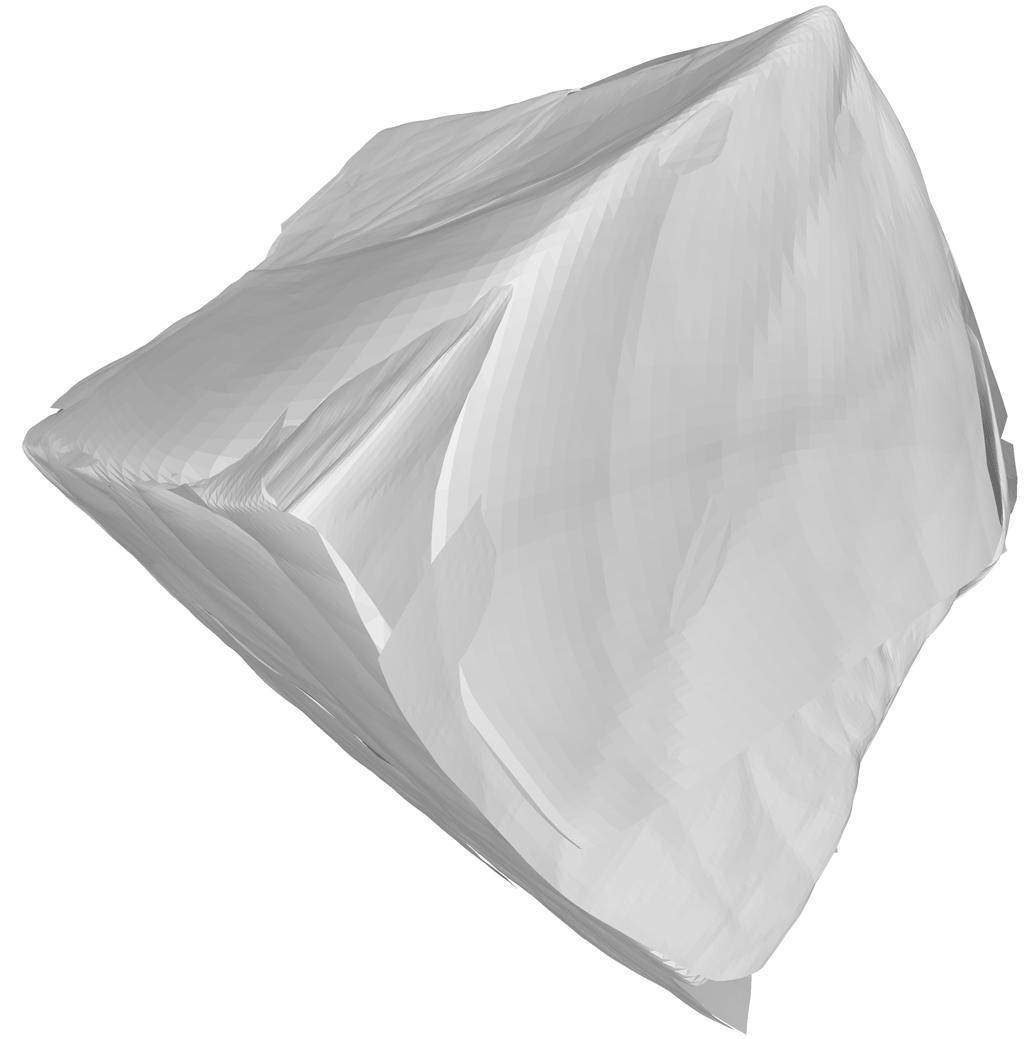}\hfill
    \includegraphics[width=.45\linewidth]{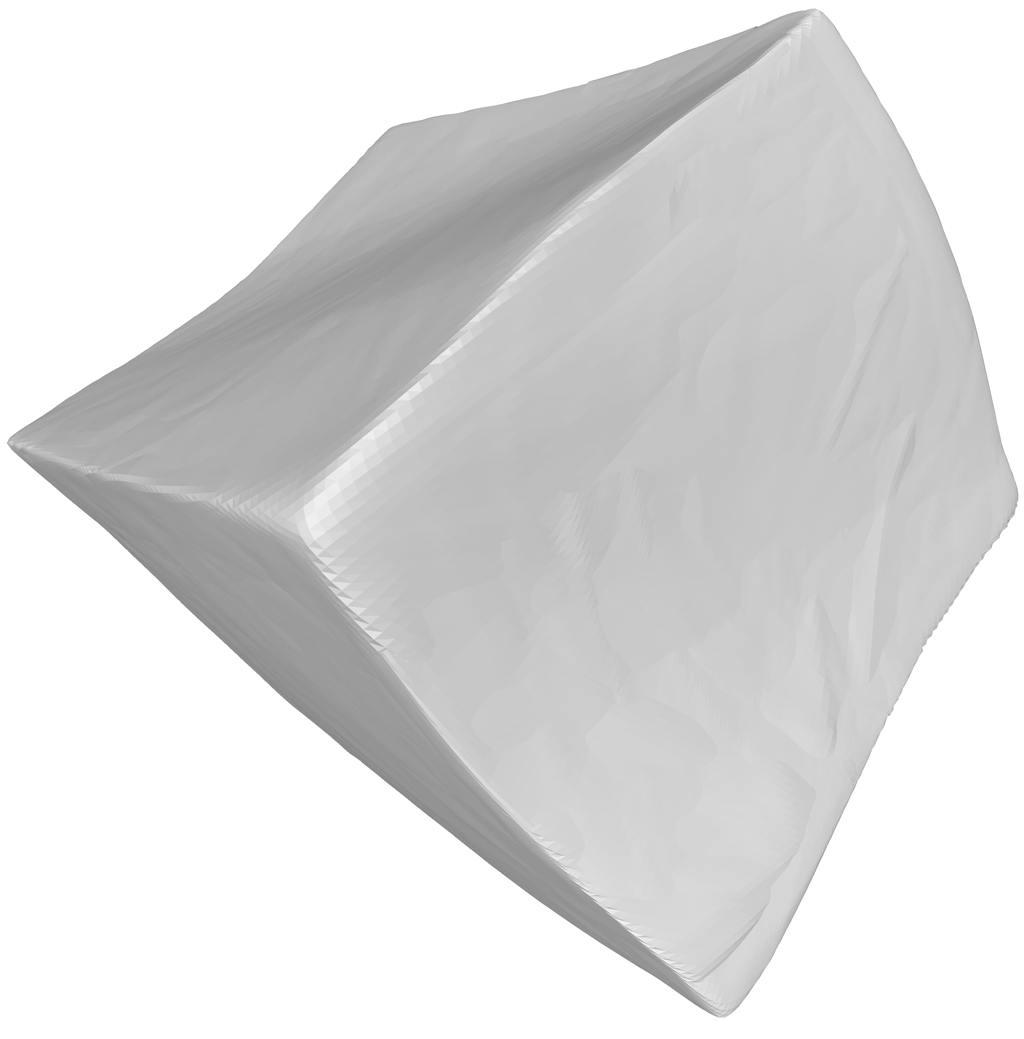}\par
    \parbox{.48\linewidth}{\centering Atlasnet}\hfill
    \parbox{.48\linewidth}{\centering Our}\\[2ex]
    \caption{Comparison with Atlasnet. Top: $d_{\mathrm{rec} \to \mathrm{GT}}$ and $d_{\mathrm{inp} \to \mathrm{rec}}$ cumulative histograms. Bottom: surface reconstruction visualization. Observe that local charts do not have consistent transitions in the Atlasnet output.}
    \label{fig:our-vs-atlasnet}
\end{figure}

We provide quantitative and qualitative comparisons to assess the impact of our architecture choices by adapting AtlasNet to a data-free setting. In this setting, we overfit AtlasNet on a single model with the same number of patches used for our method.
Figure \ref{fig:our-vs-atlasnet} reports both $d_{\mathrm{rec} \to \mathrm{GT}}$ and $d_{\mathrm{inp} \to \mathrm{rec}}$ cumulative histograms on a twisted cube surface using $10$ local charts. We verify that when the Atlasnet architecture is trained to fit the surface using our experimental setup, it is clearly outperformed both quantitatively and qualitatively by our deep geometric prior. We emphasize however that AtlasNet is designed as a data-driven approach, and as such it can leverage semantic information from large training sets. 

\section{Discussion}
\label{sec:limitations}


Neural networks -- particularly in the overparametrised regime -- are remarkably efficient at \emph{curve fitting} in high-dimensional spaces. Despite recent progress in understanding the dynamics of gradient descent in such regimes, their ability to learn and generalize by overcoming the curse of dimensionality remains a major mystery. In this paper, we bypass this high-dimensional regime and concentrate on a standard low-dimensional interpolation problem: surface reconstruction. We demonstrate that in this regime neural networks also have an intriguing capacity to reconstruct complex signal structures while providing robustness to noise. 

Our model is remarkably simple, combining two key principles: (i) construct local piece-wise linear charts by means of a vanilla ReLU fully-connected network, and (ii) use Wasserstein distances in each neighborhood, enabling consistent transitions across local charts. The resulting architecture, when combined with gradient descent, provides a ``deep geometric prior'' that is shown to outperform existing surface-reconstruction methods, which rely on domain-specific geometric assumptions. The theoretical analysis of this deep geometric prior is our next focus, which should address questions such as how the geometry of the surface informs the design the neural network architecture, or why is gradient descent biasing towards locally regular reconstructions. 

Despite these promising directions, we also note the  limitations our approach is facing. In particular, our method is currently substantially more expensive than alternatives. One natural possibility to accelerate it, would be to train a separate neural network model to provide an efficient initialization for the local chart minimisation \eqref{bibi2}, similarly as in neural sparse coders \cite{gregor2010learning}.
Another important question for future research is the adaptive patch selection, which would leverage the benefits of multiscale approximations.

\section{Acknowledgements}

We are grateful to the NYU HPC staff for providing computing cluster service. 
This project was partially supported by
the NSF CAREER award 1652515,
the NSF grant IIS-1320635,
the NSF grant DMS-1436591,
the NSF grant DMS-1835712,
the NSF grant RI-IIS 1816753,
the SNSF grant P2TIP2\_175859,
the Alfred P. Sloan Foundation,
the Moore-Sloan Data Science Environment, 
the DARPA D3M program, 
NVIDIA, 
Samsung Electronics,
Labex DigiCosme,
DOA W911NF-17-1-0438,
a gift from Adobe Research, 
and a gift from nTopology. 
Any opinions, findings, and conclusions or recommendations expressed in this material are those of the authors and do not necessarily reflect the views of DARPA. The authors would also like to thank the anonymous reviewers for their time and effort.

{
\bibliographystyle{ieee}
\bibliography{main.bib}

\begin{thebibliography}{10}\itemsep=-1pt

\bibitem{adamson2003approximating}
A.~Adamson and M.~Alexa.
\newblock {Approximating and intersecting surfaces from points}.
\newblock In {\em SGP}, page 239, 2003.

\bibitem{alexa2001point}
M.~Alexa, J.~Behr, D.~Fleishman, D.~Levin, and C.~Silva.
\newblock {Point Set Surfaces}.
\newblock In {\em {IEEE VIS}}, page~21, 2001.

\bibitem{altschuler2017near}
J.~Altschuler, J.~Weed, and P.~Rigollet.
\newblock Near-linear time approximation algorithms for optimal transport via
  sinkhorn iteration.
\newblock In {\em Advances in Neural Information Processing Systems}, pages
  1964--1974, 2017.

\bibitem{basri2016efficient}
R.~Basri and D.~Jacobs.
\newblock Efficient representation of low-dimensional manifolds using deep
  networks.
\newblock {\em arXiv preprint arXiv:1602.04723}, 2016.

\bibitem{BergerLNTS13}
M.~Berger, J.~A. Levine, L.~G. Nonato, G.~Taubin, and C.~T. Silva.
\newblock A benchmark for surface reconstruction.
\newblock {\em {ACM TOG}}, 32(2):20:1--20:17, 2013.

\bibitem{BergerTSAGLSS17}
M.~Berger, A.~Tagliasacchi, L.~M. Seversky, P.~Alliez, G.~Guennebaud, J.~A.
  Levine, A.~Sharf, and C.~T. Silva.
\newblock A survey of surface reconstruction from point clouds.
\newblock {\em {CGF}}, 36(1):301--329, 2017.

\bibitem{bowers2010parallel}
J.~Bowers, R.~Wang, L.-Y. Wei, and D.~Maletz.
\newblock Parallel poisson disk sampling with spectrum analysis on surfaces.
\newblock In {\em ACM Transactions on Graphics (TOG)}, volume~29, page 166.
  ACM, 2010.

\bibitem{cuturi2013sinkhorn}
M.~Cuturi.
\newblock Sinkhorn distances: Lightspeed computation of optimal transport.
\newblock In {\em Advances in neural information processing systems}, pages
  2292--2300, 2013.

\bibitem{dai2017shape}
A.~Dai, C.~R. Qi, and M.~Nie{\ss}ner.
\newblock Shape completion using 3d-encoder-predictor cnns and shape synthesis.
\newblock In {\em Proc. IEEE Conf. on Computer Vision and Pattern Recognition
  (CVPR)}, volume~3, 2017.

\bibitem{du2018gradient}
S.~S. Du, X.~Zhai, B.~Poczos, and A.~Singh.
\newblock Gradient descent provably optimizes over-parameterized neural
  networks.
\newblock {\em arXiv preprint arXiv:1810.02054}, 2018.

\bibitem{gregor2010learning}
K.~Gregor and Y.~LeCun.
\newblock Learning fast approximations of sparse coding.
\newblock In {\em Proceedings of the 27th International Conference on
  International Conference on Machine Learning}, pages 399--406. Omnipress,
  2010.

\bibitem{groueix2018atlasnet}
T.~Groueix, M.~Fisher, V.~G. Kim, B.~C. Russell, and M.~Aubry.
\newblock Atlasnet: A papier-m$\backslash$\^{} ach$\backslash$'e approach to
  learning 3d surface generation.
\newblock {\em arXiv preprint arXiv:1802.05384}, 2018.

\bibitem{guennebaud2007algebraic}
G.~Guennebaud and M.~Gross.
\newblock {Algebraic Point Set Surfaces}.
\newblock {\em {SIGGRAPH}}, 26:23--9, 2007.

\bibitem{pcpnet}
P.~Guerrero, Y.~Kleiman, M.~Ovsjanikov, and N.~J. Mitra.
\newblock {PCPNet}: Learning local shape properties from raw point clouds.
\newblock {\em Computer Graphics Forum}, 37(2):75--85, 2018.

\bibitem{gunasekar2018characterizing}
S.~Gunasekar, J.~Lee, D.~Soudry, and N.~Srebro.
\newblock Characterizing implicit bias in terms of optimization geometry.
\newblock {\em arXiv preprint arXiv:1802.08246}, 2018.

\bibitem{han2017high}
X.~Han, Z.~Li, H.~Huang, E.~Kalogerakis, and Y.~Yu.
\newblock High-resolution shape completion using deep neural networks for
  global structure and local geometry inference.
\newblock In {\em Proceedings of IEEE International Conference on Computer
  Vision (ICCV)}, 2017.

\bibitem{hoppe1992surface}
H.~Hoppe, T.~DeRose, T.~Duchamp, J.~McDonald, and W.~Stuetzle.
\newblock {Surface reconstruction from unorganized points}.
\newblock In {\em {SIGGRAPH}}, pages 71--78, 1992.

\bibitem{WGCAZ13}
H.~Huang, S.~Wu, M.~Gong, D.~Cohen{-}Or, U.~M. Ascher, and H.~R. Zhang.
\newblock Edge-aware point set resampling.
\newblock {\em {ACM TOG}}, 32(1):9:1--9:12, 2013.

\bibitem{Huang:2014}
Q.~Huang, F.~Wang, and L.~Guibas.
\newblock Functional map networks for analyzing and exploring large shape
  collections.
\newblock In {\em {SIGGRAPH}}, 2014.

\bibitem{kazhdan2005reconstruction}
M.~Kazhdan.
\newblock {Reconstruction of solid models from oriented point sets}.
\newblock In {\em SGP}, page~73, 2005.

\bibitem{kazhdan2006poisson}
M.~Kazhdan, M.~Bolitho, and H.~Hoppe.
\newblock {Poisson surface reconstruction}.
\newblock In {\em SGP}, page~70, 2006.

\bibitem{KazhdanH13}
M.~M. Kazhdan and H.~Hoppe.
\newblock Screened poisson surface reconstruction.
\newblock {\em {ACM TOG}}, 32(3):29:1--29:13, 2013.

\bibitem{kingma2014adam}
D.~P. Kingma and J.~Ba.
\newblock Adam: A method for stochastic optimization.
\newblock {\em arXiv preprint arXiv:1412.6980}, 2014.

\bibitem{kolluri2005provably}
R.~Kolluri.
\newblock {Provably Good Moving Least Squares}.
\newblock In {\em {SODA}}, pages 1008--1018, 2005.

\bibitem{kuhn1955hungarian}
H.~W. Kuhn.
\newblock The hungarian method for the assignment problem.
\newblock {\em Naval research logistics quarterly}, 2(1-2):83--97, 1955.

\bibitem{li2017algorithmic}
Y.~Li, T.~Ma, and H.~Zhang.
\newblock Algorithmic regularization in over-parameterized matrix recovery.
\newblock {\em arXiv preprint arXiv:1712.09203}, 2017.

\bibitem{manson2008streaming}
J.~Manson, G.~Petrova, and S.~Schaefer.
\newblock {Streaming surface reconstruction using wavelets}.
\newblock 27(5):1411--1420, 2008.

\bibitem{nagai2009smoothing}
Y.~Nagai, Y.~Ohtake, and H.~Suzuki.
\newblock {Smoothing of Partition of Unity Implicit Surfaces for Noise Robust
  Surface Reconstruction}.
\newblock 28(5):1339--1348, 2009.

\bibitem{ohtake2003multi}
Y.~Ohtake, A.~Belyaev, M.~Alexa, G.~Turk, and H.~Seidel.
\newblock {Multi-level partition of unity implicits}.
\newblock {\em {ACM TOG}}, 22(3):463--470, 2003.

\bibitem{ohtake2005integrating}
Y.~Ohtake, A.~Belyaev, and H.~Seidel.
\newblock {An integrating approach to meshing scattered point data}.
\newblock In {\em Symposium on Solid and Physical Modeling}, page~69, 2005.

\bibitem{ohtake20043d}
Y.~Ohtake, A.~G. Belyaev, and H.-P. Seidel.
\newblock 3d scattered data interpolation and approximation with multilevel
  compactly supported rbfs.
\newblock {\em Graphical Models}, 67(3):150--165, 2005.

\bibitem{pietroni2011global}
N.~Pietroni, M.~Tarini, O.~Sorkine, and D.~Zorin.
\newblock Global parametrization of range image sets.
\newblock In {\em ACM Transactions on Graphics (TOG)}, volume~30, page 149.
  ACM, 2011.

\bibitem{Qi:2017}
C.~R. Qi, H.~Su, K.~Mo, and L.~J. Guibas.
\newblock Pointnet: Deep learning on point sets for 3d classification and
  segmentation.
\newblock In {\em {CVPR}}, 2017.

\bibitem{Qi:2017a}
C.~R. Qi, L.~Yi, H.~Su, and L.~J. Guibas.
\newblock Pointnet++: Deep hierarchical feature learning on point sets in a
  metric space.
\newblock {\em arXiv preprint arXiv:1706.02413}, 2017.

\bibitem{soudry2017implicit}
D.~Soudry, E.~Hoffer, M.~S. Nacson, S.~Gunasekar, and N.~Srebro.
\newblock The implicit bias of gradient descent on separable data.
\newblock {\em arXiv preprint arXiv:1710.10345}, 2017.

\bibitem{sukhbaatar2016learning}
S.~Sukhbaatar, R.~Fergus, et~al.
\newblock Learning multiagent communication with backpropagation.
\newblock In {\em Advances in Neural Information Processing Systems}, pages
  2244--2252, 2016.

\bibitem{ulyanov2017deep}
D.~Ulyanov, A.~Vedaldi, and V.~Lempitsky.
\newblock Deep image prior.
\newblock {\em arXiv preprint arXiv:1711.10925}, 2017.

\bibitem{varley2017shape}
J.~Varley, C.~DeChant, A.~Richardson, J.~Ruales, and P.~Allen.
\newblock Shape completion enabled robotic grasping.
\newblock In {\em Intelligent Robots and Systems (IROS), 2017 IEEE/RSJ
  International Conference on}, pages 2442--2447. IEEE, 2017.

\bibitem{vinyals2015order}
O.~Vinyals, S.~Bengio, and M.~Kudlur.
\newblock Order matters: Sequence to sequence for sets.
\newblock {\em arXiv preprint arXiv:1511.06391}, 2015.

\bibitem{xiong2014robust}
S.~Xiong, J.~Zhang, J.~Zheng, J.~Cai, and L.~Liu.
\newblock Robust surface reconstruction via dictionary learning.
\newblock {\em ACM Transactions on Graphics (TOG)}, 33(6):201, 2014.

\bibitem{yu2018pu}
L.~Yu, X.~Li, C.-W. Fu, D.~Cohen-Or, and P.-A. Heng.
\newblock Pu-net: Point cloud upsampling network.
\newblock In {\em Proceedings of the IEEE Conference on Computer Vision and
  Pattern Recognition}, pages 2790--2799, 2018.

\end{thebibliography}
}

\clearpage
\appendix

\begin{figure}
\parbox{\textwidth}{
    \includegraphics[width=.14\linewidth]{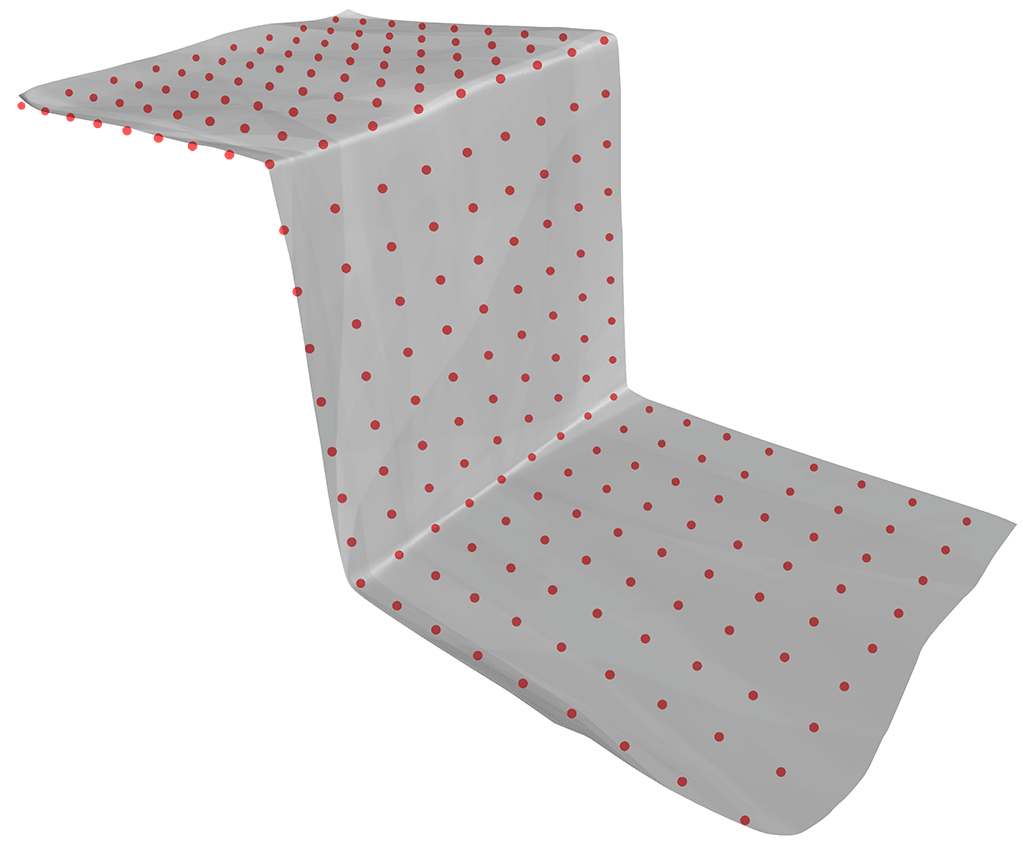}\hfill
    \includegraphics[width=.14\linewidth]{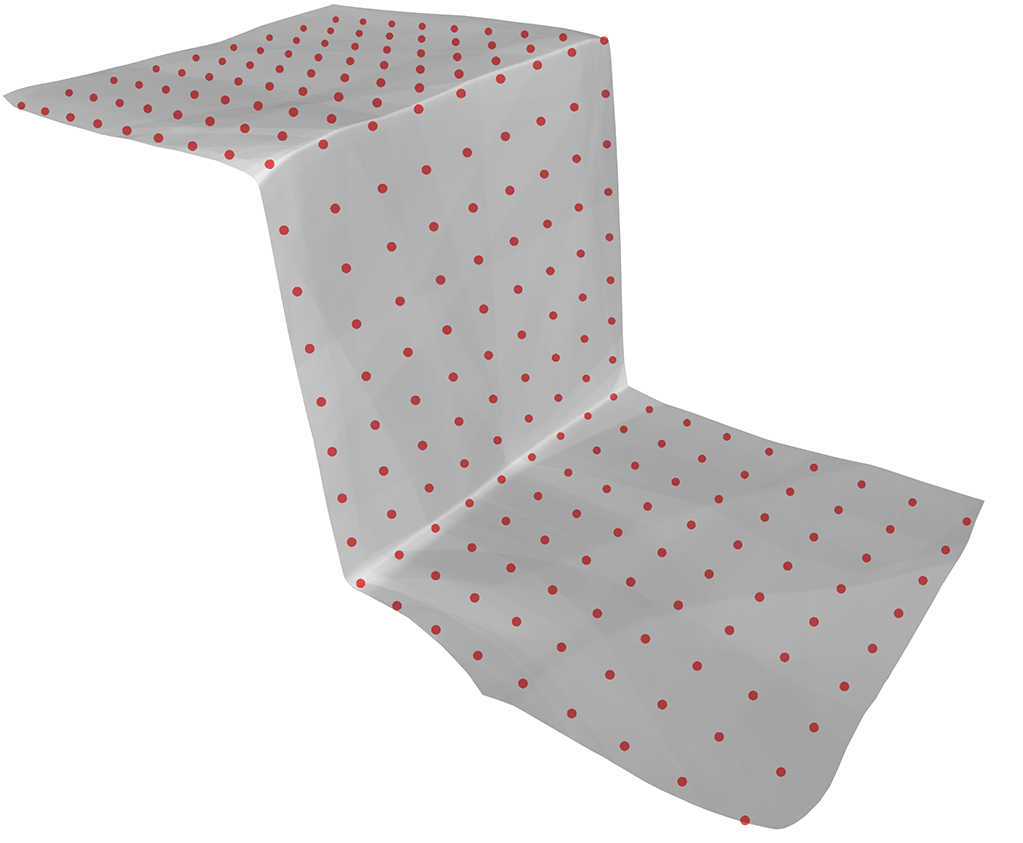}\hfill
    \includegraphics[width=.14\linewidth]{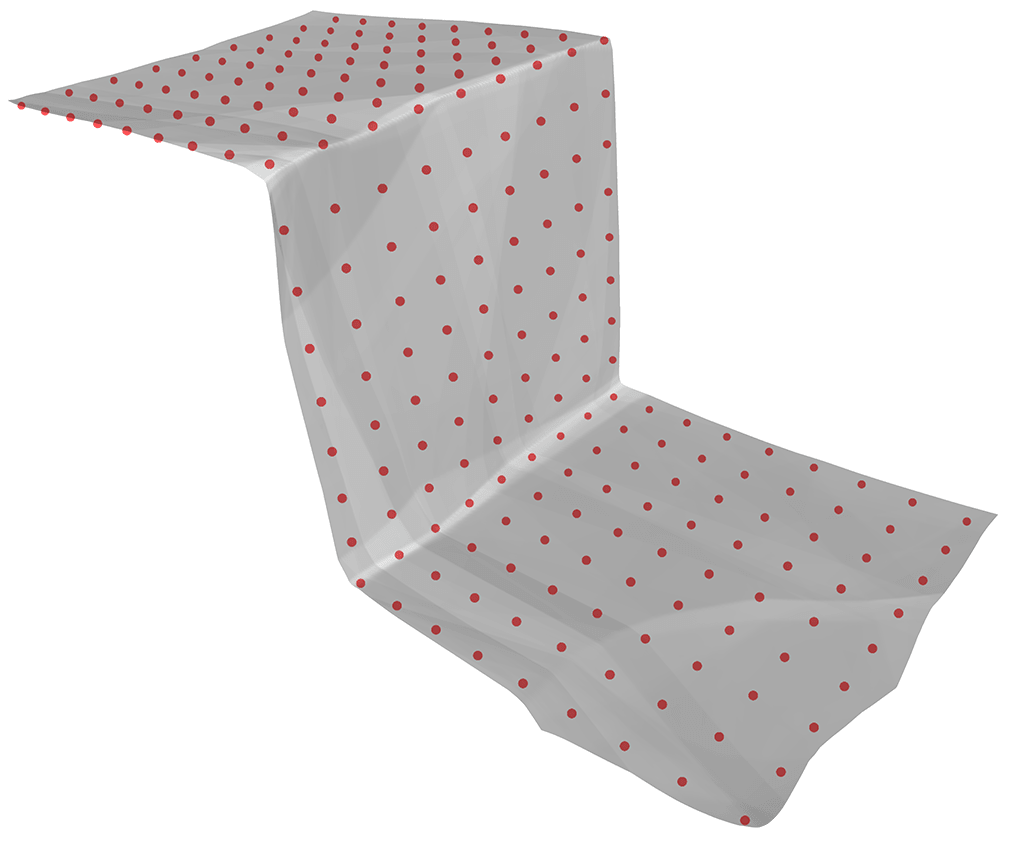}\hfill
    \includegraphics[width=.14\linewidth]{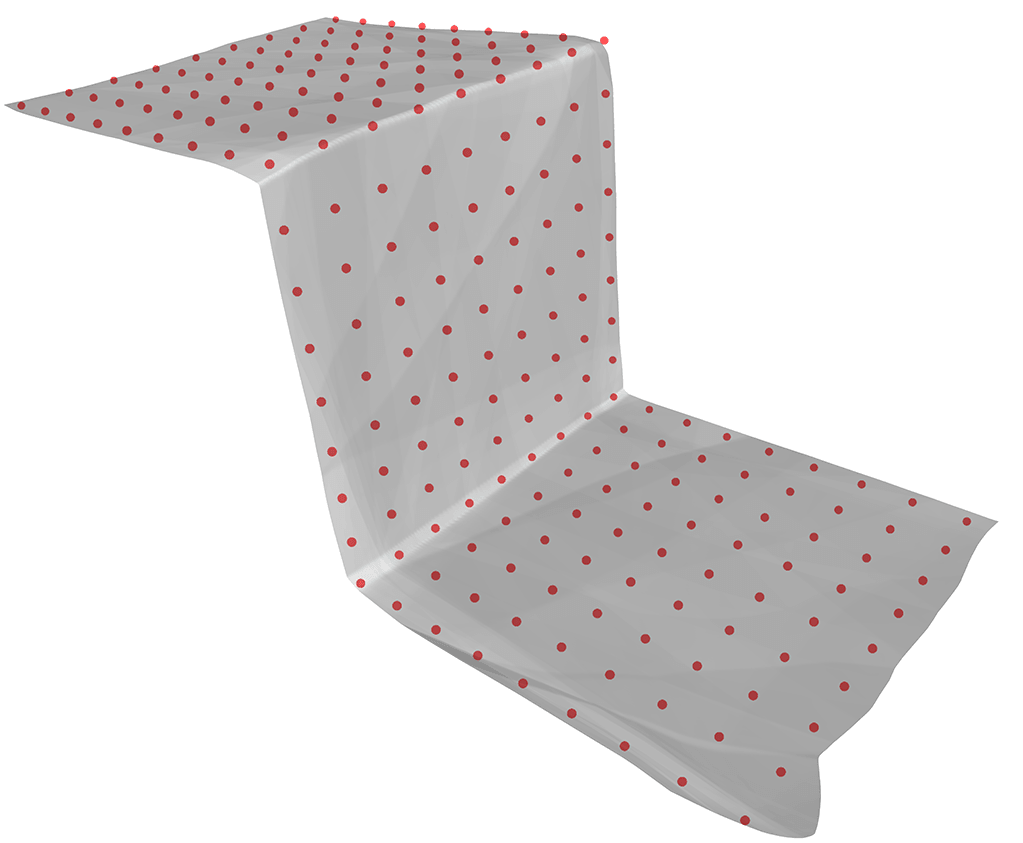}\hfill
    \includegraphics[width=.14\linewidth]{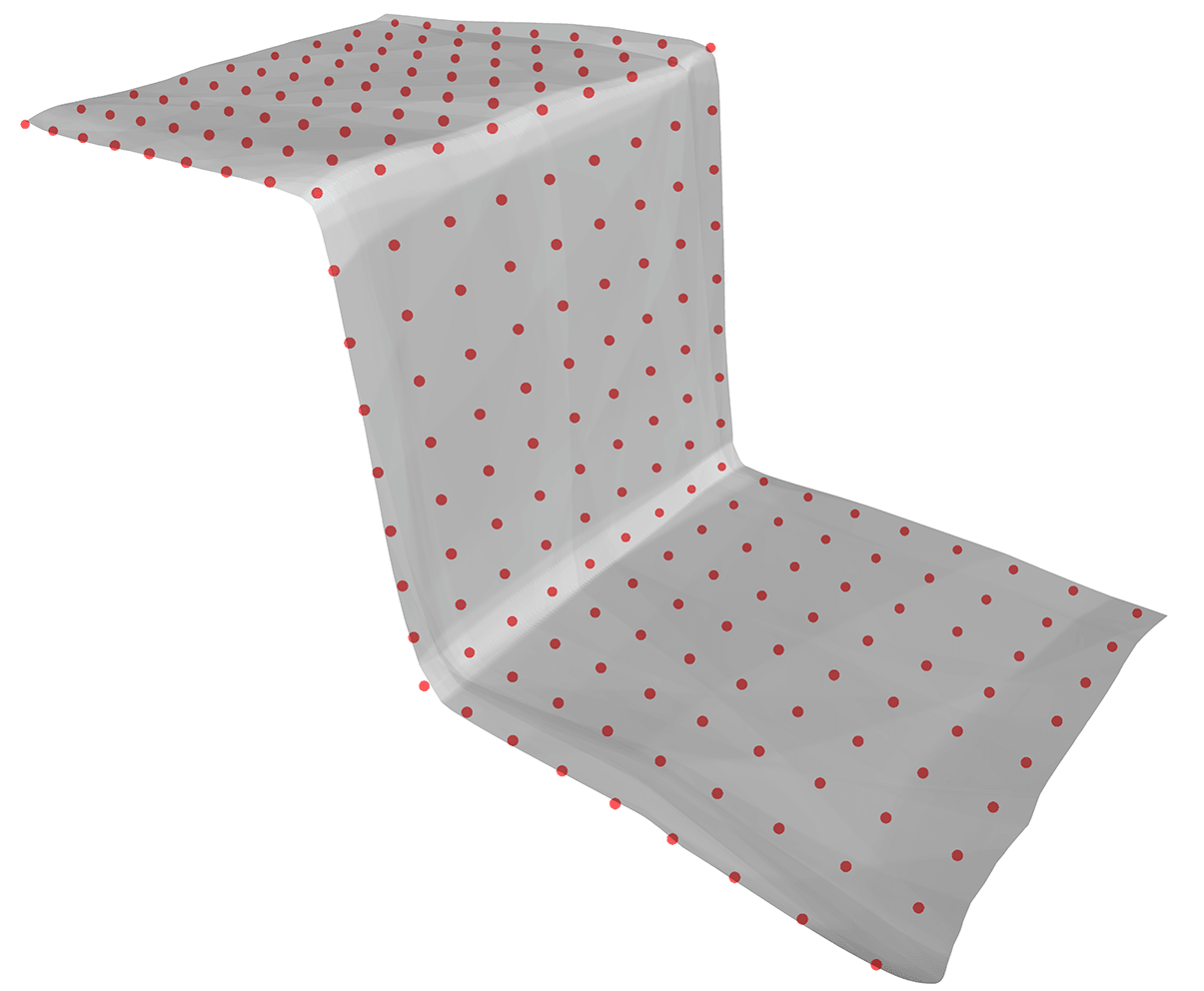}\hfill
    \includegraphics[width=.14\linewidth]{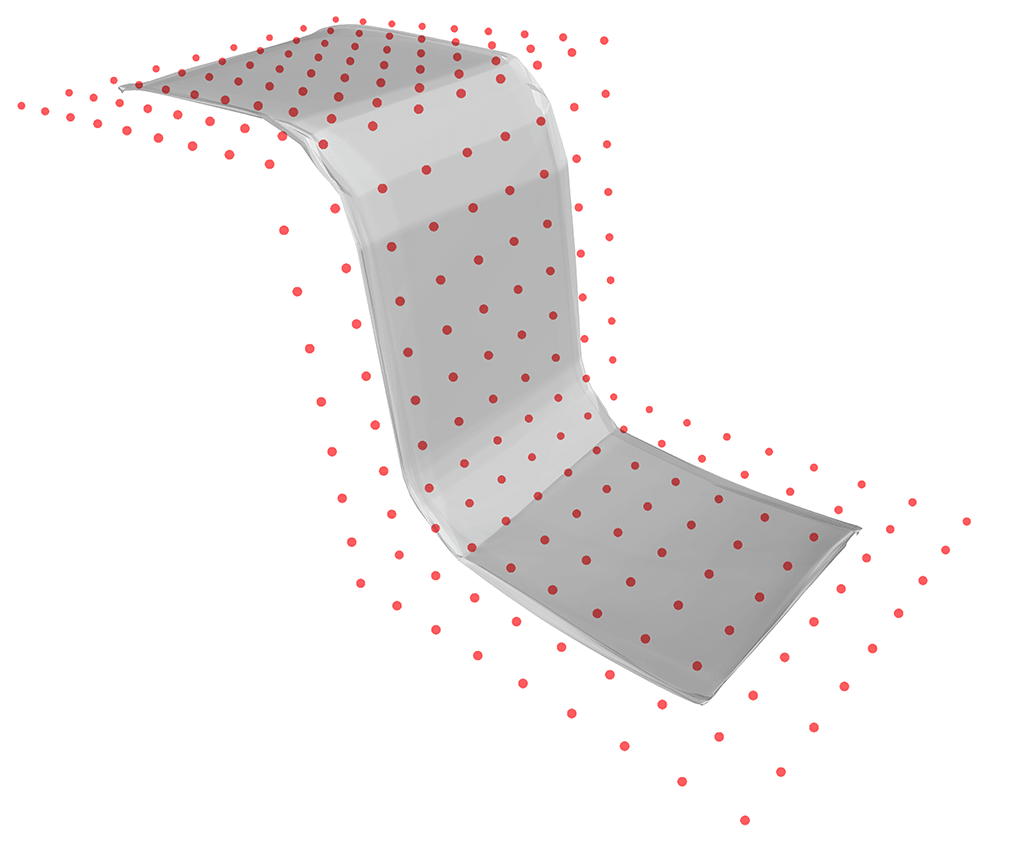}\hfill
    \includegraphics[width=.14\linewidth]{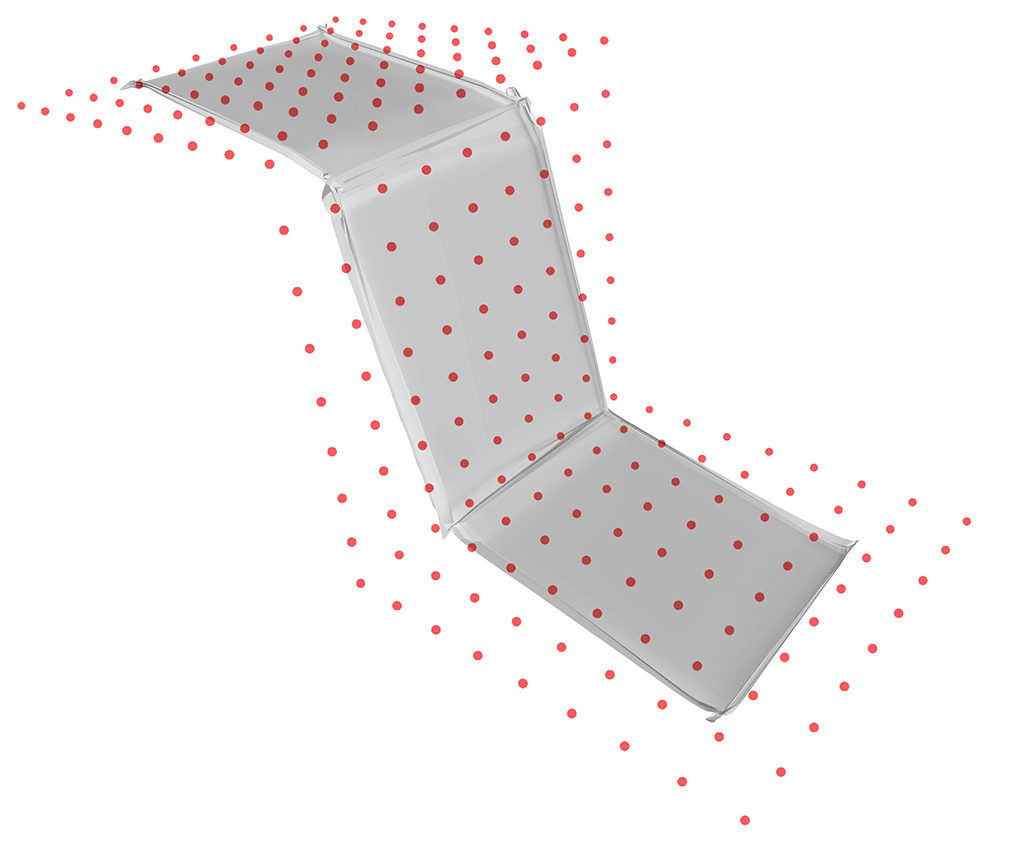}\\[1em]
    \parbox{.14\linewidth}{\centering$\lambda=0.001$\\$W_2=2.46$}\hfill
    \parbox{.14\linewidth}{\centering$\lambda=0.005$\\$W_2=2.87$}\hfill
    \parbox{.14\linewidth}{\centering$\lambda=0.01$\\$W_2=3.26$}\hfill
    \parbox{.14\linewidth}{\centering$\lambda=0.05$\\$W_2=10.76$}\hfill
    \parbox{.14\linewidth}{\centering$\lambda=0.1$\\$W_2=20.98$}\hfill
    \parbox{.14\linewidth}{\centering$\lambda=0.5$\\$W_2=92.91$}\hfill
    \parbox{.14\linewidth}{\centering$\lambda=1$\\$W_2=174.09$}\\[1em]
    \caption{Effect of the parameter $\lambda$ on the reconstruction and the loss $W_2$.}
    \label{fig:lambda-fig}
}
\end{figure}

\section{Supplementary Experiments}
\label{sec:results_supplementary}

\subsection{Effect of the parameter $\lambda$}
In Figure~\ref{fig:lambda-fig}, we demonstrate the effect of varying the Sinkhorn regularization parameter on the final reconstruction of a surface. Smaller values of $\lambda$ yield a better approximation of the Wasserstein distance, and thus, produce better reconstructions of the original points.

\subsection{Kinect reconstruction}
\begin{figure}
    \centering
    \includegraphics[width=0.4\linewidth]{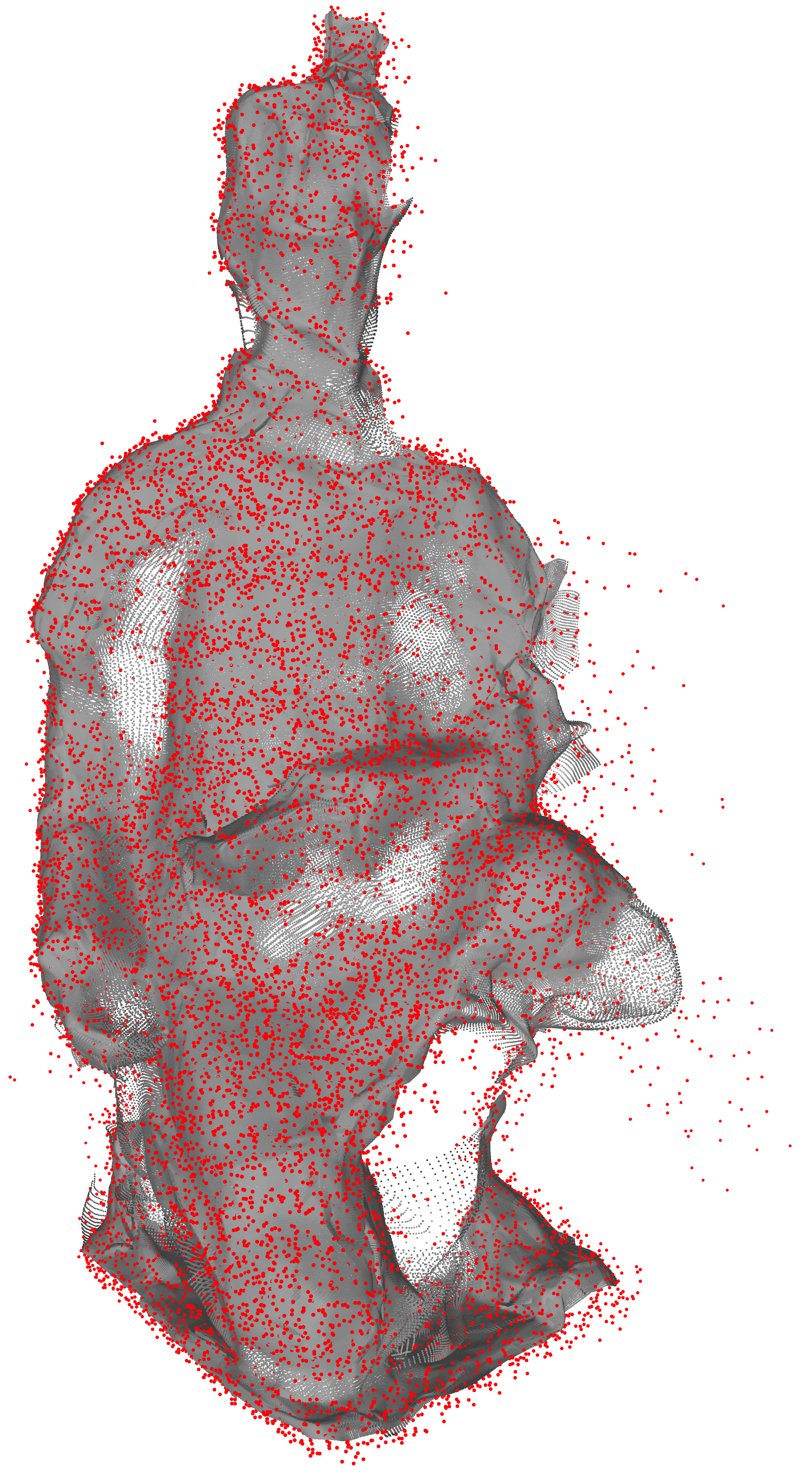}
    \caption{Example of reconstruction form Kinect data (red dots).}
    \label{fig:kinect}
\end{figure}

To demonstrate the effectiveness of our technique on reconstructing point clouds with large quantities of noise and highly non-uniform sampling, we reconstruct a raw point cloud acquired with a Kinect V2 (Figure~\ref{fig:kinect}). In spite of the challenging input, we are still able to produce a smooth reconstruction approximating the geometry of the original object.

\newpage
~\vspace{10.8em}

\subsection{Surface Reconstruction Benchmark}
\begin{figure*}\centering\footnotesize
\parbox{.16\linewidth}{~}\hfill
\includegraphics[width=.32\linewidth]{figs/srb/1/anchor}\hfill
\includegraphics[width=.32\linewidth]{figs/srb/1/gargoyle}\hfill
\parbox{.16\linewidth}{~}\par
\parbox{.16\linewidth}{~}\hfill
\parbox{.32\linewidth}{\centering Anchor}\hfill
\parbox{.32\linewidth}{\centering Gargoyle}\hfill
\parbox{.16\linewidth}{~}\\[4ex]
\includegraphics[width=.32\linewidth]{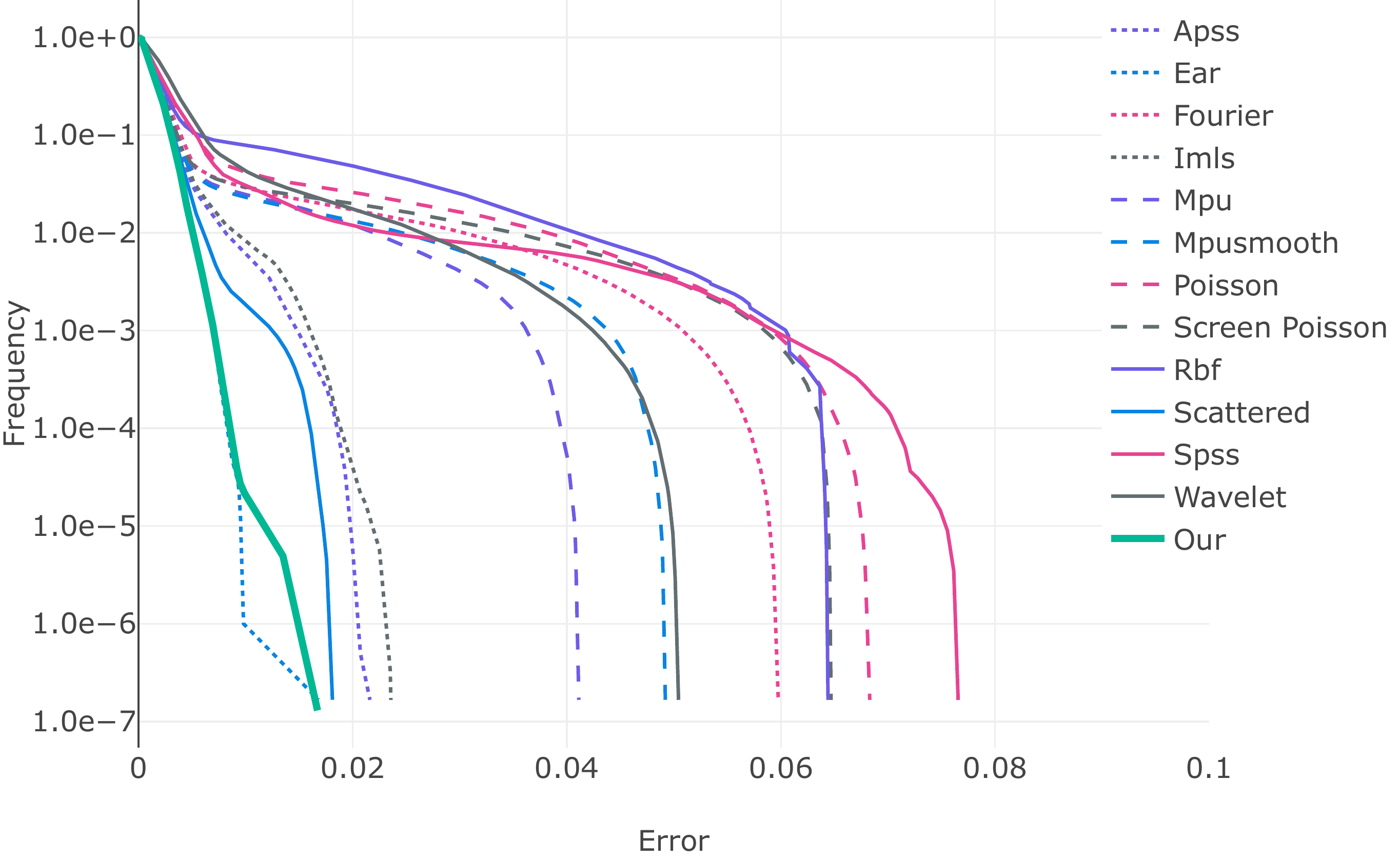}\hfill
\includegraphics[width=.32\linewidth]{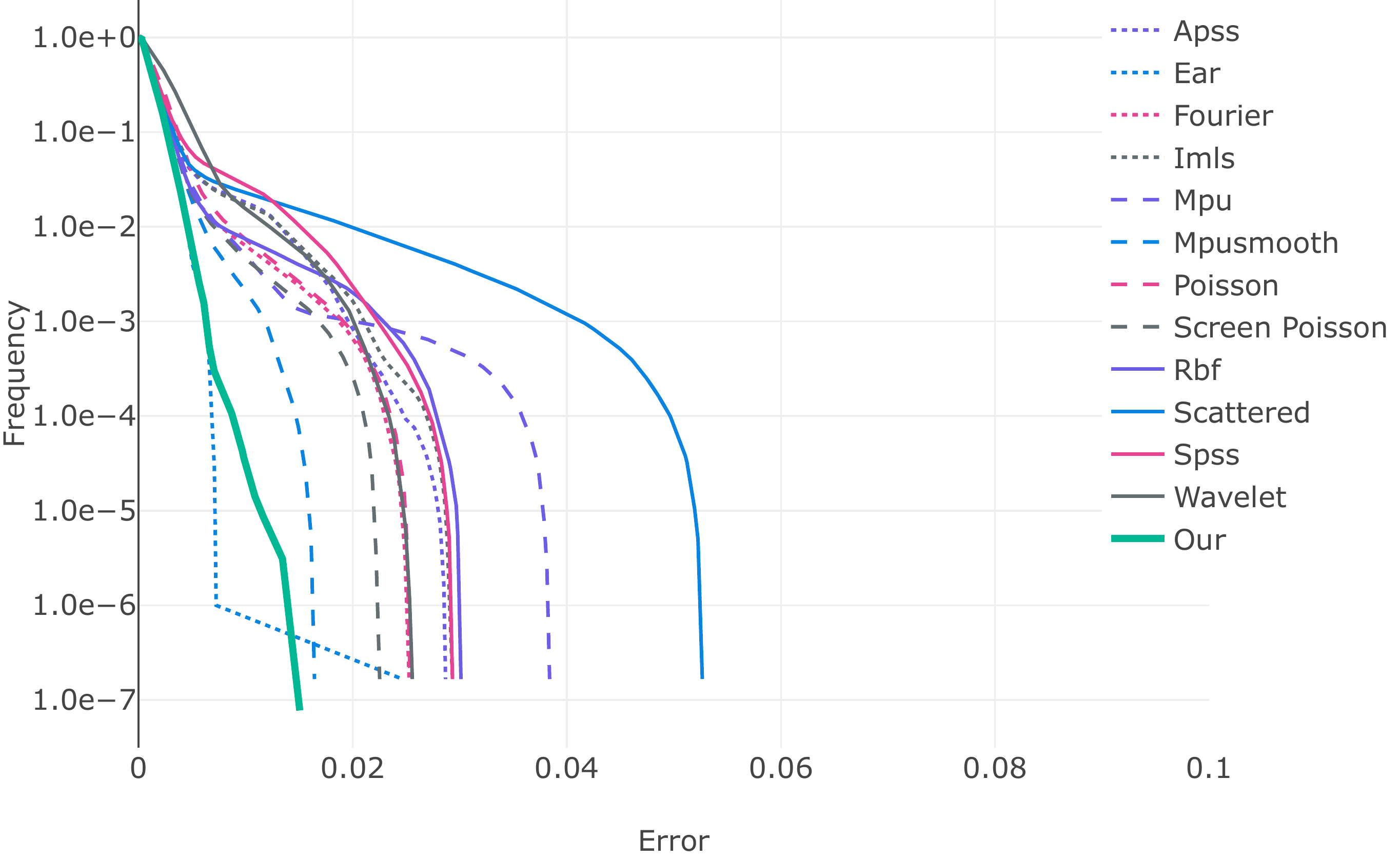}\hfill
\includegraphics[width=.32\linewidth]{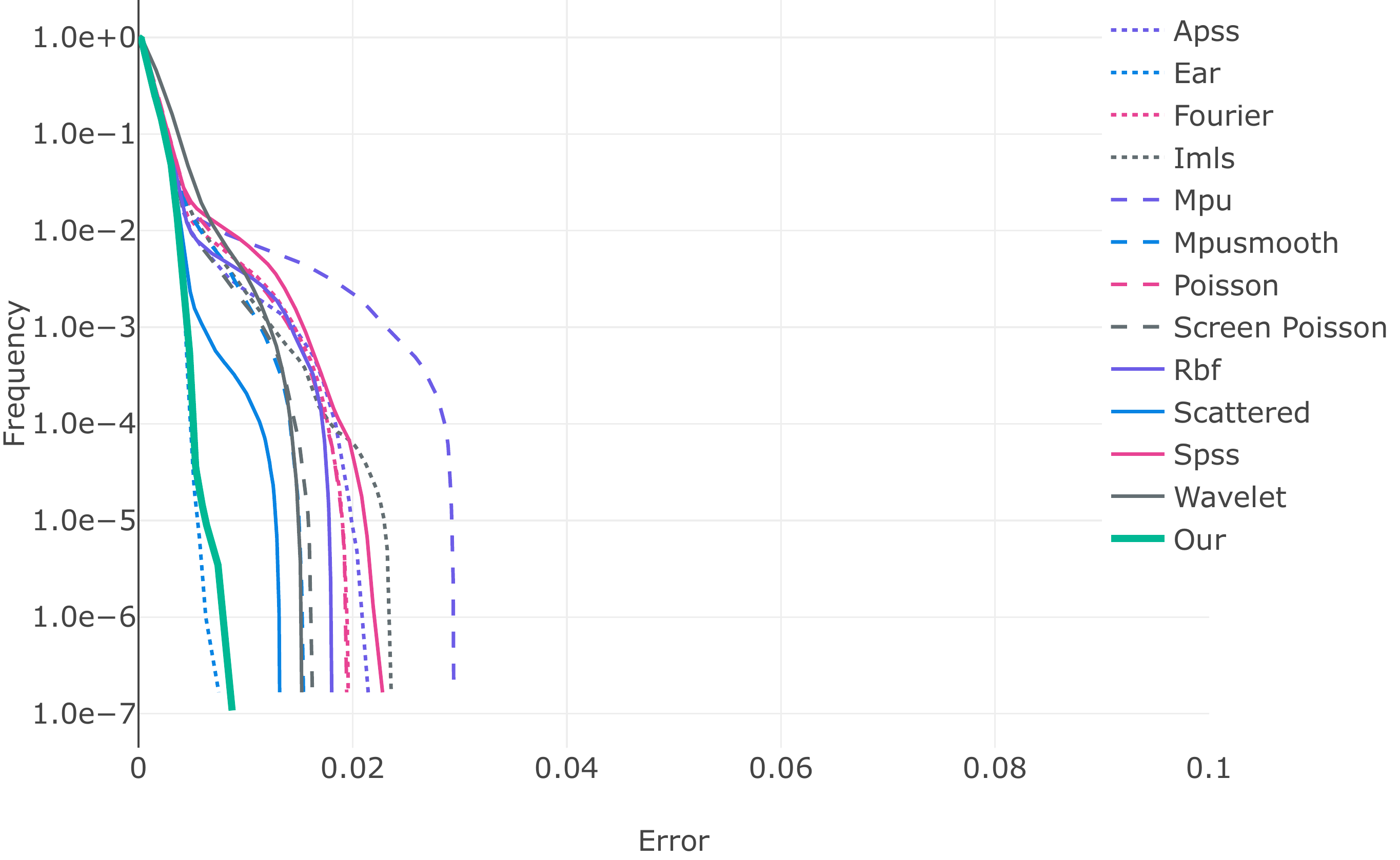}\hfill
\parbox{.32\linewidth}{\centering Daratech}\hfill
\parbox{.32\linewidth}{\centering Danching Children}\hfill
\parbox{.32\linewidth}{\centering Lord Quas}\par
\caption{Percentage of fitted vertices ($y$-axis, log scale) to reach a certain error level  ($x$-axis) for different methods. The errors are computed from the fitted surface to the ground truth.}
\label{fig:first-result-supp}
\end{figure*}

\begin{figure*}\centering\footnotesize
\parbox{.16\linewidth}{~}\hfill
\includegraphics[width=.32\linewidth]{figs/srb/2/anchor}\hfill
\includegraphics[width=.32\linewidth]{figs/srb/2/gargoyle}\hfill
\parbox{.16\linewidth}{~}\par
\parbox{.16\linewidth}{~}\hfill
\parbox{.32\linewidth}{\centering Anchor}\hfill
\parbox{.32\linewidth}{\centering Gargoyle}\hfill
\parbox{.16\linewidth}{~}\\[4ex]
\includegraphics[width=.32\linewidth]{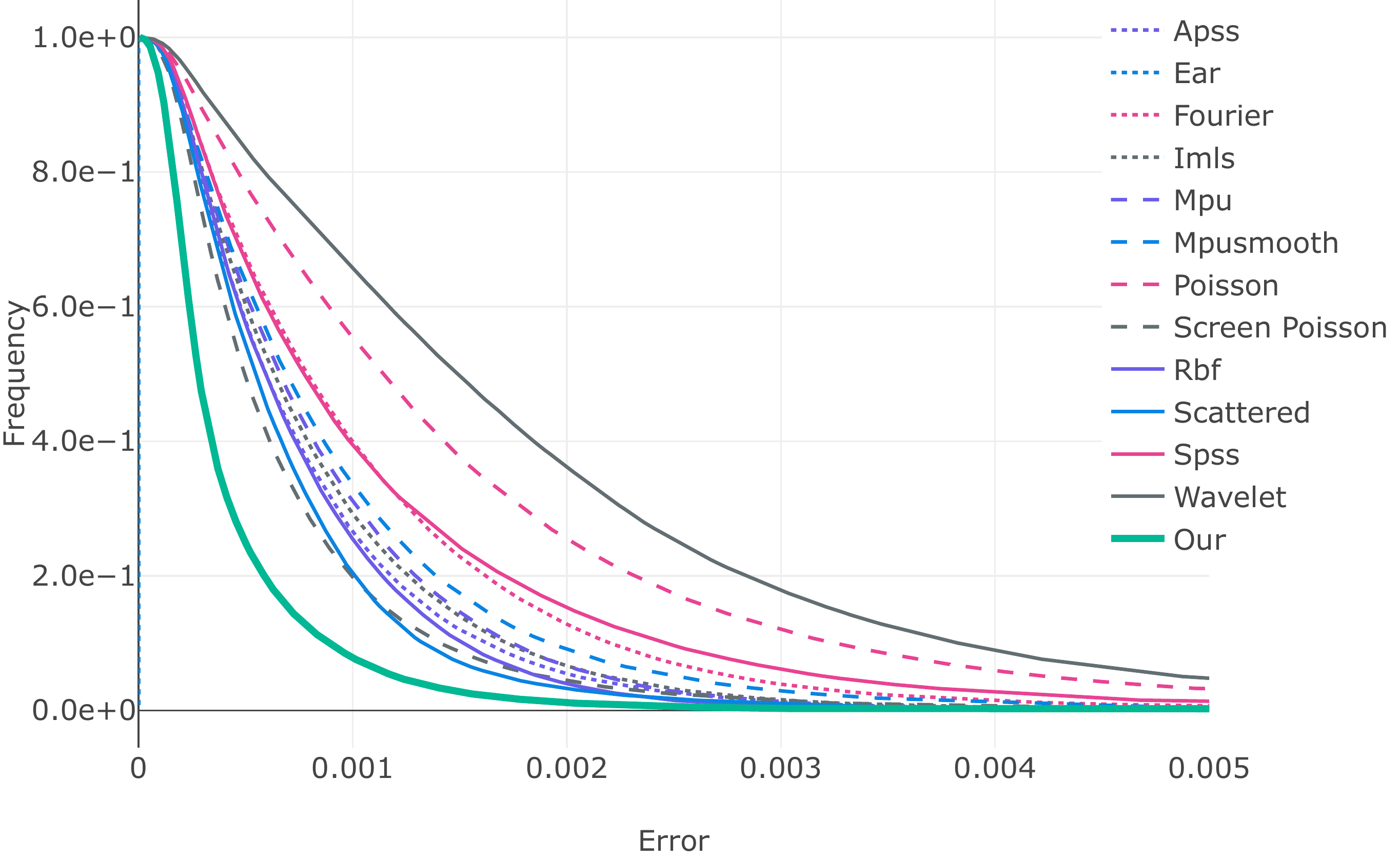}\hfill
\includegraphics[width=.32\linewidth]{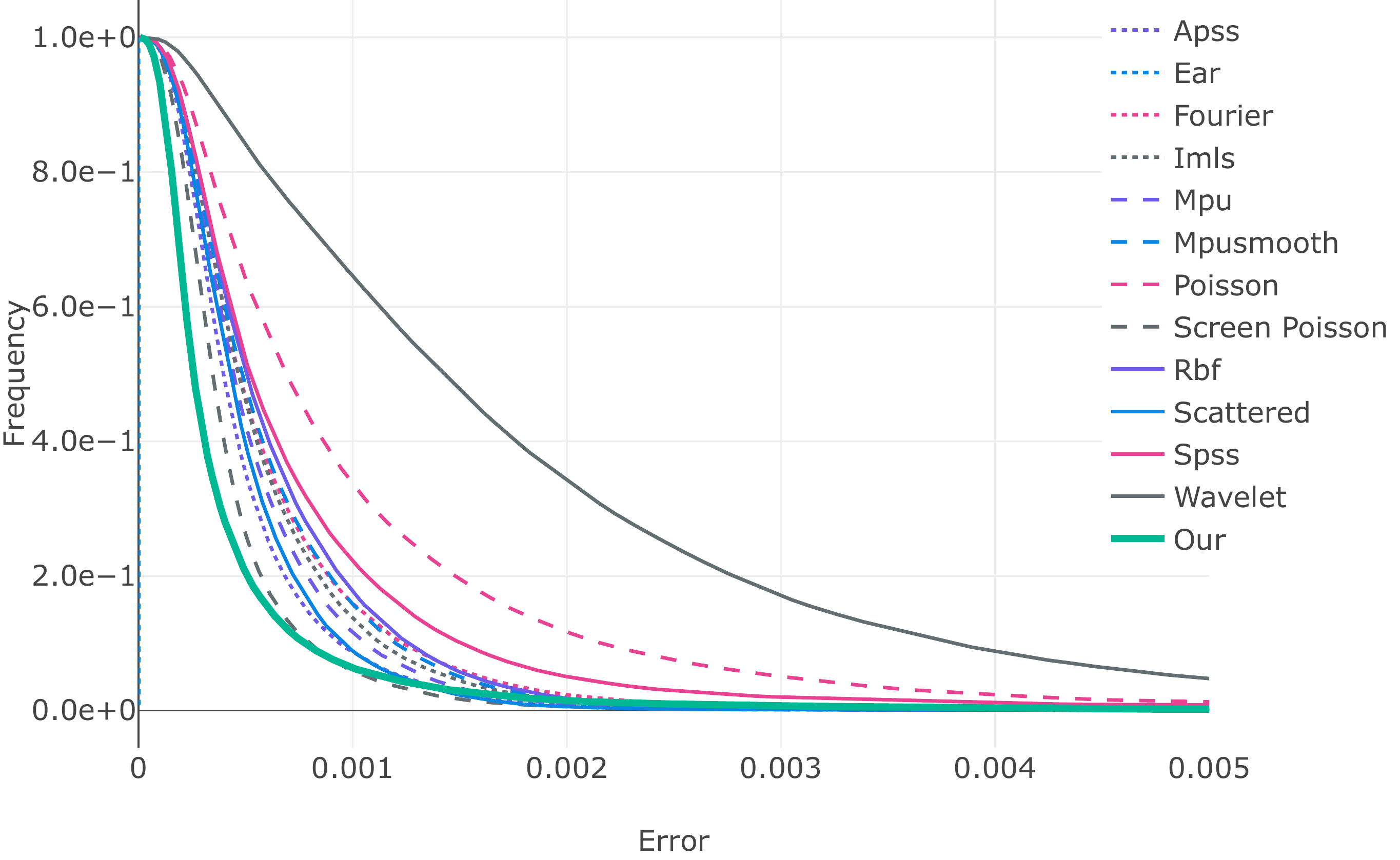}\hfill
\includegraphics[width=.32\linewidth]{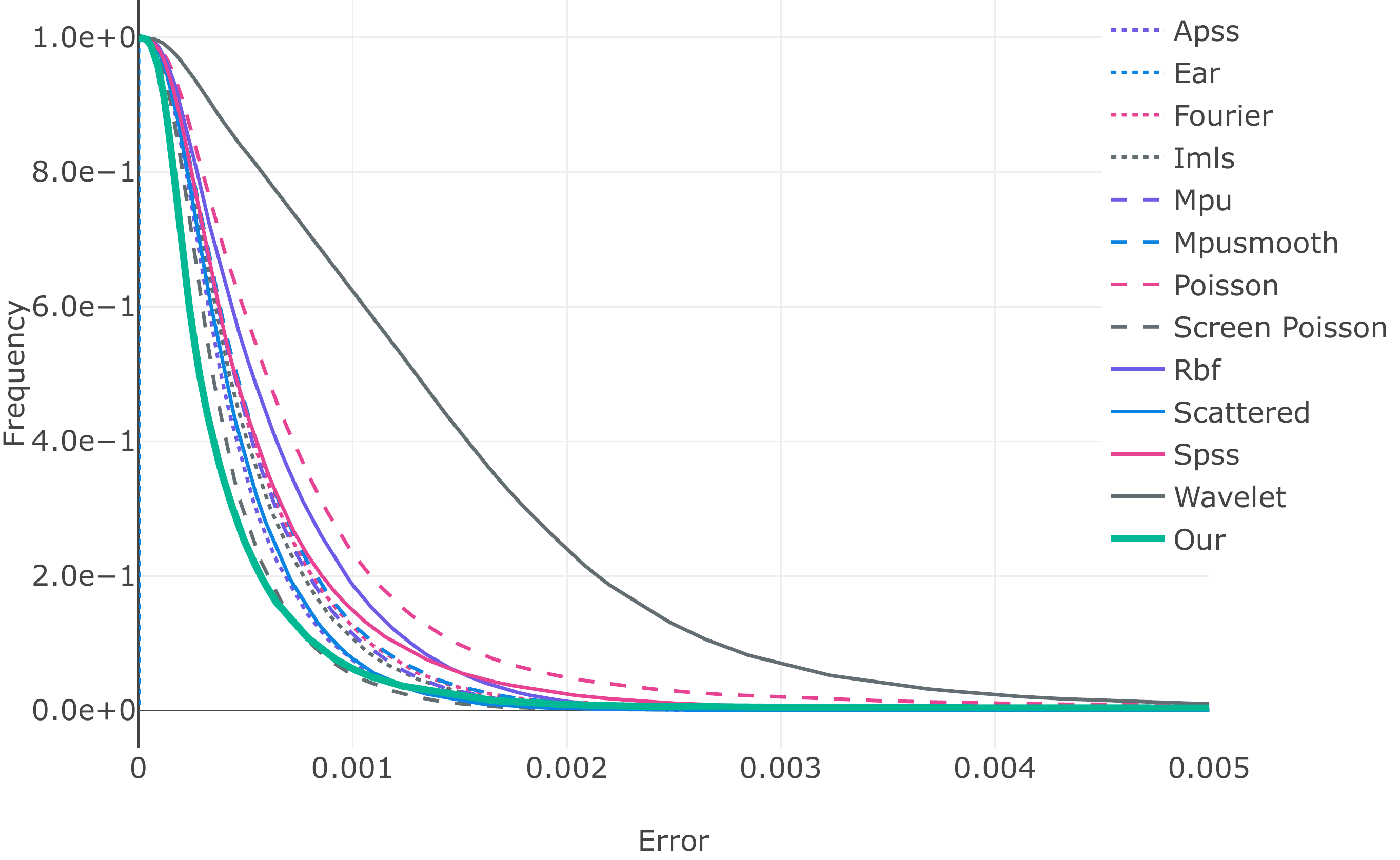}\hfill
\parbox{.32\linewidth}{\centering Daratech}\hfill
\parbox{.32\linewidth}{\centering Danching Children}\hfill
\parbox{.32\linewidth}{\centering Lord Quas}\par
    \caption{Percentage of fitted vertices ($y$-axis) to reach a certain error level ($x$-axis) for different methods. The errors are measured as distance from the input data to the fitted surface.}
    \label{fig:second-result-supp}
\end{figure*}

We provide cumulative histograms for the results of the Surface Reconstruction Benchmark \cite{BergerLNTS13} on all 5 models shown in Figure \ref{fig:srb-modesl}. 
Figures \ref{fig:first-result-supp} and~\ref{fig:second-result-supp} show respectively the percentage of vertices of $\widehat{\mathcal{Y}}$ and $\mathcal{X}$ such that $d_{\mathrm{rec} \to \mathrm{GT}}$ and $d_{\mathrm{inp} \to \mathrm{rec}}$ is below a given error.

\subsection{Surface Reconstruction Benchmark Statistics}
In addition to the cumulative histograms above, we tabulate the mean, standard deviation, and maximum values for each method and model in the benchmark. Table~\ref{tab:gt_to_recon_last} show the distance from the input to the reconstruction ($d_{\mathrm{inp} \to \mathrm{rec}}$) and Table \ref{tab:recon_to_gt_last} show the distance from the reconstruction to the input ($d_{\mathrm{rec} \to \mathrm{GT}}$).

\clearpage
\begin{table}\centering\footnotesize
\begin{tabular}{clcccc}
		&&min&		avg&		std&		max\\
\hline
\multirow{13}{*}{\begin{sideways}Anchor\end{sideways}}
&Apss		&1.32e-05		&8.78e-04		&8.78e-04		&9.21e-03\\
&Ear		&3.30e-09		&3.18e-07		&3.18e-07		&7.75e-07\\
&Fourier		&8.95e-06		&1.34e-03		&1.34e-03		&1.34e-01\\
&Imls		&1.11e-05		&9.00e-04		&9.00e-04		&9.43e-03\\
&Mpu		&4.07e-06		&8.73e-04		&8.73e-04		&8.45e-03\\
&Mpusmooth		&9.50e-06		&9.13e-04		&9.13e-04		&1.06e-02\\
&Poisson		&1.09e-05		&1.63e-03		&1.63e-03		&1.17e-01\\
&Screen Poisson		&6.99e-06		&7.45e-04		&7.45e-04		&1.79e-02\\
&Rbf		&1.46e-05		&8.61e-04		&8.61e-04		&9.89e-03\\
&Scattered		&1.33e-05		&8.18e-04		&8.18e-04		&1.06e-02\\
&Spss		&8.67e-06		&1.03e-03		&1.03e-03		&1.06e-02\\
&Wavelet		&8.30e-06		&2.19e-03		&2.19e-03		&6.27e-02\\
&Our		&4.91e-06		&7.21e-04		&7.21e-04		&2.55e-02\\
\hline
\multirow{13}{*}{\begin{sideways}Daratech\end{sideways}}
&Apss		&9.98e-06		&7.87e-04		&7.87e-04		&1.05e-02\\
&Ear		&1.34e-09		&2.79e-07		&2.79e-07		&8.18e-07\\
&Fourier		&7.86e-06		&1.06e-03		&1.06e-03		&1.94e-02\\
&Imls		&5.73e-06		&8.35e-04		&8.35e-04		&1.05e-02\\
&Mpu		&5.33e-06		&8.47e-04		&8.47e-04		&8.55e-03\\
&Mpusmooth		&9.87e-06		&9.31e-04		&9.31e-04		&1.87e-02\\
&Poisson		&1.28e-05		&1.58e-03		&1.58e-03		&3.18e-02\\
&Screen Poisson		&3.80e-06		&6.98e-04		&6.98e-04		&1.72e-02\\
&Rbf		&2.12e-06		&7.52e-04		&7.52e-04		&1.08e-02\\
&Scattered		&7.48e-06		&6.97e-04		&6.97e-04		&1.70e-02\\
&Spss		&8.36e-06		&1.12e-03		&1.12e-03		&1.12e-02\\
&Wavelet		&6.13e-06		&1.88e-03		&1.88e-03		&2.27e-02\\
&Our		&5.30e-06		&4.23e-04		&4.23e-04		&1.79e-02\\
\hline
\multirow{13}{*}{\begin{sideways}Dc\end{sideways}}
&Apss		&6.20e-06		&4.98e-04		&4.98e-04		&1.45e-02\\
&Ear		&9.81e-10		&3.18e-07		&3.18e-07		&8.28e-07\\
&Fourier		&7.69e-06		&6.23e-04		&6.23e-04		&2.65e-02\\
&Imls		&9.09e-06		&5.88e-04		&5.88e-04		&1.43e-02\\
&Mpu		&1.07e-05		&5.54e-04		&5.54e-04		&7.08e-03\\
&Mpusmooth		&8.10e-06		&6.14e-04		&6.14e-04		&2.75e-02\\
&Poisson		&6.76e-06		&1.02e-03		&1.02e-03		&2.63e-02\\
&Screen Poisson		&7.12e-06		&4.34e-04		&4.34e-04		&2.70e-02\\
&Rbf		&9.15e-06		&6.40e-04		&6.40e-04		&2.77e-02\\
&Scattered		&4.45e-06		&5.20e-04		&5.20e-04		&2.69e-02\\
&Spss		&3.76e-06		&7.66e-04		&7.66e-04		&1.55e-02\\
&Wavelet		&1.76e-05		&1.82e-03		&1.82e-03		&2.68e-02\\
&Our		&6.10e-06		&3.98e-04		&3.98e-04		&2.48e-02\\
\hline
\multirow{13}{*}{\begin{sideways}Gargoyle\end{sideways}}
&Apss		&7.80e-06		&5.62e-04		&5.62e-04		&6.92e-03\\
&Ear		&1.73e-09		&3.18e-07		&3.18e-07		&7.52e-07\\
&Fourier		&3.94e-06		&7.02e-04		&7.02e-04		&2.55e-02\\
&Imls		&8.39e-06		&6.09e-04		&6.09e-04		&6.49e-03\\
&Mpu		&1.10e-05		&6.27e-04		&6.27e-04		&6.75e-03\\
&Mpusmooth		&4.57e-06		&7.25e-04		&7.25e-04		&8.81e-03\\
&Poisson		&1.20e-05		&1.05e-03		&1.05e-03		&2.73e-02\\
&Screen Poisson		&1.04e-05		&4.87e-04		&4.87e-04		&1.81e-02\\
&Rbf		&6.44e-06		&7.30e-04		&7.30e-04		&5.86e-03\\
&Scattered		&7.73e-06		&5.78e-04		&5.78e-04		&1.03e-02\\
&Spss		&5.30e-06		&7.74e-04		&7.74e-04		&1.34e-02\\
&Wavelet		&1.14e-05		&1.56e-03		&1.56e-03		&2.73e-02\\
&Our		&5.40e-06		&4.50e-04		&4.50e-04		&1.81e-02\\
\hline
\multirow{13}{*}{\begin{sideways}Lord Quas\end{sideways}}
&Apss		&8.64e-06		&4.76e-04		&4.76e-04		&7.55e-03\\
&Ear		&1.05e-09		&3.22e-07		&3.22e-07		&8.91e-07\\
&Fourier		&1.15e-05		&5.64e-04		&5.64e-04		&1.79e-02\\
&Imls		&8.29e-06		&5.29e-04		&5.29e-04		&8.60e-03\\
&Mpu		&7.94e-06		&5.44e-04		&5.44e-04		&5.01e-03\\
&Mpusmooth		&1.07e-05		&5.70e-04		&5.70e-04		&8.18e-03\\
&Poisson		&5.70e-06		&8.24e-04		&8.24e-04		&4.38e-02\\
&Screen Poisson		&4.74e-06		&4.29e-04		&4.29e-04		&1.08e-02\\
&Rbf		&1.01e-05		&6.48e-04		&6.48e-04		&7.27e-03\\
&Scattered		&6.72e-06		&4.88e-04		&4.88e-04		&1.69e-02\\
&Spss		&6.66e-06		&6.03e-04		&6.03e-04		&7.66e-03\\
&Wavelet		&9.27e-06		&1.49e-03		&1.49e-03		&4.71e-02\\
&Our		&1.38e-06		&4.15e-04		&4.15e-04		&2.14e-02\\
\end{tabular}\\[2ex]
\caption{Distance from the input to the reconstruction.}
\label{tab:gt_to_recon_last}
\end{table}

\begin{table}\centering\footnotesize
\begin{tabular}{clcccc}
		&&min&		avg&		std&		max\\
\hline
\multirow{13}{*}{\begin{sideways}Anchor\end{sideways}}
&Apss		&6.28e-06		&1.79e-03		&1.79e-03		&2.80e-02\\
&Ear		&4.96e-06		&1.53e-03		&1.53e-03		&9.93e-03\\
&Fourier		&1.45e-06		&2.01e-03		&2.01e-03		&6.59e-02\\
&Imls		&6.49e-06		&1.92e-03		&1.92e-03		&2.82e-02\\
&Mpu		&2.23e-06		&2.08e-03		&2.08e-03		&4.59e-02\\
&Mpusmooth		&4.81e-06		&1.87e-03		&1.87e-03		&3.66e-02\\
&Poisson		&8.75e-06		&2.27e-03		&2.27e-03		&6.59e-02\\
&Screen Poisson		&7.12e-06		&2.15e-03		&2.15e-03		&6.59e-02\\
&Rbf		&1.62e-06		&2.98e-03		&2.98e-03		&6.60e-02\\
&Scattered		&4.30e-06		&2.16e-03		&2.16e-03		&6.57e-02\\
&Spss		&4.15e-06		&4.39e-03		&4.39e-03		&9.00e-02\\
&Wavelet		&5.36e-06		&3.01e-03		&3.01e-03		&6.59e-02\\
&Our		&3.82e-06		&1.53e-03		&1.53e-03		&9.69e-03\\
\hline
\multirow{13}{*}{\begin{sideways}Daratech\end{sideways}}
&Apss		&2.18e-06		&1.68e-03		&1.68e-03		&2.16e-02\\
&Ear		&3.15e-06		&1.50e-03		&1.50e-03		&1.68e-02\\
&Fourier		&4.68e-06		&2.39e-03		&2.39e-03		&5.98e-02\\
&Imls		&3.46e-06		&1.79e-03		&1.79e-03		&2.36e-02\\
&Mpu		&5.25e-06		&2.06e-03		&2.06e-03		&4.11e-02\\
&Mpusmooth		&5.98e-06		&2.14e-03		&2.14e-03		&4.92e-02\\
&Poisson		&3.25e-06		&2.98e-03		&2.98e-03		&6.83e-02\\
&Screen Poisson		&4.30e-06		&2.42e-03		&2.42e-03		&6.47e-02\\
&Rbf		&2.28e-06		&3.59e-03		&3.59e-03		&6.44e-02\\
&Scattered		&5.40e-06		&1.60e-03		&1.60e-03		&1.81e-02\\
&Spss		&5.99e-06		&2.73e-03		&2.73e-03		&7.66e-02\\
&Wavelet		&4.04e-06		&3.29e-03		&3.29e-03		&5.04e-02\\
&Our		&1.96e-06		&1.51e-03		&1.51e-03		&1.67e-02\\
\hline
\multirow{13}{*}{\begin{sideways}Dc\end{sideways}}
&Apss		&5.53e-06		&1.68e-03		&1.68e-03		&2.87e-02\\
&Ear		&2.54e-06		&1.32e-03		&1.32e-03		&2.46e-02\\
&Fourier		&3.51e-06		&1.60e-03		&1.60e-03		&2.53e-02\\
&Imls		&6.12e-06		&1.75e-03		&1.75e-03		&2.93e-02\\
&Mpu		&3.98e-06		&1.53e-03		&1.53e-03		&3.84e-02\\
&Mpusmooth		&3.51e-06		&1.47e-03		&1.47e-03		&1.64e-02\\
&Poisson		&5.86e-06		&1.87e-03		&1.87e-03		&2.53e-02\\
&Screen Poisson		&3.96e-06		&1.49e-03		&1.49e-03		&2.25e-02\\
&Rbf		&1.15e-06		&1.55e-03		&1.55e-03		&3.01e-02\\
&Scattered		&5.26e-06		&1.89e-03		&1.89e-03		&5.27e-02\\
&Spss		&5.12e-06		&1.96e-03		&1.96e-03		&2.93e-02\\
&Wavelet		&4.70e-06		&2.63e-03		&2.63e-03		&2.56e-02\\
&Our		&3.10e-06		&1.31e-03		&1.31e-03		&1.51e-02\\
\hline
\multirow{13}{*}{\begin{sideways}Gargoyle\end{sideways}}
&Apss		&2.64e-06		&1.50e-03		&1.50e-03		&3.41e-02\\
&Ear		&4.04e-06		&1.22e-03		&1.22e-03		&8.81e-03\\
&Fourier		&2.29e-06		&1.37e-03		&1.37e-03		&2.15e-02\\
&Imls		&2.11e-06		&1.71e-03		&1.71e-03		&4.37e-02\\
&Mpu		&4.60e-06		&1.57e-03		&1.57e-03		&2.98e-02\\
&Mpusmooth		&1.20e-06		&1.37e-03		&1.37e-03		&2.17e-02\\
&Poisson		&6.48e-06		&1.57e-03		&1.57e-03		&2.17e-02\\
&Screen Poisson		&2.82e-06		&1.30e-03		&1.30e-03		&2.09e-02\\
&Rbf		&3.19e-06		&7.66e-03		&7.66e-03		&9.17e-02\\
&Scattered		&2.48e-06		&1.36e-03		&1.36e-03		&2.17e-02\\
&Spss		&5.03e-06		&1.85e-03		&1.85e-03		&4.65e-02\\
&Wavelet		&4.79e-06		&1.89e-03		&1.89e-03		&2.11e-02\\
&Our		&2.34e-06		&1.19e-03		&1.19e-03		&1.45e-02\\
\hline
\multirow{13}{*}{\begin{sideways}Lord Quas\end{sideways}}
&Apss		&3.90e-06		&1.24e-03		&1.24e-03		&2.14e-02\\
&Ear		&2.26e-06		&1.14e-03		&1.14e-03		&7.49e-03\\
&Fourier		&5.01e-06		&1.30e-03		&1.30e-03		&1.96e-02\\
&Imls		&3.30e-06		&1.31e-03		&1.31e-03		&2.36e-02\\
&Mpu		&2.30e-06		&1.35e-03		&1.35e-03		&2.94e-02\\
&Mpusmooth		&4.70e-06		&1.28e-03		&1.28e-03		&1.54e-02\\
&Poisson		&1.91e-06		&1.39e-03		&1.39e-03		&1.94e-02\\
&Screen Poisson		&2.07e-06		&1.24e-03		&1.24e-03		&1.62e-02\\
&Rbf		&2.26e-06		&1.29e-03		&1.29e-03		&1.80e-02\\
&Scattered		&4.48e-06		&1.17e-03		&1.17e-03		&1.32e-02\\
&Spss		&1.49e-06		&1.38e-03		&1.38e-03		&2.28e-02\\
&Wavelet		&4.42e-06		&1.87e-03		&1.87e-03		&1.52e-02\\
&Our		&3.94e-06		&1.14e-03		&1.14e-03		&8.74e-03\\
\end{tabular}\\[2ex]
\caption{Distance from the reconstruction to the input.}
\label{tab:recon_to_gt_last}
\end{table}

\end{document}